\documentclass{article}
\usepackage[preprint]{cpal_2024}

\usepackage{graphicx} 
\usepackage{algorithm}
\usepackage{algorithmic}
\usepackage{booktabs}
\usepackage{diagbox}
\usepackage{multirow}
\usepackage{color}
\usepackage{subfigure}
\usepackage{wrapfig}
\usepackage{amsmath}
\usepackage[pagebackref,breaklinks,colorlinks]{hyperref}
\usepackage{enumitem}
\usepackage{newfloat}
\usepackage{listings}
\floatstyle{ruled}
\newfloat{listing}{tb}{lst}{}
\floatname{listing}{Listing}
\title{ASP: Automatic Selection of Proxy dataset for efficient AutoML}
\author{
Peng Yao, Chao Liao, Jiyuan Jia, Jianchao Tan\thanks{Corresponding Author.}, ~Bin Chen, Chengru Song, Di Zhang
\\
Kuaishou Technology
}

\begin{document}

\maketitle

\begin{abstract}
Deep neural networks have gained great success due to the increasing amounts of data, and diverse effective neural network designs. However, it also brings a heavy computing burden as the amount of training data is proportional to the training time. In addition, a well-behaved model requires repeated trials of different structure designs and hyper-parameters, which may take a large amount of time even with state-of-the-art (SOTA) hyper-parameter optimization (HPO) algorithms and neural architecture search (NAS) algorithms. In this paper, we propose an Automatic Selection of Proxy dataset framework (ASP) aimed to dynamically find the informative proxy subsets of training data at each epoch, reducing the training data size as well as saving the AutoML processing time. We verify the effectiveness and generalization of ASP on CIFAR10, CIFAR100, ImageNet16-120, and ImageNet-1k, across various public model benchmarks. The experiment results show that ASP can obtain better results than other data selection methods at all selection ratios. ASP can also enable much more efficient AutoML processing with a speedup of 2x-20x while obtaining better architectures and better hyper-parameters compared to utilizing the entire dataset.
\end{abstract}

\section{Introduction}
Automated machine learning (AutoML), is the process of automating the model development and iteration with large scalability, high efficiency, and less expert knowledge~\cite{zoller2021benchmark}. As the two main tasks of AutoML, neural architecture search (NAS) and hyper-parameter optimization (HPO) have become very active research fields.
However, both of them still suffer from huge computing costs (for example, GPU days). 
Recently, some acceleration NAS algorithms have been proposed, such as One-shot NAS~\cite{bender2018understanding, guo2020single} and differentiable NAS~\cite{liu2018darts, wu2019fbnet}. The predictor-based and zero-cost metric-based performance estimator can further speed up the search procedure.
Similarly, in the field of HPO~\cite{falkner2018bohb}, it combined the Bayesian optimization~\cite{bergstra2011algorithms} and Hyper-band search to obtain fast and better hyper-parameters with the early-exit mechanism.
Although the above algorithms have greatly improved their efficiency, they still require prohibitive computing resources and a mass of training data.

To further accelerate the AutoML procedure in terms of the dataset,~\cite{nickson2014automated, krueger2015fast} discovered that using randomly sampled small-size subsets can quickly find better hyper-parameters. 
\cite{park2019data} proposed a core-set method to select a fixed unique and informative subset by a pre-trained model for NAS acceleration. 
\cite{katharopoulos2018not, zhang2019autoassist} adopted importance sampling methods on the training dataset to accelerate the training of neural networks. They calculated the gradients of each sample as importance scores and only used the data with larger scores during the training.
\cite{killamsetty2022automata} attempted to apply HPO on a sub-dataset dynamically selected using data gradients as information during training, with an aim of exploring appropriate hyper-parameters more effectively.
The above methods use proxy data (either pre-selected or dynamically selected during training) to achieve obvious acceleration, however, there are several existing problems:
Firstly, the subset pre-selection methods often require a pre-trained model which will require extra computational steps and resource consumption. 
Secondly, the subset data selected by a pre-trained model will have different distributions from the original dataset. It will introduce a prior bias into the training of the new model and possibly lead to the degradation of model performance.
Thirdly, even for the dynamic selecting methods of the dataset, \textbf{it is noted that different importance metrics play different roles at different stages of model training}.
For example, using the data gradient as an importance metric for all epochs is resource-unfriendly and maybe not optimal. 
\textbf{Thus, the dynamic selection of metrics during the training procedure may be optimal.}

Inspired by the aforementioned works, we proposed a unified framework \textbf{A}utomatic \textbf{S}election of \textbf{P}roxy dataset (ASP) for various AutoML tasks, the method of core-set selection and importance sampling can be considered as two special cases of our ASP.
Under the ASP framework, all the samples can have chances to participate in model training, where each sample is dynamically activated and deactivated according to the importance metric ranking, and the importance metric is also dynamically chosen at different stages of training.
The importance metric of a sample can be computed from different views. The training loss or sample gradient is suitable for any training tasks; the training entropy or prediction accuracy~\cite{na2021accelerating} are mostly suitable for classification tasks.
Our ASP framework maintains a Dict type data structure to record the various importance metrics of each sample during the training process.
Samples with the higher importance rankings at the current moment are more likely to be activated and participate in model training.
\textbf{With the convergence of training, the model's ability to recognize data becomes stronger, and its importance ranking will gradually decrease. Conversely, as the importance ranking of previously deactivated samples rises, their probability of being activated will gradually increase, and they will participate in the training process again. Thus, the option of dynamic data selection with replacement is necessary}.
To the best of our knowledge, ASP is the SOTA data selection framework for AutoML which performs a dynamic data selection with replacement during the training or searching.
\begin{figure}[!t]
\centering
\includegraphics[width=0.45\columnwidth]{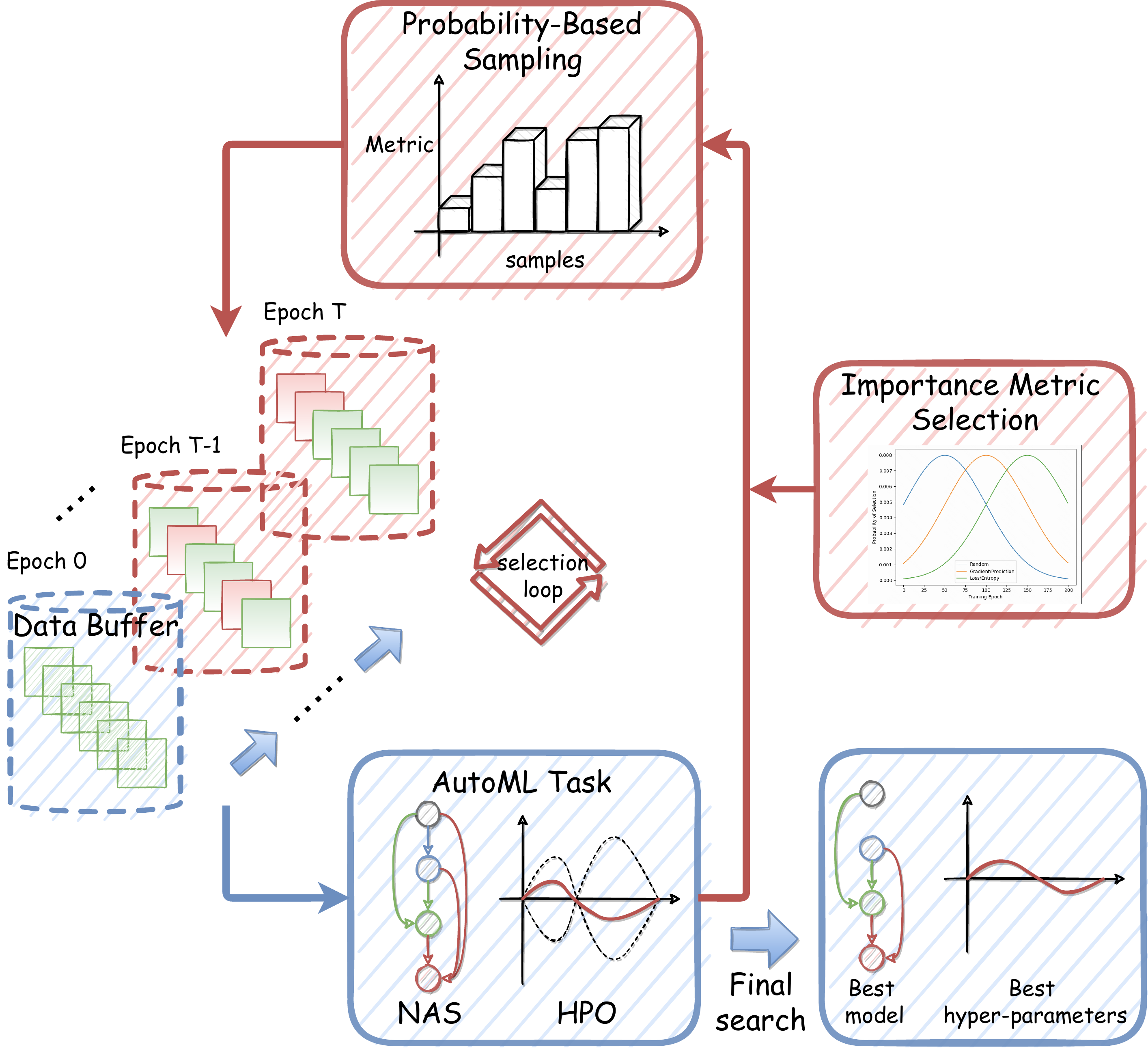} 
\caption{The overview of ASP framework. The red squares inside the Data Buffer represent the currently deactivated samples, and the green squares represent the currently activated samples; The previously deactivated samples may be activated again after one epoch of selection.}
\label{fig:asp}
\end{figure}

Our contributions are as follows:
\begin{itemize}[itemsep=2pt,topsep=2pt,parsep=1pt]
\item We proposed a data selection framework to dynamically activate the training data at each epoch and only exploit the proxy subset with the highest contribution to the current model. The dynamic sample ratio and dynamic metric selection accelerate model convergence.
\item Our proposed ASP framework unifies the core-set pre-selection and importance sampling during training in a concise way, which is general to any metrics for selecting proxy data.
\item Extensive experimental results demonstrate the superiority of ASP. We achieve better performance compared with other state-of-the-art.
The relevant NAS and HPO experiments show that one can use the 10\% proxy data selected by ASP to reduce the search time by 10 times and can still obtain potentially competitive architectures and hyper-parameters.
\end{itemize}

\section{Related Works}

\subsection{AutoML}
AutoML~\cite{hutter2019automated} aims to automate the model development and iteration with less expert knowledge.
However, AutoML usually suffers from heavy computational resources. 
In terms of NAS, some differentiable and one-shot NAS methods are proposed to reduce the search cost, they adopt acceleration strategies such as weight sharing~\cite{pham2018efficient} and continuous relaxation of the discrete search space~\cite{liu2018darts, zoph2018learning}. 
HPO aims to search for effective training hyper-parameters on the given model and dataset. A naive way is Grid Search~\cite{claesen2014easy} or Random Search~\cite{pinto2009high}, however, both of them are time-consuming since the search space may be very large. Compared to the naive methods, sequential model-based optimization (SMBO) methods aim to learn a regression model in an iteration way, where~\cite{bergstra2011algorithms} adopted the Bayesian optimization to optimize the hyper-parameters and proposed GP and TPE algorithms, but it still needs to train all the model from scratch to end. 
Sequential Halving algorithm (SHA)~\cite{jamieson2016non} adopts an early-exist mechanism and allocates an increasingly exponential amount of resources to better configures.~\cite{li2020system} improves the SHA algorithm and devises an asynchronous parallel training framework.

\subsection{Proxy Data Selection}
Proxy data selection is a well-established methodology in deep learning that can save computing resources and improve research efficiency. 
Core-set selection~\cite{shleifer2019using, killamsetty2021grad, killamsetty2021glister} can be defined as a subset of original data that maintains a similar “quality” as the full dataset for cheap search experiments.~\cite{mirzasoleiman2020coresets} tries to select a core-set dataset to better estimate the gradients of the entire dataset. 
Importance sampling assumes that samples are not equally important, and the limited computational resources should be used to train those samples that produce larger gradients for the parameters of models, and without an additional pre-trained model.
~\cite{katharopoulos2018not, zhang2019autoassist}.~\cite{johnson2018training} derives a tractable upper bound to each sample gradient norm to accelerate the gradient computation.~\cite{jiang2019accelerating} adopt loss as a cheap alternative to gradients, reducing the number of computationally expensive backpropagation steps performed and accelerating training. 
Dataset Distillation~\cite{wang2018dataset} and Dataset Condensation~\cite{zhao2021dataset} propose similar approaches in which models train on small-sized datasets of the synthetic dataset and reach comparable performance with the original datasets.

\subsection{Proxy strategy for AutoML}
~\cite{coleman2019selection} use a pre-trained model to perform forward propagation on the entire dataset, and then select a proxy subset according to the forgetting events as the training data.
EcoNAS~\cite{zhou2020econas} proposed four strategies to accelerate the NAS, including searching architectures on a proxy subset randomly sampled from the whole dataset.
~\cite{na2021accelerating} draw support of pre-trained model and adopted probabilistic sample strategy to eliminate those samples whose entropy value is in the middle among datasets for NAS acceleration.  
~\cite{killamsetty2022automata} attempted to apply HPO on sub-datasets selected based on data gradients to explore appropriate hyper-parameters faster and more effectively.
In this work, we propose ASP, an efficient and adaptive proxy data selection framework, to further improve the efficiency of AutoML algorithms while maintaining performance.

\section{Methodology}


\subsection{Problem formulation}
The goal of NAS or HPO is to find the best architecture parameters $\mathcal{A}^{*}$ or the best training hyper-parameters $\mathcal{H}^{*}$ in the huge search space. To be specific, in terms of NAS, architecture parameters $\mathcal{A}$ can be optimized by minimizing loss function $\mathcal{L}_{\mathcal{D}_{val}}(\mathcal{W}^{*}(\mathcal{A}), \mathcal{A})$ on validation dataset, where the weight of supernet $\mathcal{W(A)}$ are associated with $\mathcal{A}$ and the optimal weight $\mathcal{W}^{*}(\mathcal{A})$ can be obtained by minimizing the training loss $\mathcal{L}_{\mathcal{D}_{train}}(\mathcal{W}(\mathcal{A}), \mathcal{A}^{*})$. 
Similarly, training hyper-parameters also can be optimized by minimizing the validation loss $\mathcal{L}_{\mathcal{D}_{val}}(\mathcal{W}^{*}, \mathcal{H})$ and the optimal weight $\mathcal{W}^{*}$ can be obtained by minimizing the training loss $\mathcal{L}_{\mathcal{D}_{train}}(\mathcal{W}, \mathcal{H}^{*})$.

The above descriptions reveal a bi-level optimization problem, the architecture parameters $\mathcal{A}$ of NAS or training hyper-parameters $\mathcal{H}$ of HPO can be considered as an upper-level variable while the weight $\mathcal{W}$ is the lower-level variable~\cite{colson2007overview, franceschi2018bilevel}. To avoid the time-consuming process of alternately training one variable to convergence and then training another variable to convergence, some literature uses the first-order approximate gradient information in training set ${\mathcal{D}_{train}}$ and validation set ${\mathcal{D}_{val}}$ to update these two variables at each iteration alternately~\cite{liu2018darts}, which indeed accelerated the searching process. 
In this paper, we propose a novel data selection framework ASP and combine it with NAS and HPO tasks to further accelerate the optimization process while maintaining high performance. \textbf{Data selection and AutoML can be jointly optimized, and we re-formulate it as}: 
\begin{align*}
   \mathop{\min}_{\mathcal{A}, \mathcal{H}}\ \mathcal{L}_{\mathcal{D}_{val}}& (\mathcal{W}^{*}(\mathcal{A}), \mathcal{A}, \mathcal{H}, \mathcal{D}^{*}_{proxy})\\
s.t.\quad  \mathcal{W}^{*}(\mathcal{A}) &= \mathop{\arg\min}_{\mathcal{W}}\ \mathcal{L}_{D^{*}_{proxy}}(\mathcal{W}, \mathcal{A}, \mathcal{H})\\
\quad\quad  \mathcal{D}_{proxy}^{*} &= \mathop{\arg\max}_{\mathcal{D}_{train}}\ \textit{Metric}(\mathcal{D}_{train})
\end{align*}

where $\mathcal{D}_{train}$ denotes the training dataset containing $N$ samples, $\mathcal{D}_{proxy}$ denotes the subset containing $M$ samples. The samples of $\mathcal{D}_{proxy}$ are selected from $\mathcal{D}_{train}$ according to the importance $\textit{Metric}$ (we will introduce the $\textit{Metric}$ in the following section), thus $\mathcal{D}_{proxy} \in \mathcal{D}_{train}$ and $M < N$. 

We can observe that ASP uses $\mathcal{D}_{proxy}$ to update $\mathcal{W}$ instead of the entire training dataset $\mathcal{D}_{train}$. $\mathcal{D}_{proxy}$ is generated according to the $\textit{Metric}$ of the samples in $\mathcal{D}_{train}$, which is a tri-level optimization problem that includes three variables: the parameters of AutoML $\mathcal{A}$ (or $\mathcal{H}$), the network weight $\mathcal{W}$, and $\mathcal{D}_{proxy}$. It should be noted that we re-formulate these three tasks: regular model training, NAS, and HPO in a unified form. For regular deep learning model training, only $\mathcal{W}$ and $\mathcal{D}_{proxy}$ need to be optimized, and the tri-level optimization problem will be downgraded to bi-level optimization. For NAS and HPO, training hyper-parameters $\mathcal{H}$ and architecture parameters $\mathcal{A}$ are fixed respectively.

Inspired by previous work~\cite{liu2018darts}, we use the method of alternately updating three variables to address this problem. Specifically, we repeat the following two steps within an epoch: First, we update $\mathcal{W}$ using the first-order approximation information on a mini-batch of samples in $\mathcal{D}_{proxy}$, which means that we approximate $\mathcal{W}^{*}$ by updating $\mathcal{W}$ with only a single training iteration (iteration-level); then, we update $\mathcal{H}$ or $\mathcal{A}$ on a mini-batch of samples in $\mathcal{D}_{val}$. 
However, the first-order approximation solution is inherently uncertain because it ignores the convergence of the algorithm. And, in each update, we actually only use the local information (a mini-batch) of the entire training set. In particular, we record the importance metrics $\textit{metric}$ on the entire dataset $\mathcal{D}_{proxy}$ during the training and update $\mathcal{D}_{proxy}$ after one epoch of training according to the recorded $\textit{metric}$, which means an epoch-level update for variable $\mathcal{D}_{proxy}$. We suggest that it significantly increases the robustness of tri-level optimization problems, as epoch-level updates better guarantee algorithm convergence and utilize the global information from the entire dataset.

\subsection{Overview of ASP}

As illustrated in Figure \ref{fig:asp}, we proposed an end-to-end data selection framework ASP to jointly optimize AutoML tasks and the Data Selection task.
In the first training epoch, all samples are used for training (all squares in the Data Buffer of  Figure \ref{fig:asp} are green). Next, we perform the training of AutoML, and the importance metric of each sample will be recorded and updated. Then, ASP will dynamically select an importance metric and define a sampling ratio according to a scheduler.
Finally, select top-m activated samples based on the probabilistic methods to form the proxy dataset $\mathcal{D}_{proxy}$. In the next training epoch, only samples in $\mathcal{D}_{proxy}$ will be activated for training. The above process will be repeated multiple times until the training time is exhausted. We put the overall algorithm pseudocode of ASP in \textbf{the appendix.}




\subsection{Mixture of Multi-metrics}\label{sec:metric}
The type of importance metric directly influences the quality of data selection. Previous data selection methods use a single importance metric to judge the importance of samples, we experimentally find that only one metric is not sufficient. We propose a mixture of multi-metrics in ASP to adaptively select different metrics during the various stages of training. 
Here we provide a brief description of five commonly used importance metrics, the alternate appearance of which during training constitutes our mixture metric.

\paragraph{Random Metric} 
Each sample in the training dataset has an equal probability of being chosen and all the data will be shuffled when a new epoch of training starts~\cite{moser2022less}. 

\paragraph{Gradient Metric}
Gradient metric is often used in importance sampling~\cite{aljundi2019gradient}, which claims that the model should more focus on those samples that can generate large updates to model parameters.


\paragraph{Loss Metric}
The forward loss is always used to judge how hard the sample is to train and is regarded as a computationally cheap proxy for the gradient norm in importance sampling works~\cite{eric2021important}.
Usually, the low loss corresponds to gradients with small norms and thus contributes little to the parameters update, which has been confirmed by experiments~\cite{jiang2019accelerating}. 


\paragraph{Entropy Metric}
For the classification task, entropy is also an important metric that the higher entropy indicates the lower confidence and vice versa~\cite{settles2009active}. Compared to loss, entropy can measure how well the model fits the sample in a finer-grained way. 



\paragraph{Prediction Metric}
The model prediction shows whether a sample is correctly predicted by the model and has been adopted in many works~\cite{toneva2018empirical, zhang2021efficient, na2021accelerating}, which counted transitions number of a sample taking from being predicted correctly to being predicted wrong, called "forgetting events". 


\paragraph{Mixture Metric} 
In general, we experimentally find that the above five importance metrics can be divided into 3 categories, and using different metrics at different stages of training can significantly improve the quality of data selection. 
In the early training stage, data diversity is more important, so \textit{Random} metric is better.
In the middle training stage, the model needs to converge quickly and improve the sample prediction rate, so \textit{Prediction} and \textit{Gradient} are more important. 
In the later training stage, the model enters a stable convergence period and the fine-grained predictive distribution calculated by \textit{Loss} and \textit{Entropy} are more critical. 
Additionally, to better balance exploration and exploitation, we do not fix the metric in each stage but combine 5 metrics based on a certain probability distribution. Due to the page limitation, more details and the probability distribution of each metric during the training procedure are in \textbf{the appendix}.

\subsection{Probability-based Data Selection with Replacement}
From Figure~\ref{fig:asp}, we can see that only partial samples will be activated during training and other samples will be deactivated. 
We avoid a situation where once a sample is deactivated in the current training stage, it will never participate in later training, from the selection mechanism. Specifically, under the ASP framework, the samples are selected based on probability corresponding to importance metric, the deactivated sample's probability of being activated again will gradually increase with training, and it will eventually participate in the later training.

\paragraph{Probability-based Sampling} 
We first propose a proxy memory mechanism (PMM) $V = \{v_{0},...,v_{N}\}$ to record the importance metric of all samples, $v_{i}$ presents the importance of $i_{th}$ sample which is positively correlated to the probability of being activated. 
During the forward propagation and backward propagation of the model, PMM automatically collects the importance metric of each sample and updates the proxy memory $V$ at the end of each epoch. 
In addition, the importance metric is further normalized with the $\mathbf{softmax}$ function to obtain better sampling and ranking properties. The final sampling probability is calculated as: $\mathcal{P}(V)=\textit{Softmax}(V)$.
Then, $M$ samples will be activated based on the probability distribution $\mathcal{P}(V)$, and samples with higher sampling probability are more likely to be selected.

\paragraph{Selection with Replacement}
The proxy data selected by ASP has a weak critical representation for the current model. With the model learning on the proxy data, the model's ability to fit these samples becomes higher, and then the importance metric of the proxy data will gradually decrease.
Since the deactivated samples do not participate in training, their importance metric will not be updated. Due to the $\mathbf{softmax}$ normalization, the decrease in the sampling probability of the activated samples will indirectly lead to an increase in the sampling probability of the deactivated samples, which means these deactivated samples will have the opportunity to participate in later training.





\subsection{Sampling Ratio Scheduler}
Without considering the bottleneck effect of CPU on data preprocessing, the amount of data is always proportional to the training time.
Given a resource budget $|\mathcal{D}_{proxy}|=|\mathcal{D}_{train}| * r$, we can control resource allocation in two ways: constant and dynamic allocation. Constant allocation selects samples with a constant ratio $r$ every epoch and dynamic allocation selects samples with dynamic $r^{'}$ every epoch but expected ratio $E(r^{'})=r$. The detailed calculation is provided in \textbf{the appendix}.

\begin{table*}
\centering
\scriptsize
\begin{tabular}{l|c|c|c|c|c|c|c}
\toprule
\multirow{2}{*}{Method}    & Search     & \multicolumn{2}{c|}{CIFAR-10}                        & \multicolumn{2}{c|}{CIFAR-100}                         & \multicolumn{2}{c}{ImageNet-16-120}          \\ \cline{2-8} 
                                    & (Seconds)  & \multicolumn{1}{l|}{validation}  & test     & \multicolumn{1}{l|}{validation}  & test       & \multicolumn{1}{l|}{validation}  & test       \\ \hline
Optimal                             & -                   & 91.61                                     & 94.37             & 73.49                                     & 73.51                     & 46.77                               & 47.31      \\ \hline
RSPS                       & 80007.13   & 80.42$\pm$3.58                  & 84.07$\pm$3.61   & 52.12$\pm$5.55                      & 52.31$\pm$5.77            & 27.22$\pm$3.24                      & 26.28$\pm$3.09 \\
DARTS-V1                   & 11625.77   & 39.77$\pm$0.00                  & 54.30$\pm$0.00   & 15.03$\pm$0.00                      & 15.61$\pm$0.00            & 16.43$\pm$0.00                      & 16.32$\pm$0.00 \\
DARTS-V2                   & 35781.80   & 39.77$\pm$0.00                  & 54.30$\pm$0.00   & 15.03$\pm$0.00                      & 15.61$\pm$0.00            & 16.43$\pm$0.00                      & 16.32$\pm$0.00 \\
GDAS                       & 31609.80   & 89.89$\pm$0.08                  & 93.61$\pm$0.09   & 71.34$\pm$0.04                      & 70.70$\pm$0.30            & 41.59$\pm$1.33                      & 41.71$\pm$0.98 \\
SETN                       & 34139.53   & 84.04$\pm$0.28                  & 87.64$\pm$0.00   & 58.86$\pm$0.06                      & 59.05$\pm$0.24            & 33.06$\pm$0.02                      & 32.52$\pm$0.21 \\
ENAS                       & 14058.80   & 37.51$\pm$3.19                  & 53.89$\pm$0.58   & 13.37$\pm$2.35                      & 13.96$\pm$2.33            & 15.06$\pm$1.95                      & 14.84$\pm$2.10 \\ 
\hline
SPOS                       & 7127.67    & 89.63$\pm$1.05                  & 92.47$\pm$0.83   & 68.24$\pm$1.39                      & 68.07$\pm$1.34            & 40.77$\pm$2.18                      & 40.53$\pm$2.27 \\ 
SPOS+Random                & 3499.51   & 89.39$\pm$1.14                  & 92.46$\pm$0.63   & 68.42$\pm$1.28                      & 68.75$\pm$1.39            & 41.22$\pm$1.32                      & 41.36$\pm$1.22 \\
SPOS+Loss                  & 3505.29   & 90.45$\pm$0.9                   & 93.14$\pm$0.77   & 70.63$\pm$0.94                      & 70.7$\pm$1.03             & 44.28$\pm$0.88                      & 44.29$\pm$0.75 \\
SPOS+Entropy               & 3521.25   & 89.91$\pm$0.96                  & 92.95$\pm$0.71   & 70.54$\pm$0.94                      & 70.66$\pm$0.68            & 43.35$\pm$0.21                      & 43.70$\pm$0.67 \\
SPOS+Gradient              & \textbf{3444.61}   & 90.38$\pm$0.25                   & 93.21$\pm$0.43   & 70.67$\pm$0.67           & 70.29$\pm$0.14   & 43.61$\pm$0.50                      & 43.85$\pm$1.06 \\ 
SPOS+Prediction            & 3519.36   & 90.25$\pm$0.72                  & 92.97$\pm$0.55   & 70.24$\pm$0.3                       & 70.05$\pm$0.53            & 43.22$\pm$0.25                      & 43.54$\pm$1.05 \\
SPOS+Mixture              & 3455.48   & \textbf{90.76$\pm$0.21}        & \textbf{93.79$\pm$0.17}   & \textbf{70.72$\pm$0.21}              & \textbf{71.08$\pm$0.45}            & \textbf{44.63$\pm$0.51}             & \textbf{44.73$\pm$0.56} \\
\bottomrule
\end{tabular}
\caption{Results from three independent runs on NAS-bench-201 search space. Results of RSPS~\cite{li2020random}, DARTS~\cite{liu2018darts}, GDAS~\cite{dong2019searching}, SETN~\cite{dong2019one} and ENAS~\cite{pham2018efficient} are also listed for comparison. The experiments are conducted in three datasets with $50\%$ sampling ratio for different importance metrics.}
\label{tab:nas201_results}
\end{table*}

\section{Experiments}

We first apply the ASP to accelerate AutoML tasks and mainly explore the acceleration effect of data selection on NAS and HPO tasks.
Then, to demonstrate the generalization and robustness of the ASP framework, we also conduct experiments on regular model training. 
Finally, we devise 36 hyper-parameters trials and compare the correlation coefficients between different sampling ratios and metrics. Due to the page limitation, we only present the results of a few datasets on the ResNet-18 model~\cite{he2016deep}. Some more implementation details, experimental results,  ablation studies, and visualization are provided in \textbf{the appendix.}

\begin{table}
\centering
\scriptsize
\begin{tabular}{l|c|c|c}
\toprule
NAS algorithm    & Test Acc.(\%) & Data Ratio &  Cost \\ \hline
 \multicolumn{4}{c}{\textbf{CIFAR-10 (Base:50K Training data)}} \\ \hline
PC-DARTS         & 97.43 $\pm$ 0.07      & 100\%      & 0.1              \\
R-DARTS          & 97.05 $\pm$ 0.21      &  100\%        & 1.6              \\
DARTS-PT         & 97.39 $\pm$ 0.08      & 100\%      & 0.8              \\
DARTS            & 97.00 $\pm$ 0.14      & 100\%      & 0.4              \\
DARTS-Accelerating        & 97.06 $\pm$ 0.00       & 10\%      & 0.03            \\ \hline
DARTS-Random     & 96.83 $\pm$ 0.11        & 10\%      & 0.04            \\
DARTS-Loss       & 96.86 $\pm$ 0.15        & 10\%      & 0.04            \\
DARTS-Entropy    & 97.08 $\pm$ 0.06        & 10\%      & 0.04            \\
DARTS-Gradient       & 96.95 $\pm$ 0.25        & 10\%      & 0.04            \\
DARTS-Prediction & 96.94 $\pm$ 0.05        & 10\%      & 0.04            \\ 
DARTS-Mixture    & \textbf{97.31 $\pm$ 0.12}        & 10\%      & 0.04            \\  \hline
 \multicolumn{4}{c}{\textbf{ImageNet (Base:128K Training data)}} \\ \hline
DARTS            & 73.30 $\pm$ 0.00      & 100\%      & -              \\
DARTS-Accelerating        & 75.40 $\pm$ 0.00       & 10\%      & 0.32            \\ \hline
DARTS-Random     & 74.28 $\pm$ 0.21        & 10\%      & 0.33            \\
DARTS-loss       & 75.36 $\pm$ 0.08        & 10\%      & 0.33            \\
DARTS-Entropy    & 75.14 $\pm$ 0.16        & 10\%      & 0.33            \\
DARTS-Gradient       & 74.86 $\pm$ 0.14        & 10\%      & 0.33            \\
DARTS-Prediction & 74.91 $\pm$ 0.11        & 10\%      & 0.33            \\
DARTS-Mixture    & \textbf{75.61 $\pm$ 0.07}        & 10\%      & 0.33            \\ 
\bottomrule
\end{tabular}
\caption{Evaluation of various differentiable search algorithms on DARTS search space. Results of PC-DARTS~\cite{xu2019pc}, R-DARTS~\cite{arber2020understanding}, DARTS-PT~\cite{wang2021rethinking} and DARTS~\cite{liu2018darts} are listed for comparison. Search cost is GPU days and the experiments are conducted with a 10\% sampling ratio for different
importance metrics.}
\label{tab:darts_results}
\end{table}

\subsection{Proxy with Neural Architecture Search}

We first conduct an experiment on NAS-Bench-201~\cite{dong2020bench} search space, which records the real performance of all sub-models in the search space. We utilize SPOS~\cite{guo2020single} algorithm as our baseline and combine it with different sampling metrics to accelerate the supernet training. After training the supernet, we directly inherit the weights from the pre-trained supernet and evaluate each sub-model. Finally, we query the real performance of the sub-model with the highest evaluation accuracy from NAS-Bench-201.
Additionally, we also conducted an experiment by combining the ASP framework with DARTS to show the effectiveness of ASP. All experimental results are averaged by three independent runs with different random seeds.
The comparison of our searched models on NAS-Bench-201 is presented in Tab. \ref{tab:nas201_results}. We set the sampling ratio as 0.5 and apply the different importance metrics to select proxy data to accelerate supernet training. 
SPOS performs well on three datasets with a search cost of only 7128s. But we can further observe that by combining SPOS with the mixture metric ASP, SPOS+Mixture can outperform the SPOS with higher test accuracy and less search cost. 
Furthermore, the mixture of multi-metric can comprehensively surpass all other data selection methods using a single metric at a similar search cost.

As shown in Tab. \ref{tab:darts_results}, we report the search cost and performance comparison on the DARTS search space. 
It can be observed: Incorporated with the ASP framework, using only 10\% of the training data, DARTS-Mixture can find a better sub-model while reducing the search time by 90\%. DARTS-Accelerating~\cite{na2021accelerating} also adopts 10\% as proxy data to complete the search process. However, the proxy data selected by the DARTS-Accelerating method is constant and is selected before model training. It also requires an additional pre-trained model to select data in advance, whose time is not included in the reported search cost.

\subsection{Proxy with Hyper Parameter Optimization}
We focus on the HPO algorithms with an early-exist mechanism for efficient search. Hyperband~\cite{li2017hyperband}, BOHB~\cite{falkner2018bohb}, and HyperTune~\cite{li2022hyper} are advances of the Sequential Halving algorithm. These algorithms have a good balance between search efficiency and performance, and we also perform asynchronous parallel acceleration~\cite{meister2020maggy} on the above algorithms to further reduce run time. We combined the proxy data with HPO algorithms and conducted an 8-way asynchronous parallel hyper-parameter search with the ResNet-18 model on the CIFAR-10 dataset.
The proxy data was selected by a mixture of metric and a dynamic sampling ratio. All the experimental results are averaged by three independent runs with different random seeds.
The performance comparison of different HPO algorithms under different resource budgets is shown in Tab. \ref{tab:hpo_results}. 
We can find that after integrating with ASP, even if only 5\% or 30\% of the data is used, the HPO algorithm in Tab. \ref{tab:hpo_results} can find the hyper-parameters which are comparable to those found by using 100\% data.

Tab. \ref{tab:hpo_ablation} shows the results of using different metrics in ASP with a 10\% sampling ratio. 
We adopt HyperTune, a state-of-the-art algorithm, as a strong baseline and combine it with ASP for the ablation study. 
From the result of Tab. \ref{tab:hpo_ablation}, we can draw a similar conclusion as in NAS: the mixture of multi-metric can comprehensively surpass all other data selection methods using a single metric, 96.92 for CIFAR-10 and 79.39 for CIFAR-100 dataset.

\begin{table}[!t]
	\centering
 \scriptsize
	\begin{tabular}{l|ccc}
		\toprule
		\diagbox [width=5em,trim=l] {Algorithm}{Ratio} & 100\% & 30\% &  5\% \\
		\hline
		Random & 94.02$\pm$0.78 & 94.55$\pm$0.41 &  93.89$\pm$0.57   \\
		HyperBand & 94.65$\pm$0.44 & 95.78$\pm$0.33 &  95.56$\pm$0.58   \\
		BOHB & 95.01$\pm$0.51 & 96.25$\pm$0.25  & 95.72$\pm$0.15   \\
		HyperTune & \textbf{95.21$\pm$0.35} & \textbf{96.74$\pm$0.17}  & \textbf{95.98$\pm$0.28}   \\
		\bottomrule
	\end{tabular}\vspace{0cm}
	\caption{Comparison of HPO algorithms with ResNet-18 model on CIFAR10 dataset. All the proxy acceleration experiments are conducted with dynamic sampling ratio and mixture metric.}
	\label{tab:hpo_results}
\end{table}

\begin{table}
	\centering
        \scriptsize
	\begin{tabular}{l|ccc}
		\toprule
		\diagbox [width=5em,trim=l] {Algorithm}{Dataset} & CIFAR-10 & CIFAR-100 &Cost \\
		\hline
		Random & 94.02$\pm$0.78 & 77.46$\pm$0.54 &  2.0   \\
		HyperBand & 94.65$\pm$0.44 & 77.90$\pm$0.45 &  2.0   \\
		BOHB & 95.01$\pm$0.51 & 78.43$\pm$0.31  & 2.0   \\
		HyperTune(HT.) & 95.21$\pm$0.35 & 78.4$\pm$0.28  & 2.0   \\
		\hline
		HT.-Random & 95.47$\pm$0.52 & 77.92$\pm$0.31 &  0.2   \\
		HT.-Loss & 95.57$\pm$0.35 & 78.13$\pm$0.24 &  0.2   \\
		HT.-Entropy & 95.16$\pm$0.25 & 78.51$\pm$0.25  & 0.2   \\
		HT.-Gradient & 95.26$\pm$0.19 & 77.85$\pm$0.52  & 0.2   \\
		HT.-Prediction & 94.94$\pm$0.51 & 79.05$\pm$0.24  & 0.2   \\
		HT.-Mixture & \textbf{96.92$\pm$0.25} & \textbf{79.39$\pm$0.11}  & 0.2   \\
		\bottomrule
	\end{tabular}\vspace{0cm}
	\caption{Ablation study of different importance metrics for HPO algorithms with dynamic sampling ratio. Search cost is GPU days and all the experiments are adopted with the ResNet-18 model on the CIFAR-10 and CIFAR-100 datasets.}
	\label{tab:hpo_ablation}
\end{table}

\subsection{Performance of proxy data selection on normal training}
\begin{figure*}[!t]
\begin{center}
\subfigure[Data Ratio 10\%]{
\includegraphics[width=.23\linewidth]{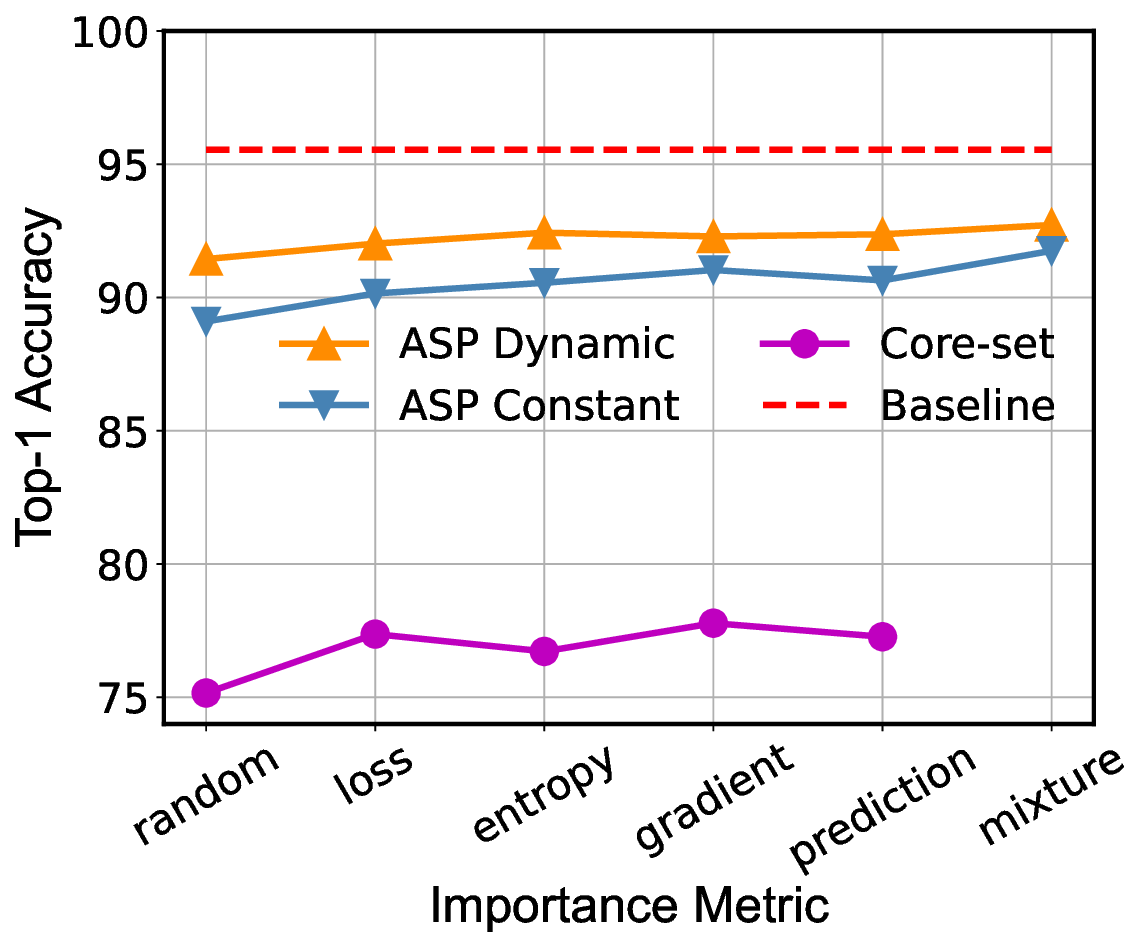}
\label{f_a}
}
\subfigure[Data Ratio 30\%]{
\includegraphics[width=.23\linewidth]{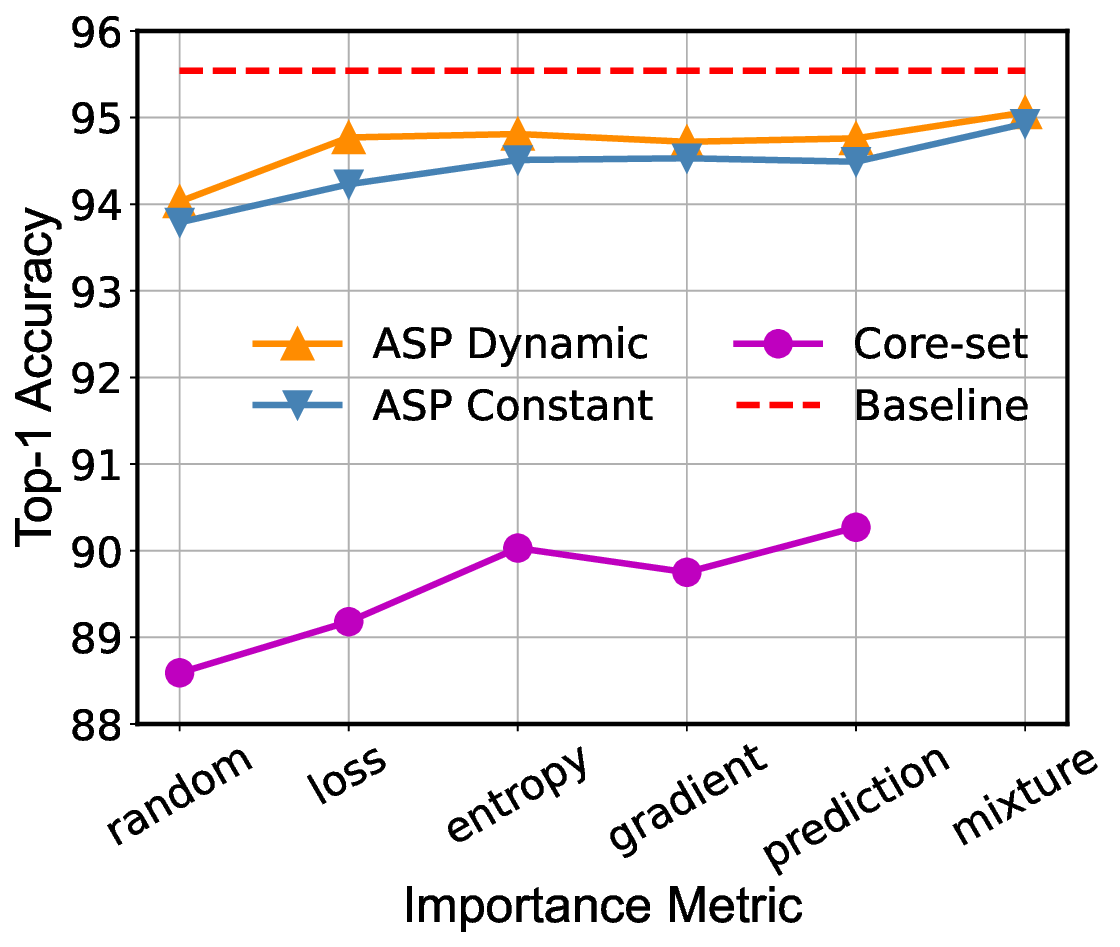}
}
\subfigure[Data Ratio 50\%]{
\includegraphics[width=.23\linewidth]{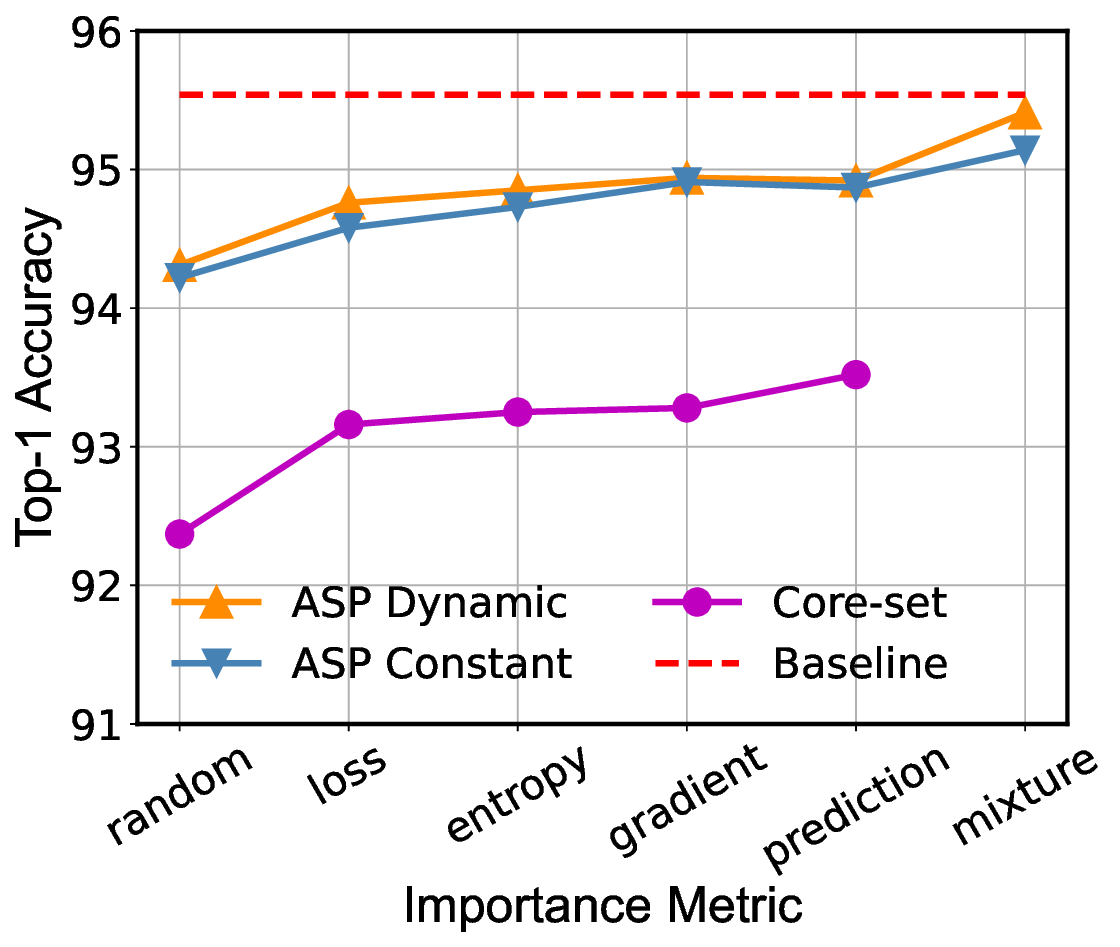}
\label{f_c}
}
\subfigure[Data Ratio 70\%]{
\includegraphics[width=.23\linewidth]{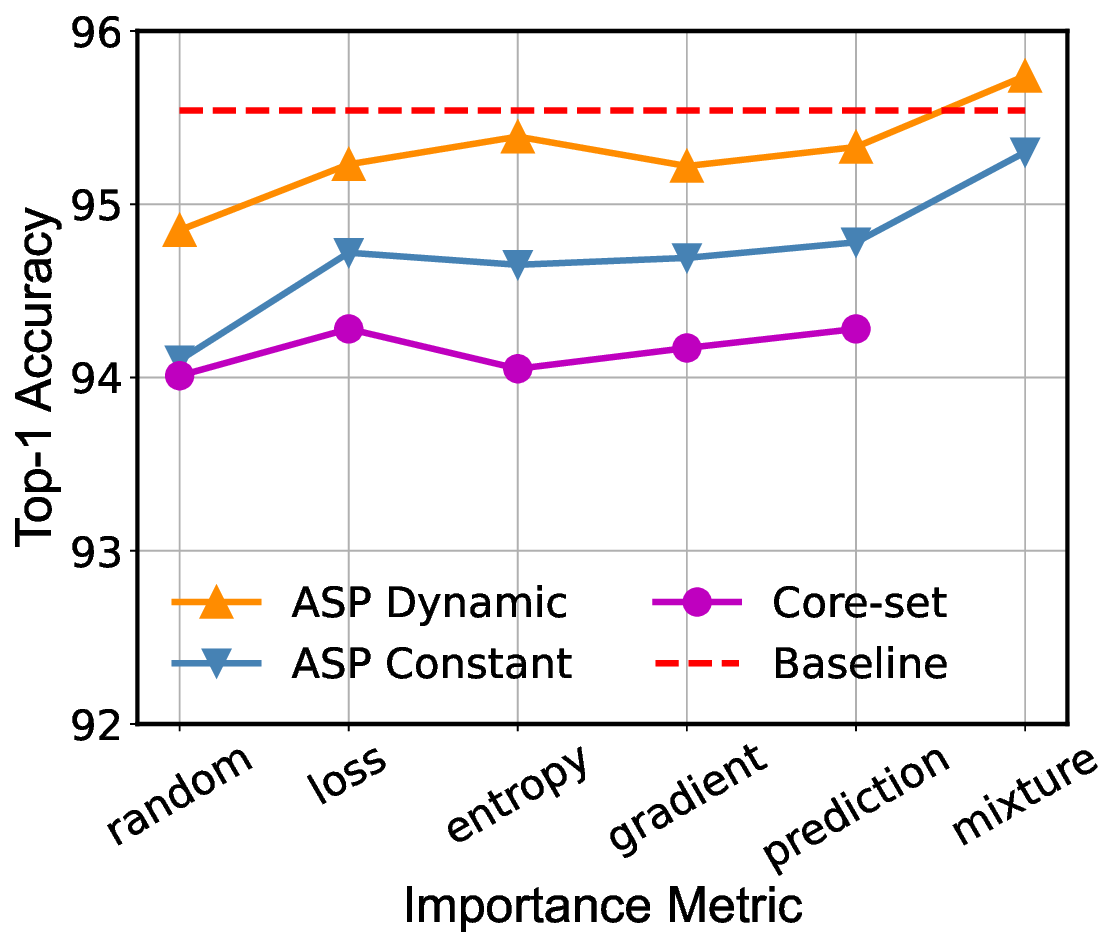}
}
\caption{Test performance (Top-1 Accuracy(\%)) of ResNet-18 on CIFAR-10 under various data sampling ratios, data selection methods, and importance metrics.}
\label{fig:perf_proxy}
\end{center}
\end{figure*}

We train the ResNet-18 model on the proxy subset of the CIFAR-10 training dataset with different sampling ratios. To compare to the core-set selection, we train a ResNet-18 model with full data in advance and then select a certain ratio of proxy data as the training data with the corresponding importance metric.
The comparison of different selection configures is shown in Fig. \ref{fig:perf_proxy}. We define four sampling ratios, e.g. 10\%, 30\%, 50\%, 70\%, to illustrate the robustness and generalization of the ASP method. The ASP dynamic and constant mean how the sampling ratio is obtained. The ASP dynamic ratio means that in the model training stage, the sample ratio gradually decreases from 100\% to 0\% but the expected value of the dynamic ratio for all epochs is equivalent to the constant ratio.
We find that the model trained with the dynamic sampling ratio has consistently better performance than the constant sampling ratio under different settings. This is mainly because the early model needs enormous training data for fast convergence and the final model only requires several hard samples to refine the model parameters.
The results show that core-set selection seriously damages the model generalization to the test data and is not suitable for training the finally submitted model.
Compared to random metrics, a simple but effective baseline~\cite{wang2019e2}, the other proxy metrics based on training results, especially for the mixture metric, can select more important proxy data and further reduce the performance drop.
It is worth mentioning that when the sampling ratio is further increased, the model performance with proxy data will be closer and even surpass the original model performance with full data, i.e. Baseline. It exactly validates an assumption: the dataset has some redundancy data. 


\subsection{Correlation Analysis}
To verify that our proposed ASP method can better maintain the performance ranking between the model trained with proxy data and the model trained with full data, we designed a simple hyper-parameter search space, which contains a total of $2\times3\times3\times2=36$ sets of hyper-parameter configures as follows: LabelSmooth (False, True), LearningRate (0.05, 0.1, 0.2), FlipProb (0.25, 0.5, 0.75) and SGDMomentum (0.8, 0.9). 
We adopt the ResNet-20 model in this experiment, a light architecture~\cite{he2016deep}, and can obtain fast training and evaluation results under the CIFAR-10 dataset.
We train the model with each hyper-parameter under different sampling ratios and finally count the correlation coefficients between them, where we adopt a dynamic sampling ratio in this experiment.

The result of the correlation experiment is shown in Tab. \ref{tab:coeficient}, which includes three types of correlation coefficients under five sampling ratios, e.g. 10\%, 30\%, 50\%, 70\%, and 85\%. We find that with the training data increasing, the correction coefficient of ranking between models gradually saturated and about 30\% of the data can have a high correction. This experimental phenomenon also reveals that proxy data can greatly improve search efficiency because performance rank between models is more important than absolute performance during the search stage, where a post-process of fine-tuning is required.
What's more, the mixture metric is significantly better than others with about 0.1 higher among all the selection metrics, demonstrating that the mixture metric can make model training more stable and approach the training trend with full training data.


\begin{table*}[!t]
    \centering
    \scriptsize
    \begin{tabular}{l|ccc|ccc|ccc|ccc|ccc}
    \toprule
    \multirow{2}{*}{\diagbox [width=5em,trim=l] {Proxy}{Ratio}} & \multicolumn{3}{c|}{$r=0.1$} & \multicolumn{3}{c|}{$r=0.3$} & \multicolumn{3}{c|}{$r=0.5$} & \multicolumn{3}{c|}{$r=0.7$}  & \multicolumn{3}{c}{$r=0.85$} \\ \cline{2-16} 
                                                                             & $\tau$      & $\rho$      & $s$      & $\tau$      & $\rho$      & $s$      & $\tau$      & $\rho$      & $s$      & $\tau$      & $\rho$      & $s$     & $\tau$      & $\rho$      & $s$\\
                                                                             \hline
    Random                                                                   & 0.52   & 0.76   & 0.69   & 0.60   & 0.85   & 0.80   & 0.65   & 0.89   & 0.83   & 0.69   & 0.85   & 0.87  & 0.64   & 0.85   & 0.84\\
    Loss                                                                     & 0.54   & 0.77   & 0.74   & 0.70   & 0.89   & 0.87   & 0.72   & 0.86   & 0.88   & 0.72   & 0.87   & 0.88  & 0.74   & 0.91   & 0.89\\
    Entropy                                                                  & 0.57   & 0.83   & 0.77   & 0.69   & 0.84   & 0.87   & 0.69   & \textbf{0.90}   & 0.87   & 0.73   & 0.90   & 0.90  & 0.73   & 0.91   & 0.90\\
    Gradient                                                                 & 0.54   & 0.75   & 0.72   & 0.71   & 0.86   & 0.89   & 0.67   & 0.85   & 0.84   & 0.69   & 0.87   & 0.88  & 0.68   & 0.88   & 0.86\\
    Prediction                                                               & 0.62   & 0.84   & 0.82   & 0.67   & 0.87   & 0.85   & 0.70   & 0.86   & 0.88   & 0.71   & 0.88   & 0.87  & 0.69   & 0.89   & 0.85\\
    Mixture                                                        & \textbf{0.67} & \textbf{0.85}   & \textbf{0.83}   & \textbf{0.73}   & \textbf{0.89}   & \textbf{0.90}   & \textbf{0.76}   & 0.89   & \textbf{0.91}   & \textbf{0.75}   & \textbf{0.91}   & \textbf{0.92}  & \textbf{0.77}   & \textbf{0.92}   & \textbf{0.90}\\
    \bottomrule
    \end{tabular}
    \caption{Experiment results of correlation coefficients under different sampling ratios $r$ and importance metrics. We record three popular correlation coefficient, Kendall tau $\tau$~\cite{kendall1938new}, Pearson $\rho$~\cite{benesty2009pearson} and Spearman $s$~\cite{spearman1961proof}.}
    \label{tab:coeficient}
\end{table*}

\section{Conclusions}
In this paper, we jointly optimize the data selection and AutoML tasks, and re-formulate this problem as a tri-level optimization. 
To solve this problem, we propose a novel and simple framework called ASP for proxy data selection to accelerate AutoML, determining whether a sample participates in model training in a dynamic manner.
We summarize and explore several common data importance metrics as well as a mixture metric, which can be flexibly inserted into the ASP framework as alternative options.
The results demonstrate that our ASP framework can minimize the performance degradation caused by the reduction of training data and the intensive ablations can demonstrate the reasonability of our design choices in ASP. 
In addition, the selected subset has a higher correlation with the full data in terms of final model performance. ASP largely reduces the search time, which has an important significance to the AutoML tasks.

\paragraph{Limitations} We mainly discuss the effect of the ASP method on classification tasks and some importance metrics used in the framework are also dependent on this task. However, the classification task is only a part of deep learning. In the future, we will verify ASP in more scenarios, e.g. NLP or speech fields. At the same time, the dynamic sampling ratio provided in this paper is prior, and an adjusted sampling ratio according to the state of the model may provide better selection in the future. The prior probability distribution for each metric is manually chosen, which will also be interesting to explore in the future.

\clearpage

{\small
\bibliography{references}

\begin{thebibliography}{61}
\providecommand{\natexlab}[1]{#1}
\providecommand{\url}[1]{\texttt{#1}}
\expandafter\ifx\csname urlstyle\endcsname\relax
  \providecommand{\doi}[1]{doi: #1}\else
  \providecommand{\doi}{doi: \begingroup \urlstyle{rm}\Url}\fi

\bibitem[Z{\"o}ller and Huber(2021)]{zoller2021benchmark}
Marc-Andr{\'e} Z{\"o}ller and Marco~F Huber.
\newblock Benchmark and survey of automated machine learning frameworks.
\newblock \emph{Journal of artificial intelligence research}, 70:\penalty0
  409--472, 2021.

\bibitem[Bender et~al.(2018)Bender, Kindermans, Zoph, Vasudevan, and
  Le]{bender2018understanding}
Gabriel Bender, Pieter-Jan Kindermans, Barret Zoph, Vijay Vasudevan, and Quoc
  Le.
\newblock Understanding and simplifying one-shot architecture search.
\newblock In \emph{International conference on machine learning}, pages
  550--559. PMLR, 2018.

\bibitem[Guo et~al.(2020)Guo, Zhang, Mu, Heng, Liu, Wei, and
  Sun]{guo2020single}
Zichao Guo, Xiangyu Zhang, Haoyuan Mu, Wen Heng, Zechun Liu, Yichen Wei, and
  Jian Sun.
\newblock Single path one-shot neural architecture search with uniform
  sampling.
\newblock In \emph{European conference on computer vision}, pages 544--560.
  Springer, 2020.

\bibitem[Liu et~al.(2018)Liu, Simonyan, and Yang]{liu2018darts}
Hanxiao Liu, Karen Simonyan, and Yiming Yang.
\newblock Darts: Differentiable architecture search.
\newblock \emph{arXiv preprint arXiv:1806.09055}, 2018.

\bibitem[Wu et~al.(2019)Wu, Dai, Zhang, Wang, Sun, Wu, Tian, Vajda, Jia, and
  Keutzer]{wu2019fbnet}
Bichen Wu, Xiaoliang Dai, Peizhao Zhang, Yanghan Wang, Fei Sun, Yiming Wu,
  Yuandong Tian, Peter Vajda, Yangqing Jia, and Kurt Keutzer.
\newblock Fbnet: Hardware-aware efficient convnet design via differentiable
  neural architecture search.
\newblock In \emph{Proceedings of the IEEE/CVF Conference on Computer Vision
  and Pattern Recognition}, pages 10734--10742, 2019.

\bibitem[Falkner et~al.(2018)Falkner, Klein, and Hutter]{falkner2018bohb}
Stefan Falkner, Aaron Klein, and Frank Hutter.
\newblock Bohb: Robust and efficient hyperparameter optimization at scale.
\newblock In \emph{International Conference on Machine Learning}, pages
  1437--1446. PMLR, 2018.

\bibitem[Bergstra et~al.(2011)Bergstra, Bardenet, Bengio, and
  K{\'e}gl]{bergstra2011algorithms}
James Bergstra, R{\'e}mi Bardenet, Yoshua Bengio, and Bal{\'a}zs K{\'e}gl.
\newblock Algorithms for hyper-parameter optimization.
\newblock \emph{Advances in neural information processing systems}, 24, 2011.

\bibitem[Nickson et~al.(2014)Nickson, Osborne, Reece, and
  Roberts]{nickson2014automated}
Thomas Nickson, Michael~A Osborne, Steven Reece, and Stephen~J Roberts.
\newblock Automated machine learning on big data using stochastic algorithm
  tuning.
\newblock \emph{arXiv preprint arXiv:1407.7969}, 2014.

\bibitem[Krueger et~al.(2015)Krueger, Panknin, and Braun]{krueger2015fast}
Tammo Krueger, Danny Panknin, and Mikio~L Braun.
\newblock Fast cross-validation via sequential testing.
\newblock \emph{J. Mach. Learn. Res.}, 16\penalty0 (1):\penalty0 1103--1155,
  2015.

\bibitem[Park(2019)]{park2019data}
Minje Park.
\newblock Data proxy generation for fast and efficient neural architecture
  search.
\newblock \emph{arXiv preprint arXiv:1911.09322}, 2019.

\bibitem[Katharopoulos and Fleuret(2018)]{katharopoulos2018not}
Angelos Katharopoulos and Fran{\c{c}}ois Fleuret.
\newblock Not all samples are created equal: Deep learning with importance
  sampling.
\newblock In \emph{International conference on machine learning}, pages
  2525--2534. PMLR, 2018.

\bibitem[Zhang et~al.(2019)Zhang, Yu, and Dhillon]{zhang2019autoassist}
Jiong Zhang, Hsiang-Fu Yu, and Inderjit~S Dhillon.
\newblock Autoassist: A framework to accelerate training of deep neural
  networks.
\newblock \emph{Advances in Neural Information Processing Systems}, 32, 2019.

\bibitem[Killamsetty et~al.(2022)Killamsetty, Abhishek, Evfimievski, Popa,
  Ramakrishnan, Iyer, et~al.]{killamsetty2022automata}
Krishnateja Killamsetty, Guttu~Sai Abhishek, Alexandre~V Evfimievski, Lucian
  Popa, Ganesh Ramakrishnan, Rishabh Iyer, et~al.
\newblock Automata: Gradient based data subset selection for compute-efficient
  hyper-parameter tuning.
\newblock \emph{arXiv preprint arXiv:2203.08212}, 2022.

\bibitem[Na et~al.(2021)Na, Mok, Choe, and Yoon]{na2021accelerating}
Byunggook Na, Jisoo Mok, Hyeokjun Choe, and Sungroh Yoon.
\newblock Accelerating neural architecture search via proxy data.
\newblock \emph{arXiv preprint arXiv:2106.04784}, 2021.

\bibitem[Hutter et~al.(2019)Hutter, Kotthoff, and
  Vanschoren]{hutter2019automated}
Frank Hutter, Lars Kotthoff, and Joaquin Vanschoren.
\newblock \emph{Automated machine learning: methods, systems, challenges}.
\newblock Springer Nature, 2019.

\bibitem[Pham et~al.(2018)Pham, Guan, Zoph, Le, and Dean]{pham2018efficient}
Hieu Pham, Melody Guan, Barret Zoph, Quoc Le, and Jeff Dean.
\newblock Efficient neural architecture search via parameters sharing.
\newblock In \emph{International conference on machine learning}, pages
  4095--4104. PMLR, 2018.

\bibitem[Zoph et~al.(2018)Zoph, Vasudevan, Shlens, and Le]{zoph2018learning}
Barret Zoph, Vijay Vasudevan, Jonathon Shlens, and Quoc~V Le.
\newblock Learning transferable architectures for scalable image recognition.
\newblock In \emph{Proceedings of the IEEE conference on computer vision and
  pattern recognition}, pages 8697--8710, 2018.

\bibitem[Claesen et~al.(2014)Claesen, Simm, Popovic, Moreau, and
  De~Moor]{claesen2014easy}
Marc Claesen, Jaak Simm, Dusan Popovic, Yves Moreau, and Bart De~Moor.
\newblock Easy hyperparameter search using optunity.
\newblock \emph{arXiv preprint arXiv:1412.1114}, 2014.

\bibitem[Pinto et~al.(2009)Pinto, Doukhan, DiCarlo, and Cox]{pinto2009high}
Nicolas Pinto, David Doukhan, James~J DiCarlo, and David~D Cox.
\newblock A high-throughput screening approach to discovering good forms of
  biologically inspired visual representation.
\newblock \emph{PLoS computational biology}, 5\penalty0 (11):\penalty0
  e1000579, 2009.

\bibitem[Jamieson and Talwalkar(2016)]{jamieson2016non}
Kevin Jamieson and Ameet Talwalkar.
\newblock Non-stochastic best arm identification and hyperparameter
  optimization.
\newblock In \emph{Artificial intelligence and statistics}, pages 240--248.
  PMLR, 2016.

\bibitem[Li et~al.(2020)Li, Jamieson, Rostamizadeh, Gonina, Ben-Tzur, Hardt,
  Recht, and Talwalkar]{li2020system}
Liam Li, Kevin Jamieson, Afshin Rostamizadeh, Ekaterina Gonina, Jonathan
  Ben-Tzur, Moritz Hardt, Benjamin Recht, and Ameet Talwalkar.
\newblock A system for massively parallel hyperparameter tuning.
\newblock \emph{Proceedings of Machine Learning and Systems}, 2:\penalty0
  230--246, 2020.

\bibitem[Shleifer and Prokop(2019)]{shleifer2019using}
Sam Shleifer and Eric Prokop.
\newblock Using small proxy datasets to accelerate hyperparameter search.
\newblock \emph{arXiv preprint arXiv:1906.04887}, 2019.

\bibitem[Killamsetty et~al.(2021{\natexlab{a}})Killamsetty, Durga,
  Ramakrishnan, De, and Iyer]{killamsetty2021grad}
Krishnateja Killamsetty, S~Durga, Ganesh Ramakrishnan, Abir De, and Rishabh
  Iyer.
\newblock Grad-match: Gradient matching based data subset selection for
  efficient deep model training.
\newblock In \emph{International Conference on Machine Learning}, pages
  5464--5474. PMLR, 2021{\natexlab{a}}.

\bibitem[Killamsetty et~al.(2021{\natexlab{b}})Killamsetty, Sivasubramanian,
  Ramakrishnan, and Iyer]{killamsetty2021glister}
Krishnateja Killamsetty, Durga Sivasubramanian, Ganesh Ramakrishnan, and
  Rishabh Iyer.
\newblock Glister: Generalization based data subset selection for efficient and
  robust learning.
\newblock In \emph{Proceedings of the AAAI Conference on Artificial
  Intelligence}, volume~35, pages 8110--8118, 2021{\natexlab{b}}.

\bibitem[Mirzasoleiman et~al.(2020)Mirzasoleiman, Bilmes, and
  Leskovec]{mirzasoleiman2020coresets}
Baharan Mirzasoleiman, Jeff Bilmes, and Jure Leskovec.
\newblock Coresets for data-efficient training of machine learning models.
\newblock In \emph{International Conference on Machine Learning}, pages
  6950--6960. PMLR, 2020.

\bibitem[Johnson and Guestrin(2018)]{johnson2018training}
Tyler~B Johnson and Carlos Guestrin.
\newblock Training deep models faster with robust, approximate importance
  sampling.
\newblock \emph{Advances in Neural Information Processing Systems}, 31, 2018.

\bibitem[Jiang et~al.(2019)Jiang, Wong, Zhou, Andersen, Dean, Ganger, Joshi,
  Kaminksy, Kozuch, Lipton, et~al.]{jiang2019accelerating}
Angela~H Jiang, Daniel L-K Wong, Giulio Zhou, David~G Andersen, Jeffrey Dean,
  Gregory~R Ganger, Gauri Joshi, Michael Kaminksy, Michael Kozuch, Zachary~C
  Lipton, et~al.
\newblock Accelerating deep learning by focusing on the biggest losers.
\newblock \emph{arXiv preprint arXiv:1910.00762}, 2019.

\bibitem[Wang et~al.(2018)Wang, Zhu, Torralba, and Efros]{wang2018dataset}
Tongzhou Wang, Jun-Yan Zhu, Antonio Torralba, and Alexei~A Efros.
\newblock Dataset distillation.
\newblock \emph{arXiv preprint arXiv:1811.10959}, 2018.

\bibitem[Zhao et~al.(2021)Zhao, Mopuri, and Bilen]{zhao2021dataset}
Bo~Zhao, Konda~Reddy Mopuri, and Hakan Bilen.
\newblock Dataset condensation with gradient matching.
\newblock \emph{ICLR}, 1\penalty0 (2):\penalty0 3, 2021.

\bibitem[Coleman et~al.(2019)Coleman, Yeh, Mussmann, Mirzasoleiman, Bailis,
  Liang, Leskovec, and Zaharia]{coleman2019selection}
Cody Coleman, Christopher Yeh, Stephen Mussmann, Baharan Mirzasoleiman, Peter
  Bailis, Percy Liang, Jure Leskovec, and Matei Zaharia.
\newblock Selection via proxy: Efficient data selection for deep learning.
\newblock \emph{arXiv preprint arXiv:1906.11829}, 2019.

\bibitem[Zhou et~al.(2020)Zhou, Zhou, Zhang, Loy, Yi, Zhang, and
  Ouyang]{zhou2020econas}
Dongzhan Zhou, Xinchi Zhou, Wenwei Zhang, Chen~Change Loy, Shuai Yi, Xuesen
  Zhang, and Wanli Ouyang.
\newblock Econas: Finding proxies for economical neural architecture search.
\newblock In \emph{Proceedings of the IEEE/CVF Conference on computer vision
  and pattern recognition}, pages 11396--11404, 2020.

\bibitem[Colson et~al.(2007)Colson, Marcotte, and Savard]{colson2007overview}
Beno{\^\i}t Colson, Patrice Marcotte, and Gilles Savard.
\newblock An overview of bilevel optimization.
\newblock \emph{Annals of operations research}, 153\penalty0 (1):\penalty0
  235--256, 2007.

\bibitem[Franceschi et~al.(2018)Franceschi, Frasconi, Salzo, Grazzi, and
  Pontil]{franceschi2018bilevel}
Luca Franceschi, Paolo Frasconi, Saverio Salzo, Riccardo Grazzi, and
  Massimiliano Pontil.
\newblock Bilevel programming for hyperparameter optimization and
  meta-learning.
\newblock In \emph{International Conference on Machine Learning}, pages
  1568--1577. PMLR, 2018.

\bibitem[Moser et~al.(2022)Moser, Raue, Hees, and Dengel]{moser2022less}
Brian Moser, Federico Raue, J{\"o}rn Hees, and Andreas Dengel.
\newblock Less is more: Proxy datasets in nas approaches.
\newblock In \emph{Proceedings of the IEEE/CVF Conference on Computer Vision
  and Pattern Recognition}, pages 1953--1961, 2022.

\bibitem[Aljundi et~al.(2019)Aljundi, Lin, Goujaud, and
  Bengio]{aljundi2019gradient}
Rahaf Aljundi, Min Lin, Baptiste Goujaud, and Yoshua Bengio.
\newblock Gradient based sample selection for online continual learning.
\newblock \emph{Advances in neural information processing systems}, 32, 2019.

\bibitem[Eric et~al.(2021)Eric, Diego, Paul, and Kevin]{eric2021important}
Arazo Eric, Ortego Diego, Albert Paul, and McGuinness Kevin.
\newblock How important is importance sampling for deep budgeted training?
\newblock 2021.

\bibitem[Settles(2009)]{settles2009active}
Burr Settles.
\newblock Active learning literature survey.
\newblock 2009.

\bibitem[Toneva et~al.(2018)Toneva, Sordoni, des Combes, Trischler, Bengio, and
  Gordon]{toneva2018empirical}
Mariya Toneva, Alessandro Sordoni, Remi~Tachet des Combes, Adam Trischler,
  Yoshua Bengio, and Geoffrey~J Gordon.
\newblock An empirical study of example forgetting during deep neural network
  learning.
\newblock In \emph{International Conference on Learning Representations}, 2018.

\bibitem[Zhang et~al.(2021)Zhang, Chen, Chen, and Wang]{zhang2021efficient}
Zhenyu Zhang, Xuxi Chen, Tianlong Chen, and Zhangyang Wang.
\newblock Efficient lottery ticket finding: Less data is more.
\newblock In \emph{International Conference on Machine Learning}, pages
  12380--12390. PMLR, 2021.

\bibitem[Li and Talwalkar(2020)]{li2020random}
Liam Li and Ameet Talwalkar.
\newblock Random search and reproducibility for neural architecture search.
\newblock In \emph{Uncertainty in artificial intelligence}, pages 367--377.
  PMLR, 2020.

\bibitem[Dong and Yang(2019{\natexlab{a}})]{dong2019searching}
Xuanyi Dong and Yi~Yang.
\newblock Searching for a robust neural architecture in four gpu hours.
\newblock In \emph{Proceedings of the IEEE/CVF Conference on Computer Vision
  and Pattern Recognition}, pages 1761--1770, 2019{\natexlab{a}}.

\bibitem[Dong and Yang(2019{\natexlab{b}})]{dong2019one}
Xuanyi Dong and Yi~Yang.
\newblock One-shot neural architecture search via self-evaluated template
  network.
\newblock In \emph{Proceedings of the IEEE/CVF International Conference on
  Computer Vision}, pages 3681--3690, 2019{\natexlab{b}}.

\bibitem[He et~al.(2016)He, Zhang, Ren, and Sun]{he2016deep}
Kaiming He, Xiangyu Zhang, Shaoqing Ren, and Jian Sun.
\newblock Deep residual learning for image recognition.
\newblock In \emph{Proceedings of the IEEE conference on computer vision and
  pattern recognition}, pages 770--778, 2016.

\bibitem[Xu et~al.(2019)Xu, Xie, Zhang, Chen, Qi, Tian, and Xiong]{xu2019pc}
Yuhui Xu, Lingxi Xie, Xiaopeng Zhang, Xin Chen, Guo-Jun Qi, Qi~Tian, and
  Hongkai Xiong.
\newblock Pc-darts: Partial channel connections for memory-efficient
  architecture search.
\newblock \emph{arXiv preprint arXiv:1907.05737}, 2019.

\bibitem[Arber~Zela et~al.(2020)Arber~Zela, Saikia, Marrakchi, Brox, and
  Hutter]{arber2020understanding}
Thomas~Elsken Arber~Zela, Tonmoy Saikia, Yassine Marrakchi, Thomas Brox, and
  Frank Hutter.
\newblock Understanding and robustifying differentiable architecture search.
\newblock In \emph{International Conference on Learning Representations},
  volume~2, 2020.

\bibitem[Wang et~al.(2021)Wang, Cheng, Chen, Tang, and
  Hsieh]{wang2021rethinking}
Ruochen Wang, Minhao Cheng, Xiangning Chen, Xiaocheng Tang, and Cho-Jui Hsieh.
\newblock Rethinking architecture selection in differentiable nas.
\newblock In \emph{International Conference on Learning Representation}, 2021.

\bibitem[Dong and Yang(2020)]{dong2020bench}
Xuanyi Dong and Yi~Yang.
\newblock Nas-bench-201: Extending the scope of reproducible neural
  architecture search.
\newblock \emph{arXiv preprint arXiv:2001.00326}, 2020.

\bibitem[Li et~al.(2017)Li, Jamieson, DeSalvo, Rostamizadeh, and
  Talwalkar]{li2017hyperband}
Lisha Li, Kevin Jamieson, Giulia DeSalvo, Afshin Rostamizadeh, and Ameet
  Talwalkar.
\newblock Hyperband: A novel bandit-based approach to hyperparameter
  optimization.
\newblock \emph{The Journal of Machine Learning Research}, 18\penalty0
  (1):\penalty0 6765--6816, 2017.

\bibitem[Li et~al.(2022)Li, Shen, Jiang, Zhang, Li, Liu, Zhang, and
  Cui]{li2022hyper}
Yang Li, Yu~Shen, Huaijun Jiang, Wentao Zhang, Jixiang Li, Ji~Liu, Ce~Zhang,
  and Bin Cui.
\newblock Hyper-tune: Towards efficient hyper-parameter tuning at scale.
\newblock \emph{arXiv preprint arXiv:2201.06834}, 2022.

\bibitem[Meister et~al.(2020)Meister, Sheikholeslami, Payberah, Vlassov, and
  Dowling]{meister2020maggy}
Moritz Meister, Sina Sheikholeslami, Amir~H Payberah, Vladimir Vlassov, and Jim
  Dowling.
\newblock Maggy: Scalable asynchronous parallel hyperparameter search.
\newblock In \emph{Proceedings of the 1st Workshop on Distributed Machine
  Learning}, pages 28--33, 2020.

\bibitem[Wang et~al.(2019)Wang, Jiang, Chen, Xu, Zhao, Lin, and
  Wang]{wang2019e2}
Yue Wang, Ziyu Jiang, Xiaohan Chen, Pengfei Xu, Yang Zhao, Yingyan Lin, and
  Zhangyang Wang.
\newblock E2-train: Training state-of-the-art cnns with over 80\% energy
  savings.
\newblock \emph{Advances in Neural Information Processing Systems}, 32, 2019.

\bibitem[Kendall(1938)]{kendall1938new}
Maurice~G Kendall.
\newblock A new measure of rank correlation.
\newblock \emph{Biometrika}, 30\penalty0 (1/2):\penalty0 81--93, 1938.

\bibitem[Benesty et~al.(2009)Benesty, Chen, Huang, and
  Cohen]{benesty2009pearson}
Jacob Benesty, Jingdong Chen, Yiteng Huang, and Israel Cohen.
\newblock Pearson correlation coefficient.
\newblock In \emph{Noise reduction in speech processing}, pages 1--4. Springer,
  2009.

\bibitem[Spearman(1961)]{spearman1961proof}
Charles Spearman.
\newblock The proof and measurement of association between two things.
\newblock 1961.

\bibitem[Coleman et~al.(2020)Coleman, Yeh, Mussmann, Mirzasoleiman, Bailis,
  Liang, Leskovec, and Zaharia]{coleman2020selection}
C~Coleman, C~Yeh, S~Mussmann, B~Mirzasoleiman, P~Bailis, P~Liang, J~Leskovec,
  and M~Zaharia.
\newblock Selection via proxy: Efficient data selection for deep learning.
\newblock In \emph{International Conference on Learning Representations
  (ICLR)}, 2020.

\bibitem[Krizhevsky et~al.(2009)]{krizhevsky2009learning}
Alex Krizhevsky et~al.
\newblock Learning multiple layers of features from tiny images.
\newblock 2009.

\bibitem[Krizhevsky et~al.(2012)Krizhevsky, Sutskever, and
  Hinton]{krizhevsky2012imagenet}
Alex Krizhevsky, Ilya Sutskever, and Geoffrey~E Hinton.
\newblock Imagenet classification with deep convolutional neural networks.
\newblock \emph{Advances in neural information processing systems}, 25, 2012.

\bibitem[Dong and Yang(2019{\natexlab{c}})]{dong2019bench}
Xuanyi Dong and Yi~Yang.
\newblock Nas-bench-201: Extending the scope of reproducible neural
  architecture search.
\newblock In \emph{International Conference on Learning Representations},
  2019{\natexlab{c}}.

\bibitem[Loshchilov and Hutter(2016)]{loshchilov2016sgdr}
Ilya Loshchilov and Frank Hutter.
\newblock Sgdr: Stochastic gradient descent with warm restarts.
\newblock \emph{arXiv preprint arXiv:1608.03983}, 2016.

\bibitem[DeVries and Taylor(2017)]{devries2017improved}
Terrance DeVries and Graham~W Taylor.
\newblock Improved regularization of convolutional neural networks with cutout.
\newblock \emph{arXiv preprint arXiv:1708.04552}, 2017.

\bibitem[Zhang et~al.(2018)Zhang, Cisse, Dauphin, and
  Lopez-Paz]{zhang2018mixup}
Hongyi Zhang, Moustapha Cisse, Yann~N Dauphin, and David Lopez-Paz.
\newblock mixup: Beyond empirical risk minimization.
\newblock In \emph{International Conference on Learning Representations}, 2018.

\end{thebibliography}
}

\appendix

\section{An Overview}

Due to page limitations, some experimental results and analyses are presented in the supplementary material.
First, we show the overall algorithm framework of ASP, the algorithm details of the sampling ratio scheduler, and more analyses of mixture metrics.
Next, we provide additional experimental results of ASP with various baseline models and datasets.
Then, we visualized the activated states of data during the proxy data selection of ASP, and the direct view of the "hard" or "easy" samples determined by our method.
Finally, we introduce the implementation details of experiments to help readers better reproduce the results.

\section{Algorithm Details}
To better understand the ASP framework, we give a more in-depth elaboration on the details of the ASP framework in this section.

\begin{algorithm}[!h]
\small
\caption{Framework of automatic selection for the proxy dataset.}
\label{alg:alg_autosampling}
\textbf{Input}: Training dataset $\mathcal{D}_{train}$, sample size m, proxy memory $V=\{v_i|i=1,...,n\}$ \\
\textbf{Output}: Proxy data $\mathcal{D}_{proxy}$, proxy memory $V$

\begin{algorithmic}[1] 
\STATE Initialize proxy data $\mathcal{D}_{proxy}$ = $\mathcal{D}_{train}$
\STATE Initialize proxy memory $v_i=0, i=1,...,n$
\FOR{$j=1$ to $n_{epochs}$}
\STATE Model training and importance metric $v_i$ updating on $\mathcal{D}_{proxy}$
\STATE Calculate sample size $m$ in Algorithm~\ref{alg:sample_rate}
\STATE Choose the importance metric.
\STATE Select samples index $I$ with top-m importance value
\STATE Update $\mathcal{D}_{proxy} = \{\mathbf{x_i}, y_i\}_{i\in I}$
\ENDFOR
\end{algorithmic}
\end{algorithm}


As shown in Algorithm \ref{alg:alg_autosampling}, we first initialize $\mathcal{D}_{proxy}$ to be original $\mathcal{D}_{train}$ in the first epoch. 
Next, the parameters of the model will be updated by gradient descent based on the $\mathcal{D}_{proxy}$. 
During the training, the importance metric of each activated sample $v_{i}$ will be updated in proxy memory $V$.
Then, we will calculate the number of samples $m$ that need to be activated in the next epoch of training according to the sampling ratio scheduler, which will be detailed in Algorithm \ref{alg:sample_rate}.
Finally, the sampling probability of each sample in $D_{train}$ will be calculated according to the importance metric, and we will select $m$ samples to form the $\mathcal{D}_{proxy}$ based on this sampling probability.
The above process will be repeated $n_{epochs}$ times.

\subsection{Sampling Ratio Scheduler}
\begin{algorithm}[!h]
\small
\caption{ The calculation of sampling ratio scheduler.}
\label{alg:sample_rate}
\textbf{Input}: Sample scheduler, sampling ratio $r$, current epoch $i$, total epochs $N$, training set $\mathcal{D}_{train}$ \\
\textbf{Output}: The size of proxy data $m$
\begin{algorithmic}[1]
\IF{Static Allocation}
\RETURN $r*\mathcal{D}_{train}$
\ELSE
    \IF{$r \leq 0.5$}
        \STATE $ratios \gets Linspace\left(1, 0.01, N\right)$
        \STATE $r \gets ratios[i] * 2 * r $
    \ELSE
        \STATE $p_{min} \gets 2*r -1$
        \STATE $ratios \gets Linspace\left(1, p_{min}, N\right)$
        \STATE $r \gets ratios[i]$
    \ENDIF
    \RETURN $r*\mathcal{D}_{train}$
\ENDIF
\end{algorithmic}
\end{algorithm}

If ignoring the bottleneck effect of CPU on data preprocessing, the training time is proportional to the size of the training dataset.
The sampling ratio determines the number of samples participating in model training in each epoch, most previous works set a constant sampling ratio ~\cite{na2021accelerating, zhang2021efficient}.
In the ASP framework, we also provide a dynamic sampling ratio scheduler, which selects more data in the early stage of the training, and fewer data in the later stage when the model converges gradually. The details of the sampling ratio scheduler are shown in Algorithm \ref{alg:sample_rate}.

\subsection{Mixture of Metrics}

\begin{figure}[!h]
\centering
\includegraphics[width=0.4\columnwidth]{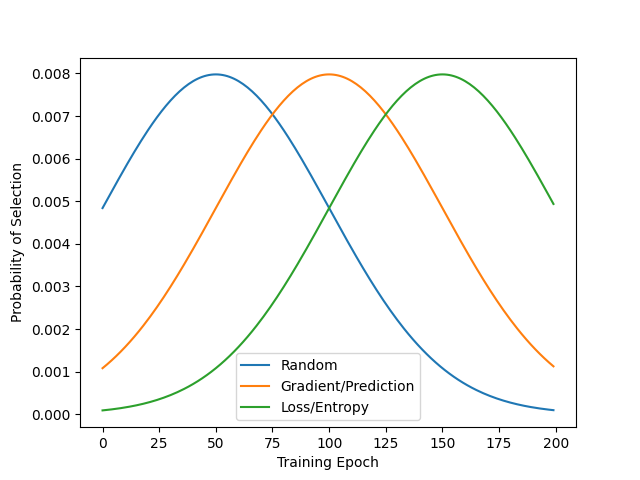} 
\caption{The probability of different proxy Metric.}.
\label{mix_prob}
\end{figure}

As mentioned before, different importance metrics will have different influences on the quality of data selection. Using a constant metric as an importance metric for all epochs may not be optimal. We propose a mixture of metrics to address this challenge. Furthermore, instead of choosing a fixed importance metric at the various training stages, we heuristically propose a probability distribution-based mixture of metrics to balance the exploration and exploitation during the training. 
  
As shown in Fig.~\ref{mix_prob}, without losing the generalizability, all metrics follow a Gaussian distribution with the same variance and different mean, and the x-axis is the number of training epochs. With this heuristical setting, the relative probability of each metric being selected at each epoch is different, which enables the dynamic selection of metrics.
Based on this probability distribution, we will choose one of these metrics as the importance metric for each epoch during training, which will be used to calculate the sampling probability of data selection.

\paragraph{Ablation analysis on our heuristical setting}
Particularly, we experimentally find that the above five importance metrics can be divided into 3 groups according to their nature.
In the early training stage, data diversity is more important, so a Random metric (group 0) should be preferred. In the middle training stage, the model needs to converge quickly and improve the sample prediction accuracy, so Prediction and Gradient (group 1) are more important. In the later training stage, the model enters a gradual convergence period and the fine-grained predictive distribution calculated by Loss and Entropy (group 2) is more critical.
In order to verify our hypothesis, we swap the Gaussian distribution position of these three groups of metrics and demonstrate the evaluation performance over each combination in Tab.\ref{tab:performance_proxy}. \textbf{The results in Tab.\ref{tab:performance_proxy} validate our hypothesis that our proposed $g012$ achieves the best performance compared to other combinations.}

\begin{table}[!b]
    \centering
    \fontsize{9}{12}\selectfont
    \begin{tabular}{l|cccccc}
    \toprule
    Mixture &g012 & g021 & g102 & g120 & g201 & g210 \\
    \hline
    Prec@1 &\textbf{95.41}        & 94.75        & 94.95        & 94.71        & 94.83        &   94.76     \\  
    \bottomrule
    \end{tabular}
    \caption{The comparison of different mixtures of metric. $gabc$ means that we set the mean axis of the Gaussian distribution of metric group a, group b, and group c to the 1/4, 2/4, and 3/4 fractions of the total epochs.}
    \label{tab:performance_proxy}
\end{table}

\subsection{Prediction Metrics for Data Selection with Replacement}

We have reviewed five commonly importance metrics, i.e. Random \cite{moser2022less}, Gradient \cite{aljundi2019gradient}, Loss\cite{jiang2019accelerating}, Entropy\cite{settles2009active}, Prediction metric, and most of them are in the format of continuous values and can be directly used as importance metric. 
For example, the loss or entropy values will decrease with the learning process and the larger value means that this sample is hard to be learned. However, the prediction metric is a discrete value, where true means correct and false means error. 
\cite{toneva2018empirical, zhang2021efficient, coleman2020selection} counted the transitions number of a sample whose prediction status (true or false) swaps, which is called the "forgetting events" importance metric.
Forgetting events only can record the samples of the continuously activated state, but if a sample is deactivated, its forgetting event will stop updating and be invalid.
In the ASP framework, we provide another replacement mechanism to present the importance of samples with the Prediction metric.
Specially, We initial a zero array $S=\{s_{i}\}, i \in \{1,2,...N\}$ as prediction memory, and each sample prediction score $s_{i}$ is: 

\begin{equation}
s_{i+1} = 
\left
\{
\begin{array}{l}
s_{i}-1, \quad if  \ \hat{y_i} == y_i \\
s_{i}+1, \quad if  \ \hat{y_i} \ != y_i \\
s_{i}+0, \quad if  \ \hat{y_i} == None
\end{array}
\right.
\end{equation}

Where $\hat{y_i}$ is the label and  $y_i$ is the prediction. $s_{i}$ presents the cumulative importance value until epoch i.
If a sample is predicted correctly, the prediction is -1, else, it is 1. What's more, if the sample doesn't participate in model training and has no prediction score, i.e. $y_i$ is None, then its score is 0. Finally, we add the current prediction score to $s_{i}$ and regard $S$ as the proxy memory $V$.

\section{More Experiment Results}
\begin{figure*}[!ht]
\begin{center}
\subfigure[Data Ratio 10\%]{
\includegraphics[width=.235\linewidth]{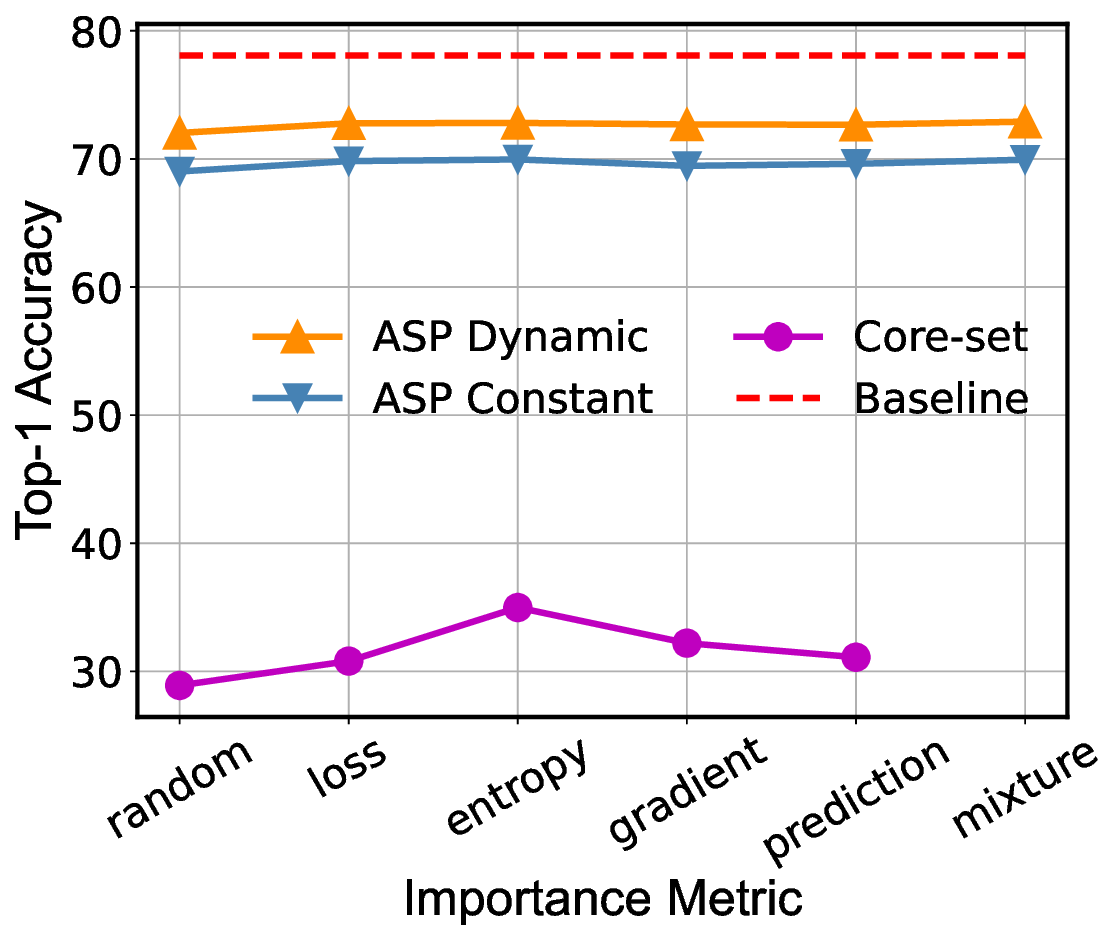}
\label{fig:18_c100_a}
}
\subfigure[Data Ratio 30\%]{
\includegraphics[width=.23\linewidth]{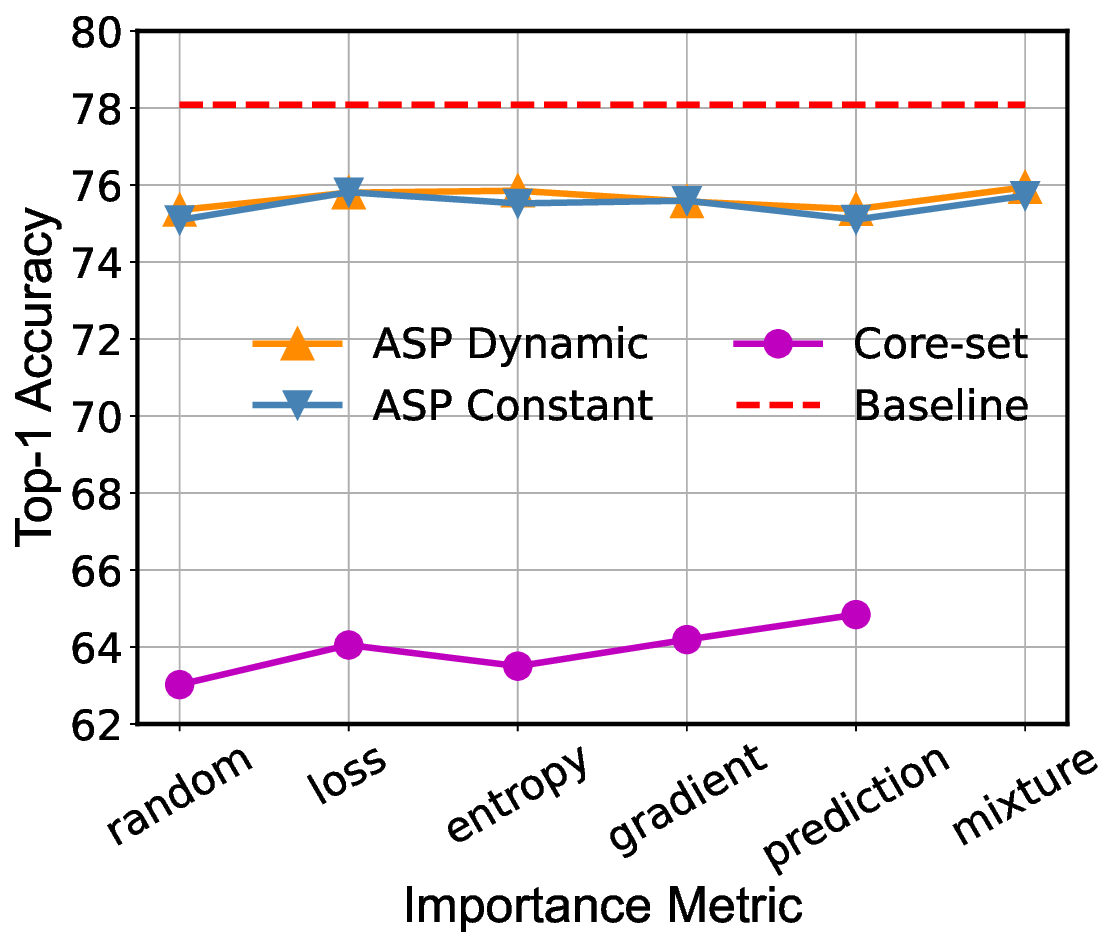}
}
\subfigure[Data Ratio 50\%]{
\includegraphics[width=.23\linewidth]{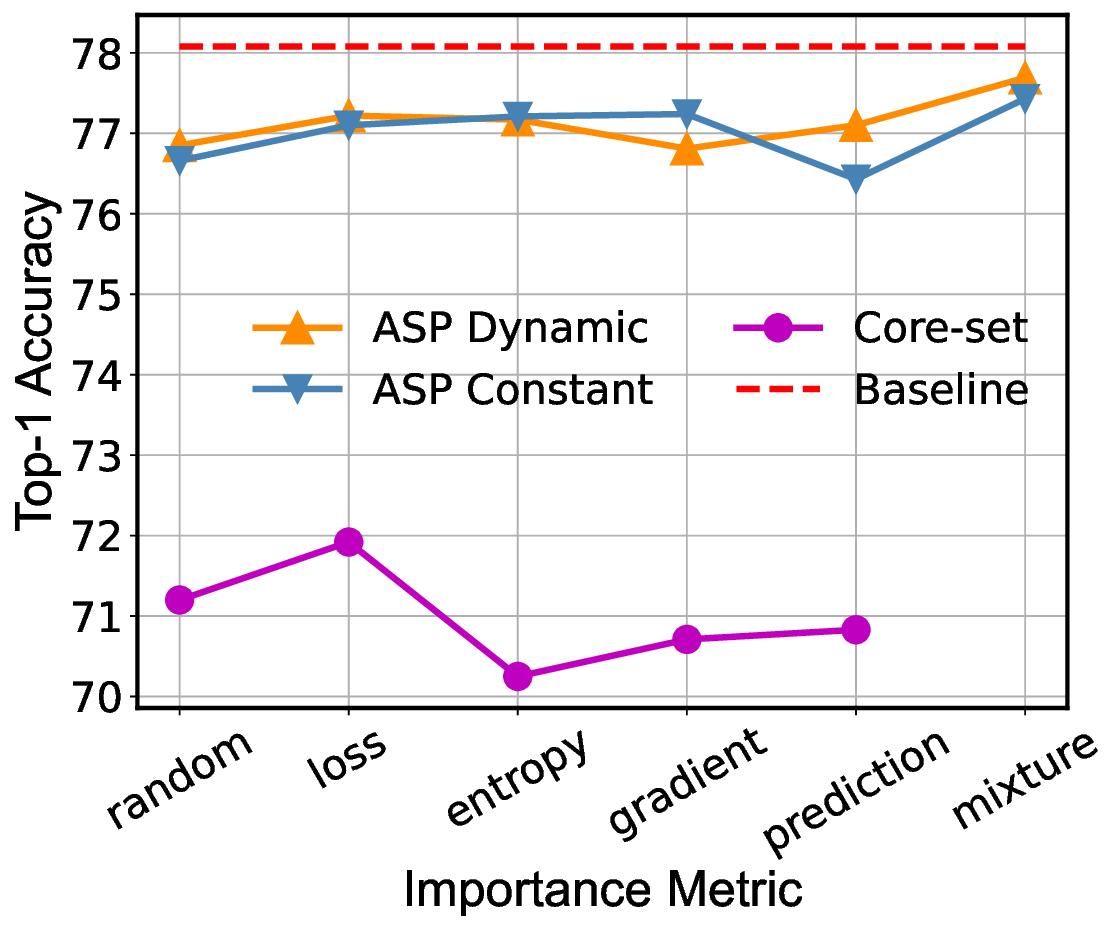}
\label{fig:18_c100_c}
}
\subfigure[Data Ratio 70\%]{
\includegraphics[width=.23\linewidth]{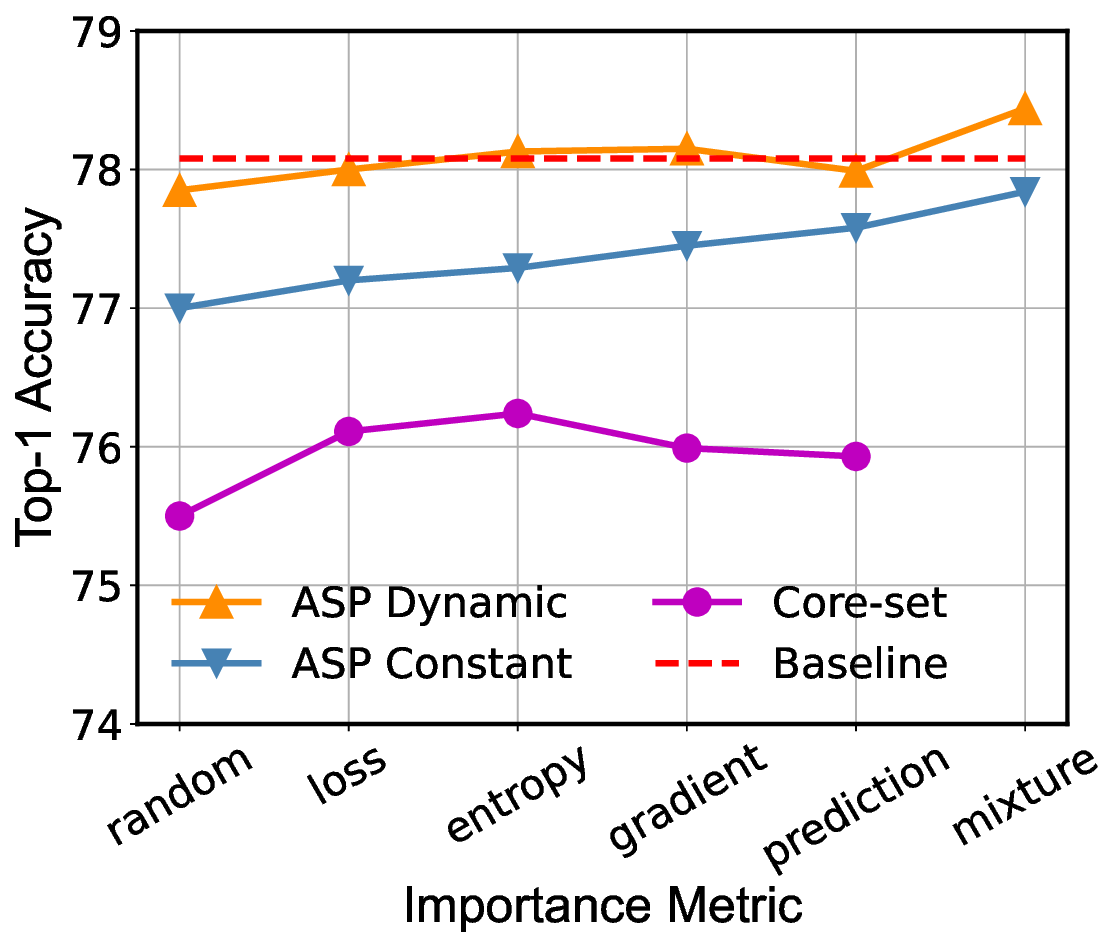}\label{fig:18_c100_d}
}
\caption{Test performance (Top-1 Accuracy(\%)) of ResNet-18 on CIFAR-100 under various data sampling ratios, data selection methods, and importance metrics.}
\label{fig:18_c100}
\end{center}
\end{figure*}

\begin{figure*}[!ht]
\begin{center}
\subfigure[Data Ratio 10\%]{
\includegraphics[width=.235\linewidth]{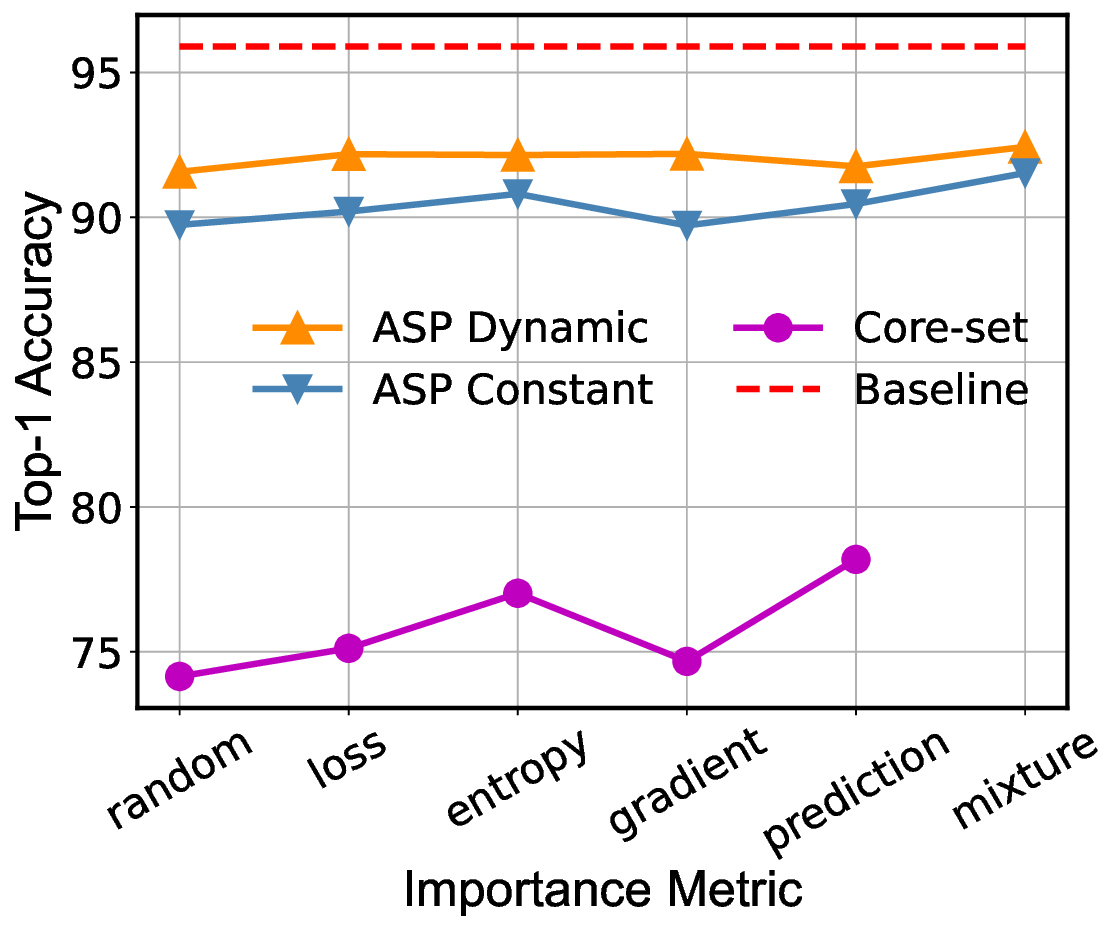}
\label{fig:34_c10_a}
}
\subfigure[Data Ratio 30\%]{
\includegraphics[width=.23\linewidth]{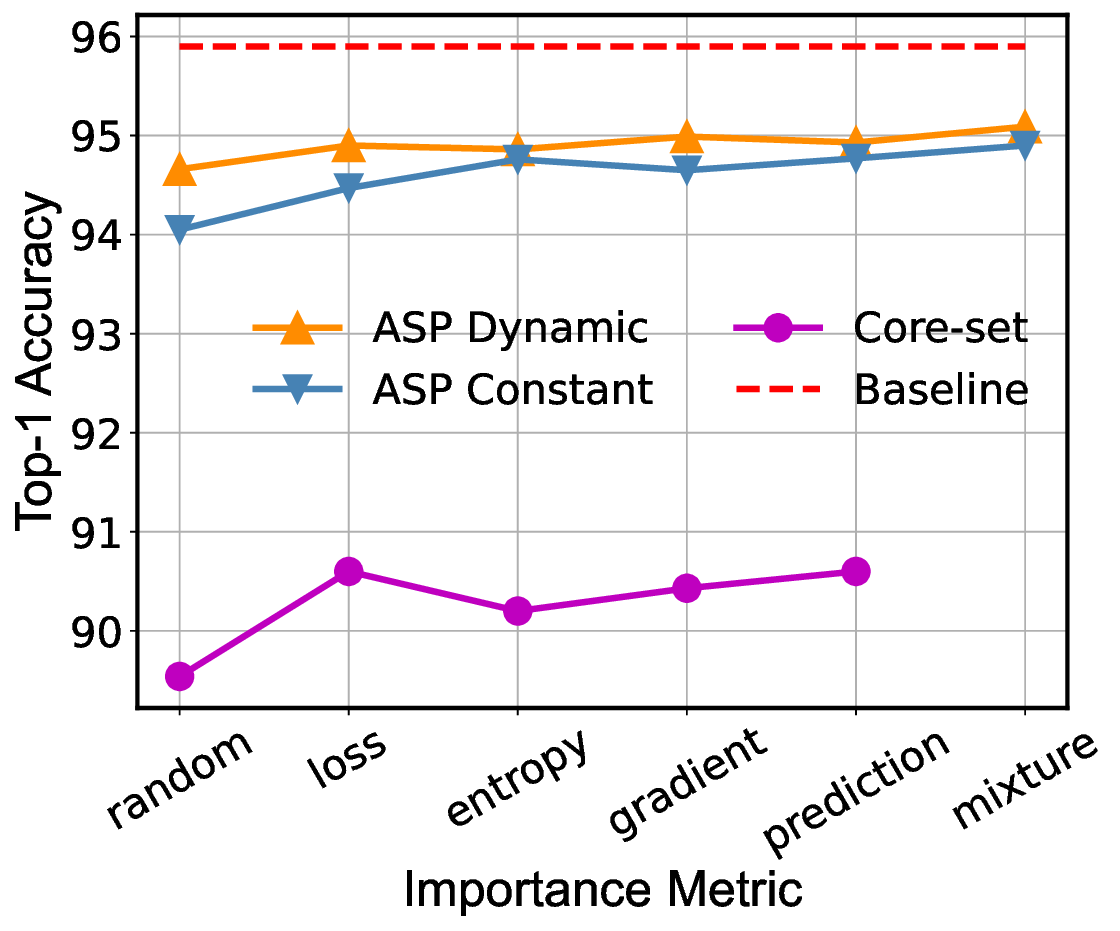}
}
\subfigure[Data Ratio 50\%]{
\includegraphics[width=.23\linewidth]{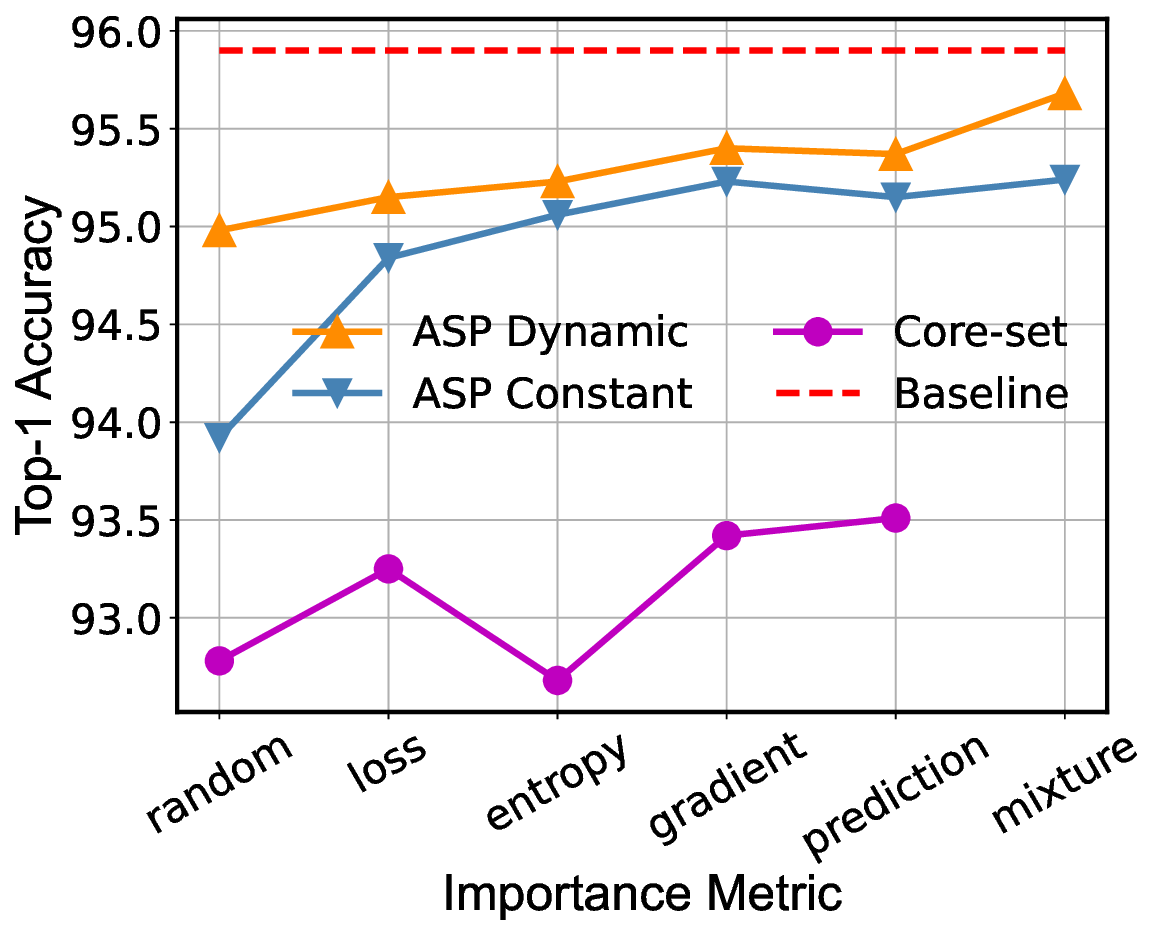}
\label{fig:34_c10_c}
}
\subfigure[Data Ratio 70\%]{
\includegraphics[width=.23\linewidth]{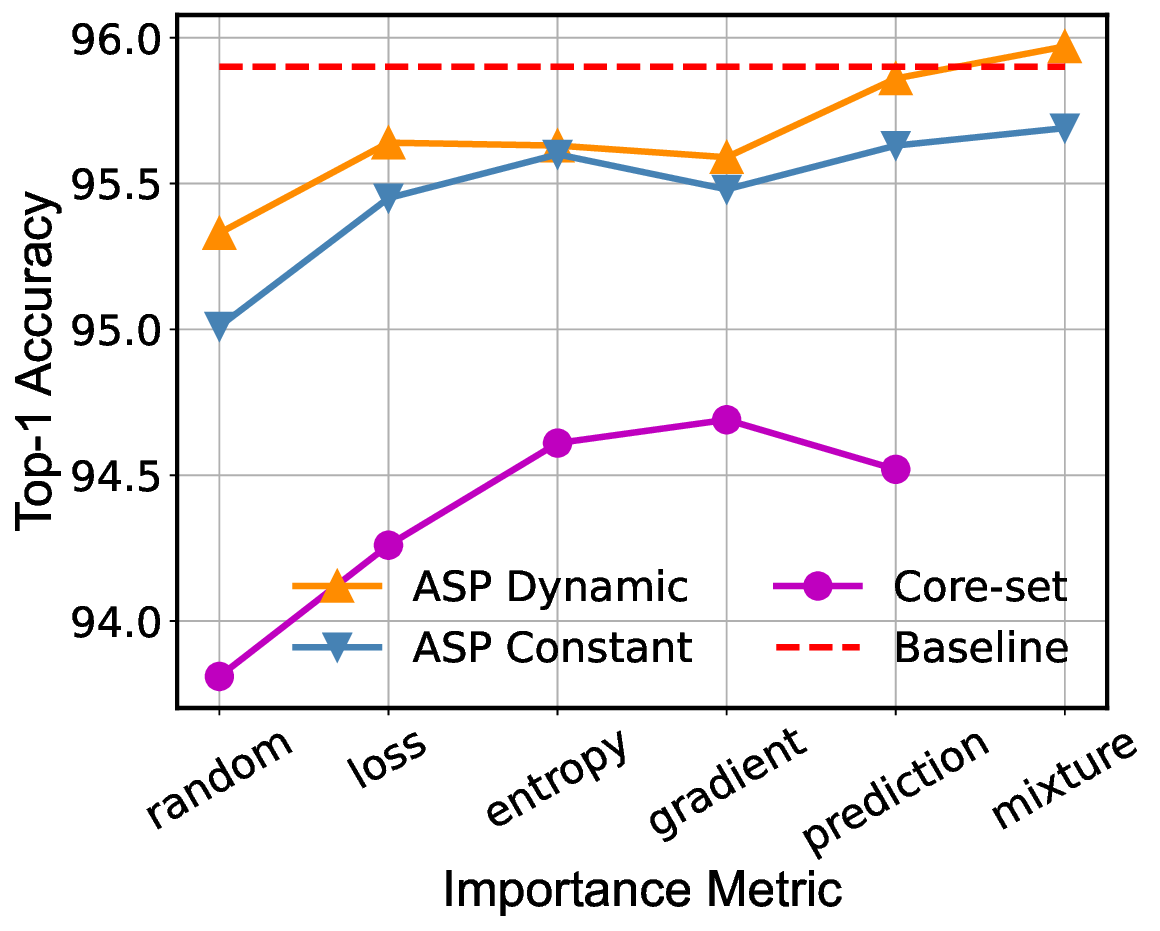}\label{fig:34_c10_d}
}
\caption{Test performance (Top-1 Accuracy(\%)) of ResNet-34 on CIFAR-10 under various data sampling ratios, data selection methods, and importance metrics.}
\label{fig:34_c10}
\end{center}
\end{figure*}

\begin{figure*}[!ht]
\begin{center}
\subfigure[Data Ratio 10\%]{
\includegraphics[width=.235\linewidth]{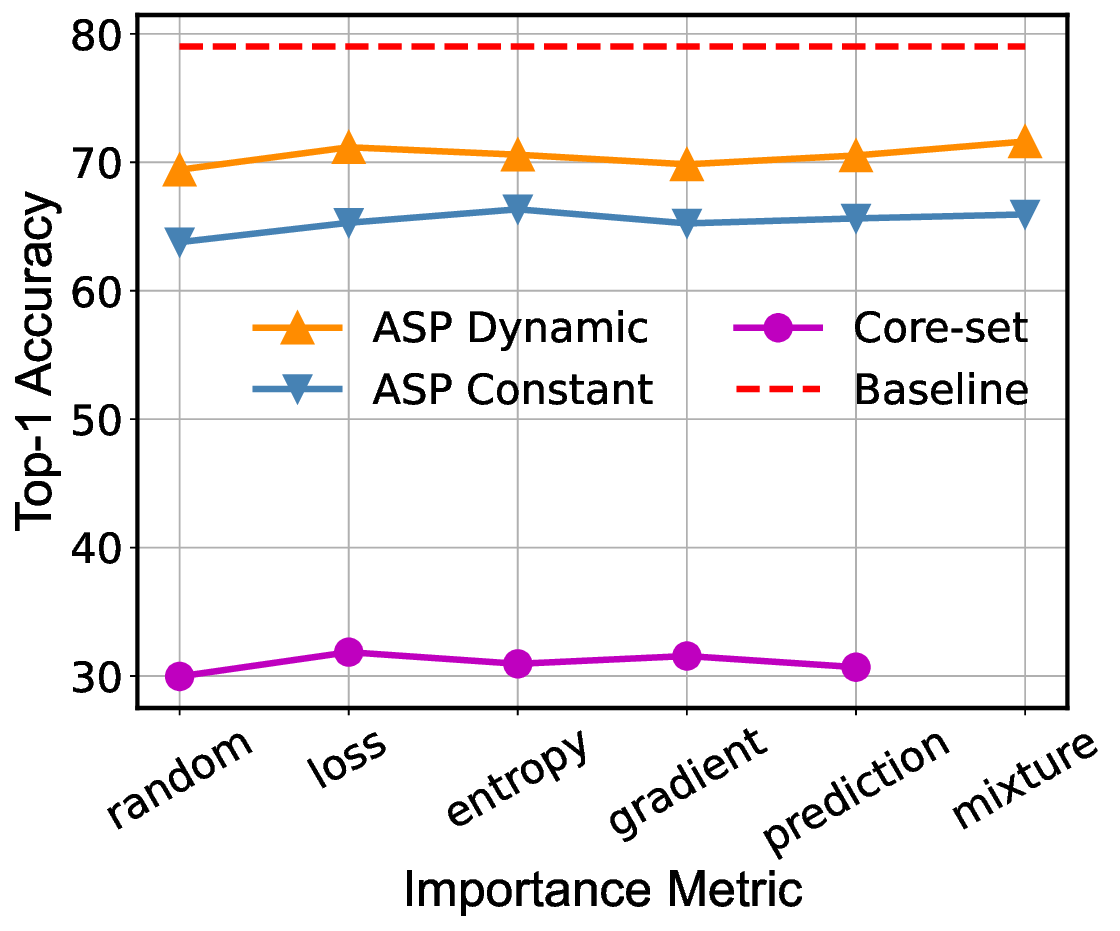}
\label{fig:34_c100_a}
}
\subfigure[Data Ratio 30\%]{
\includegraphics[width=.23\linewidth]{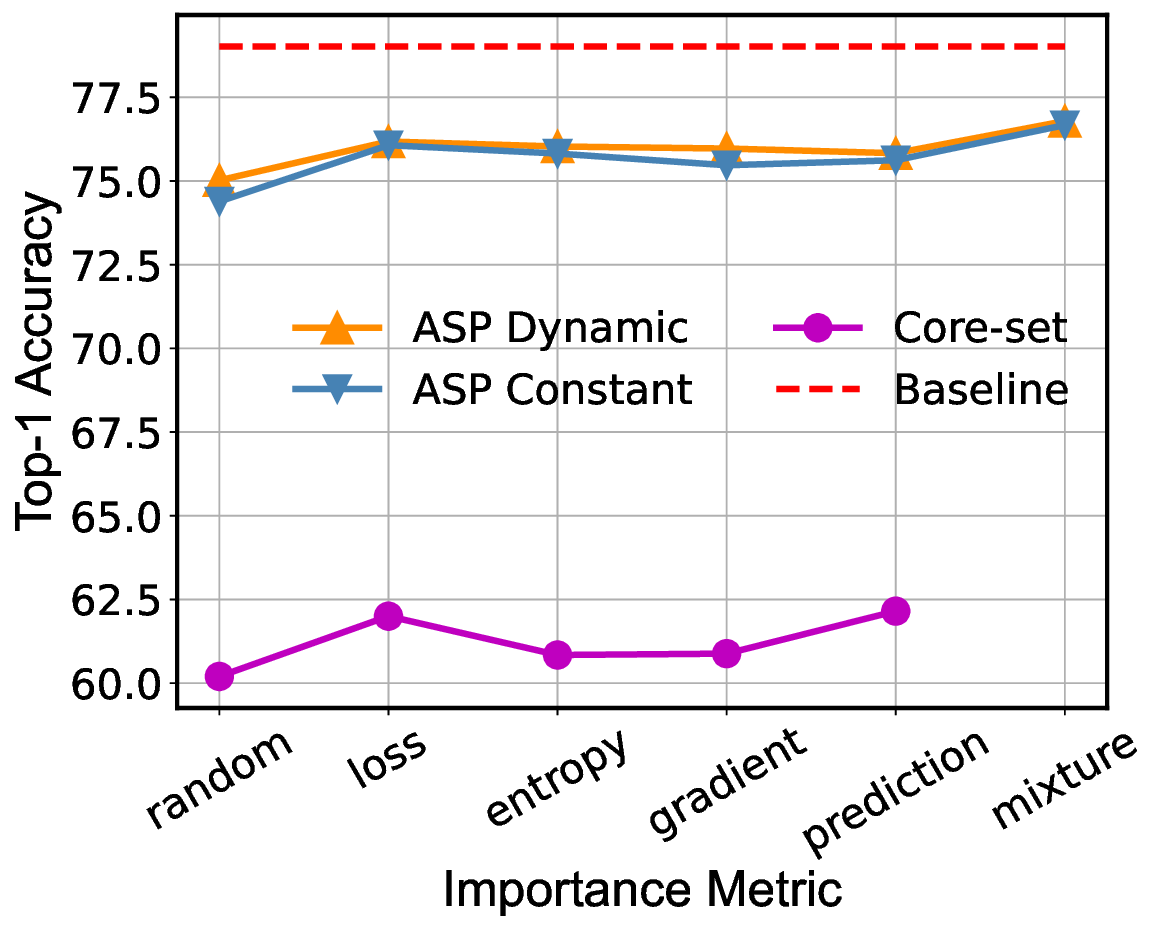}
}
\subfigure[Data Ratio 50\%]{
\includegraphics[width=.23\linewidth]{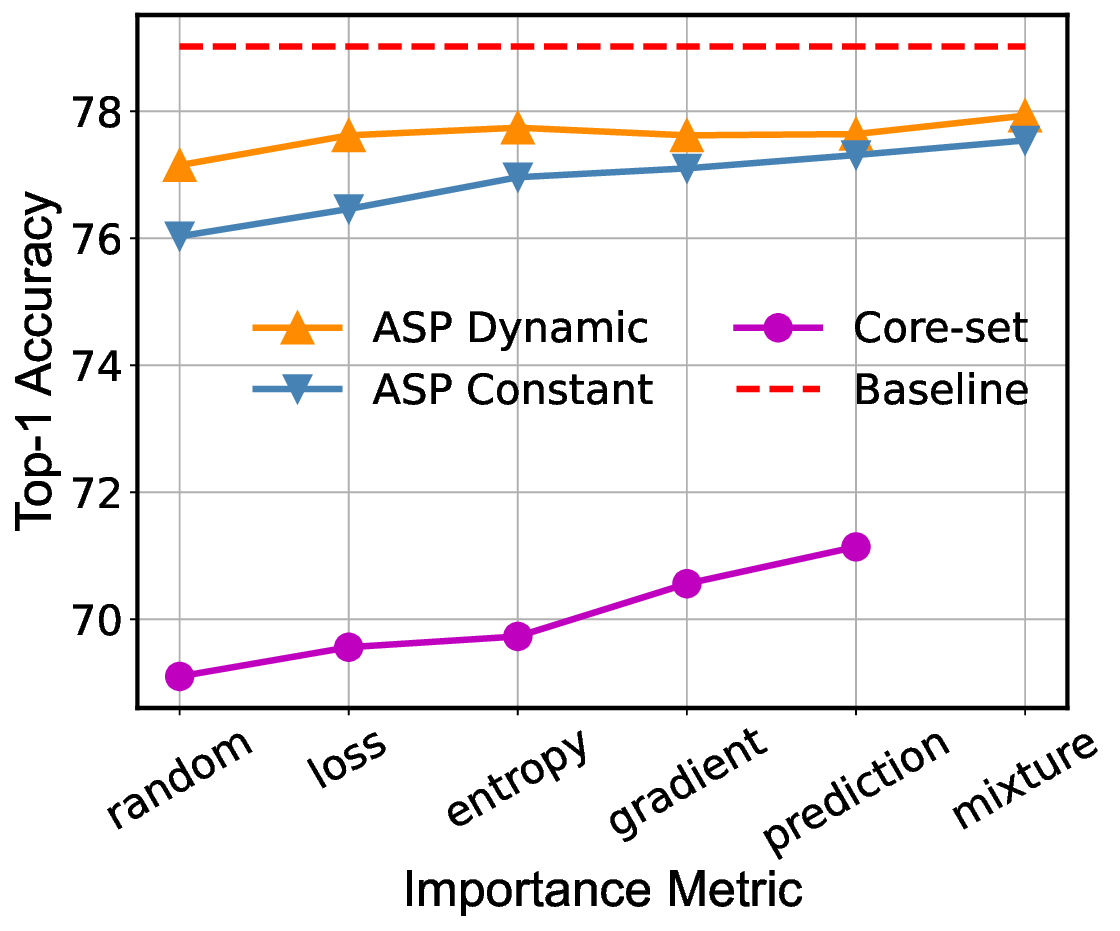}
\label{fig:34_c100_c}
}
\subfigure[Data Ratio 70\%]{
\includegraphics[width=.23\linewidth]{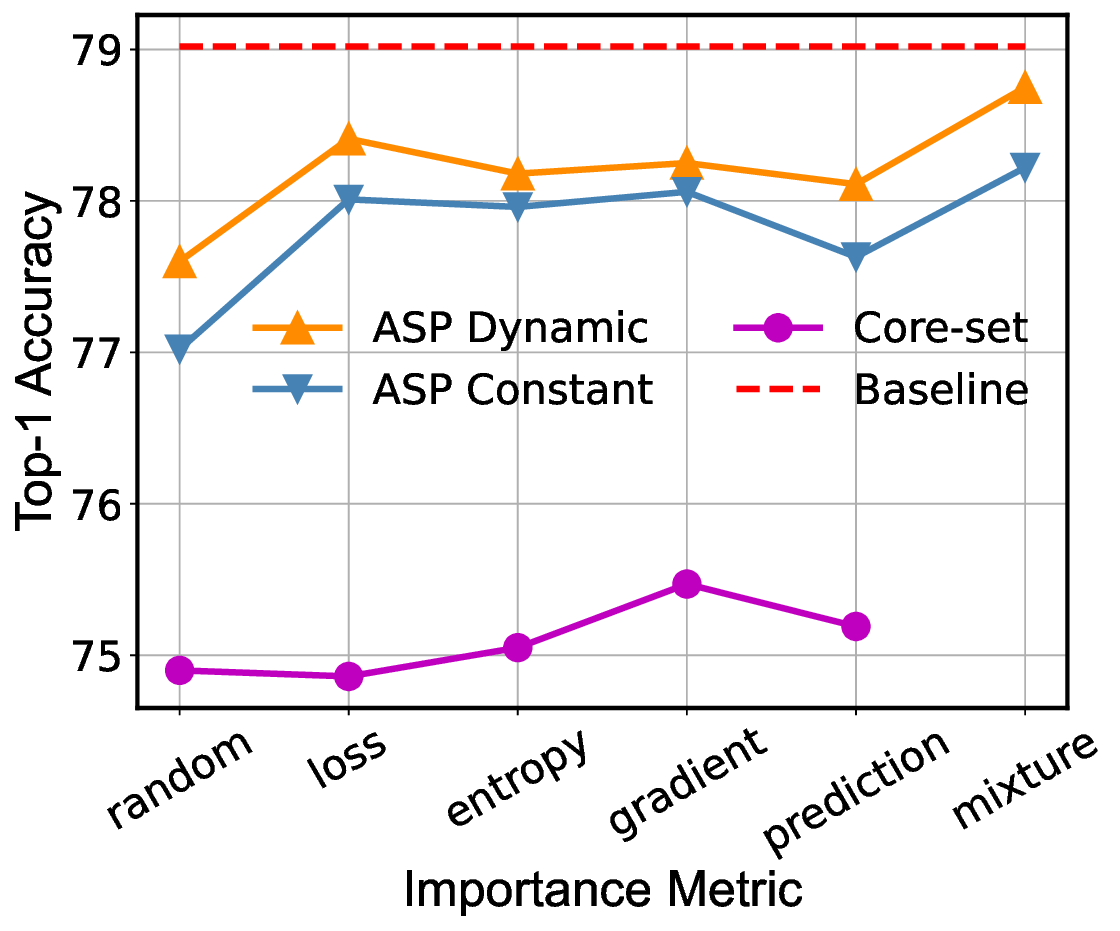}
}
\caption{Test performance (Top-1 Accuracy(\%)) of ResNet-34 on CIFAR-100 under various data sampling ratios, data selection methods, and importance metrics.}
\label{fig:34_c100}
\end{center}
\end{figure*}

\begin{figure*}[!h]
\begin{center}
\subfigure[Data Ratio 10\%]{
\includegraphics[width=.235\linewidth]{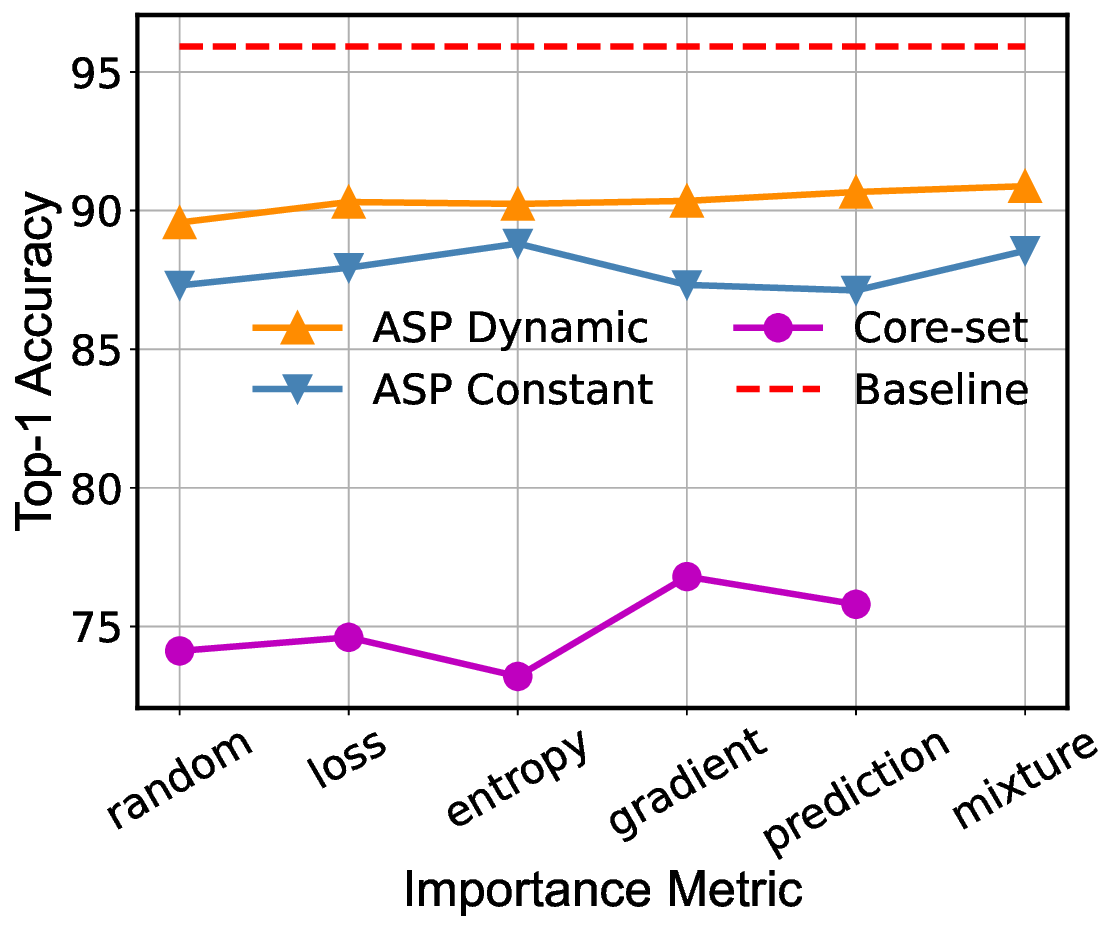}
\label{fig:101_c10_a}
}
\subfigure[Data Ratio 30\%]{
\includegraphics[width=.23\linewidth]{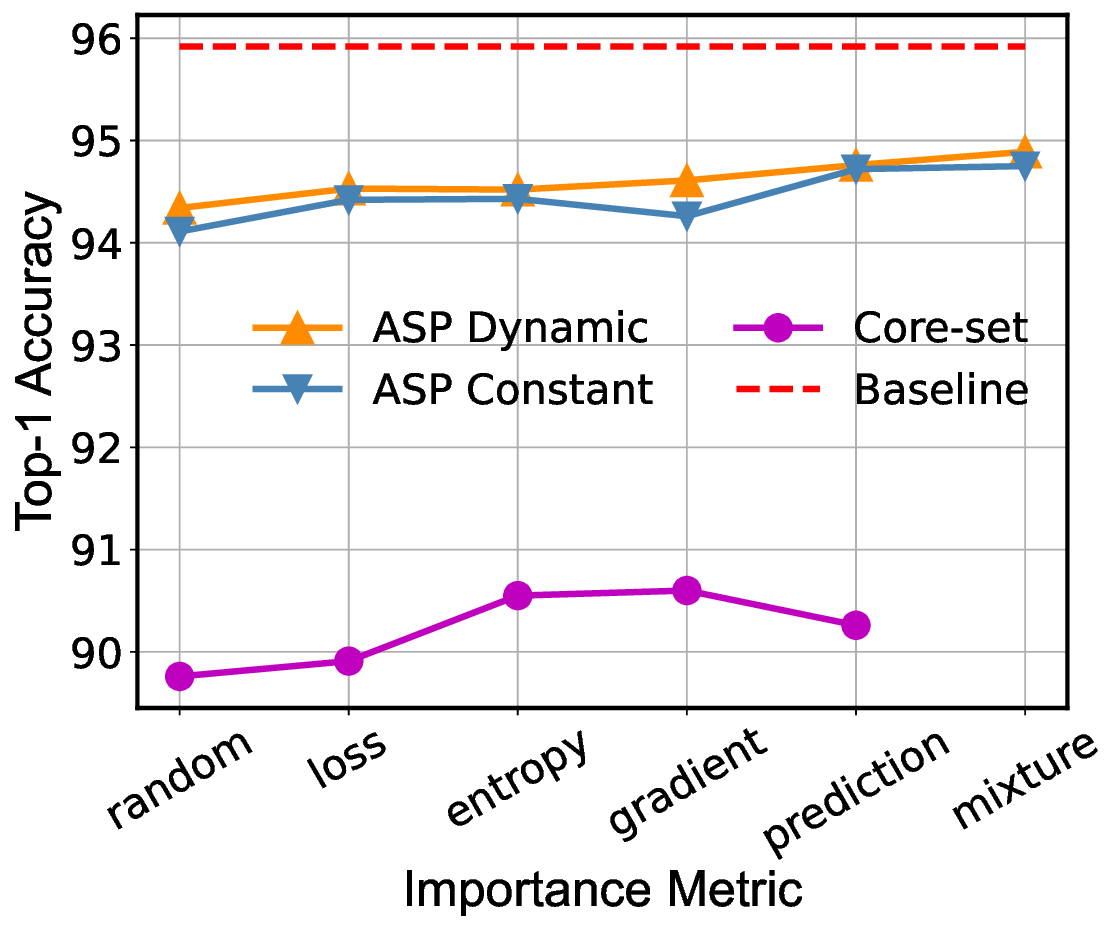}
}
\subfigure[Data Ratio 50\%]{
\includegraphics[width=.23\linewidth]{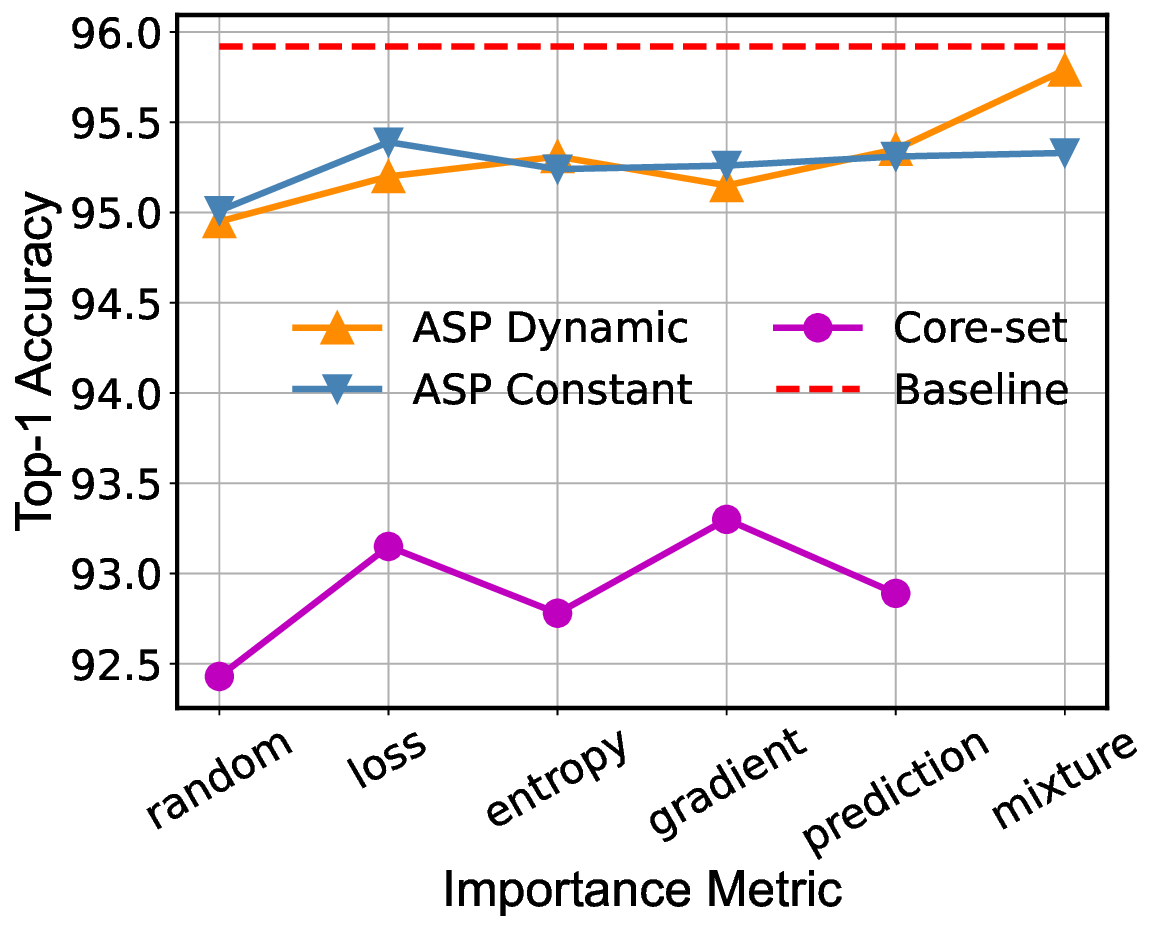}
\label{fig:101_c10_c}
}
\subfigure[Data Ratio 70\%]{
\includegraphics[width=.23\linewidth]{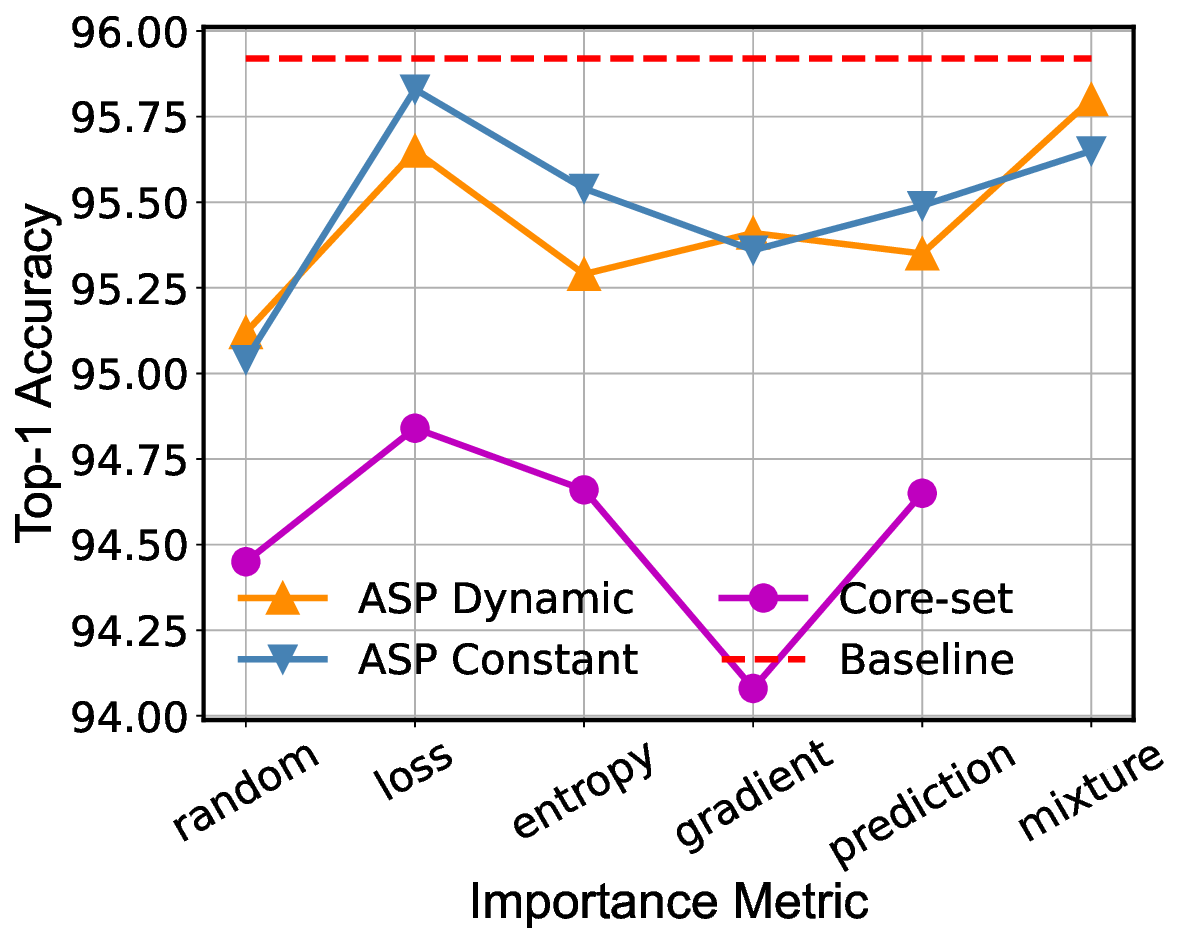}
}
\caption{Test performance (Top-1 Accuracy(\%)) of ResNet-101 on CIFAR-10 under various data sampling ratios, data selection methods, and importance metrics.}
\label{fig:101_c10}
\end{center}
\end{figure*}

\begin{figure*}[!h]
\begin{center}
\subfigure[Data Ratio 10\%]{
\includegraphics[width=.235\linewidth]{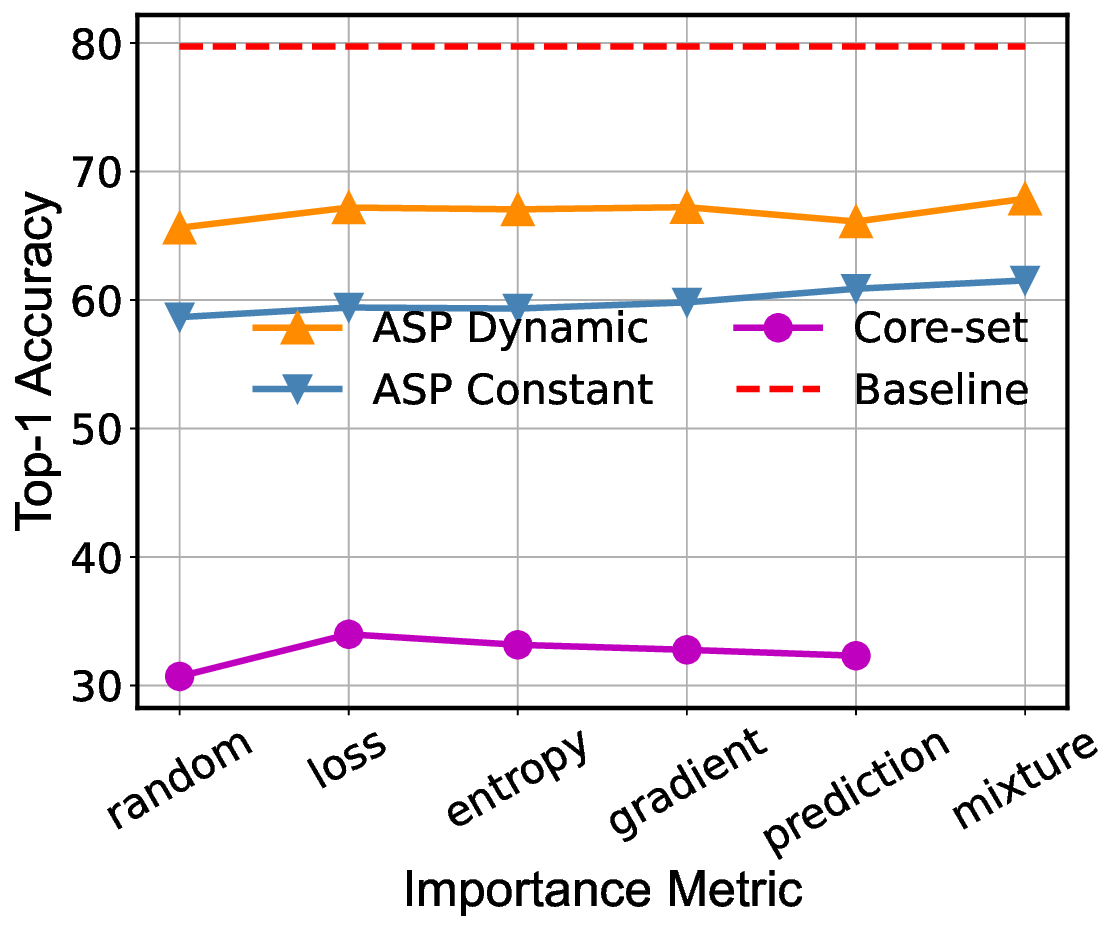}
\label{fig:101_c100_a}
}
\subfigure[Data Ratio 30\%]{
\includegraphics[width=.23\linewidth]{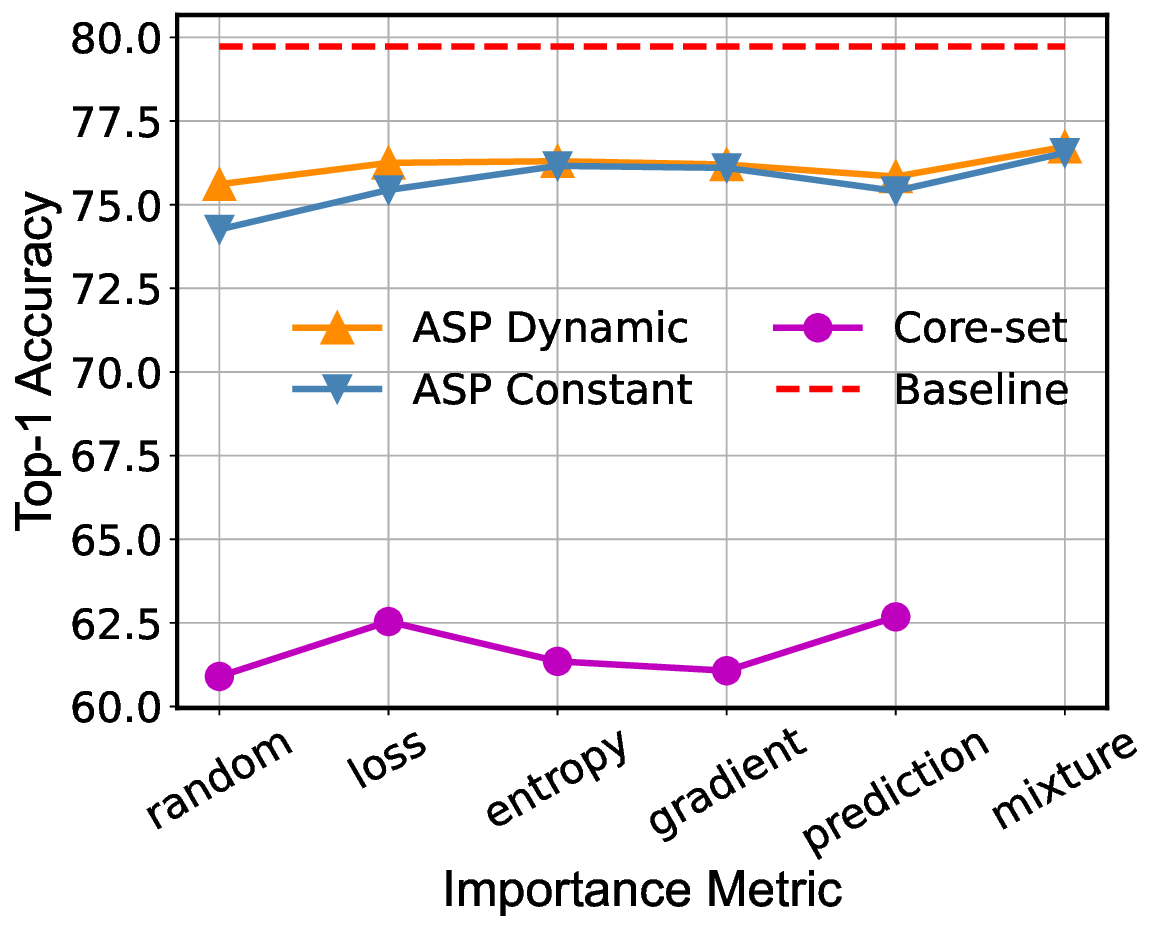}
}
\subfigure[Data Ratio 50\%]{
\includegraphics[width=.23\linewidth]{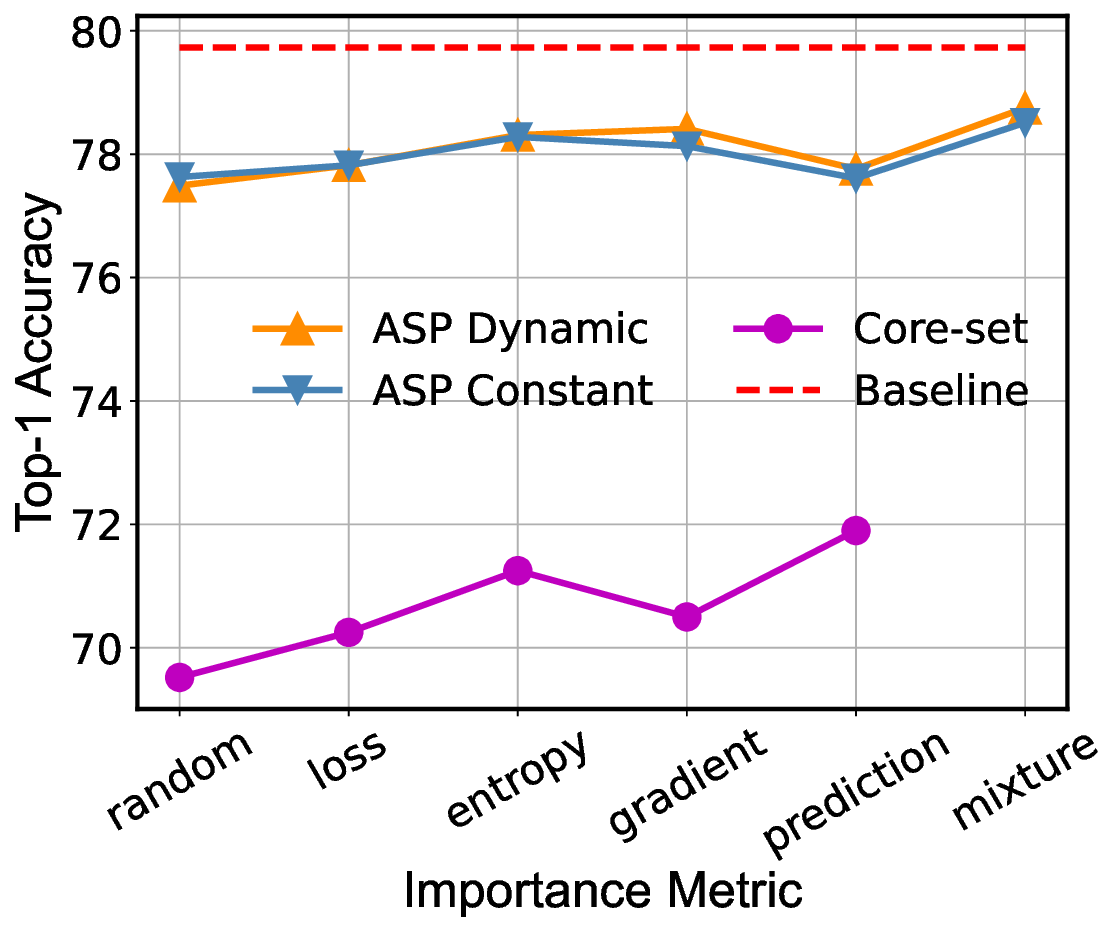}
\label{fig:101_c100_c}
}
\subfigure[Data Ratio 70\%]{
\includegraphics[width=.23\linewidth]{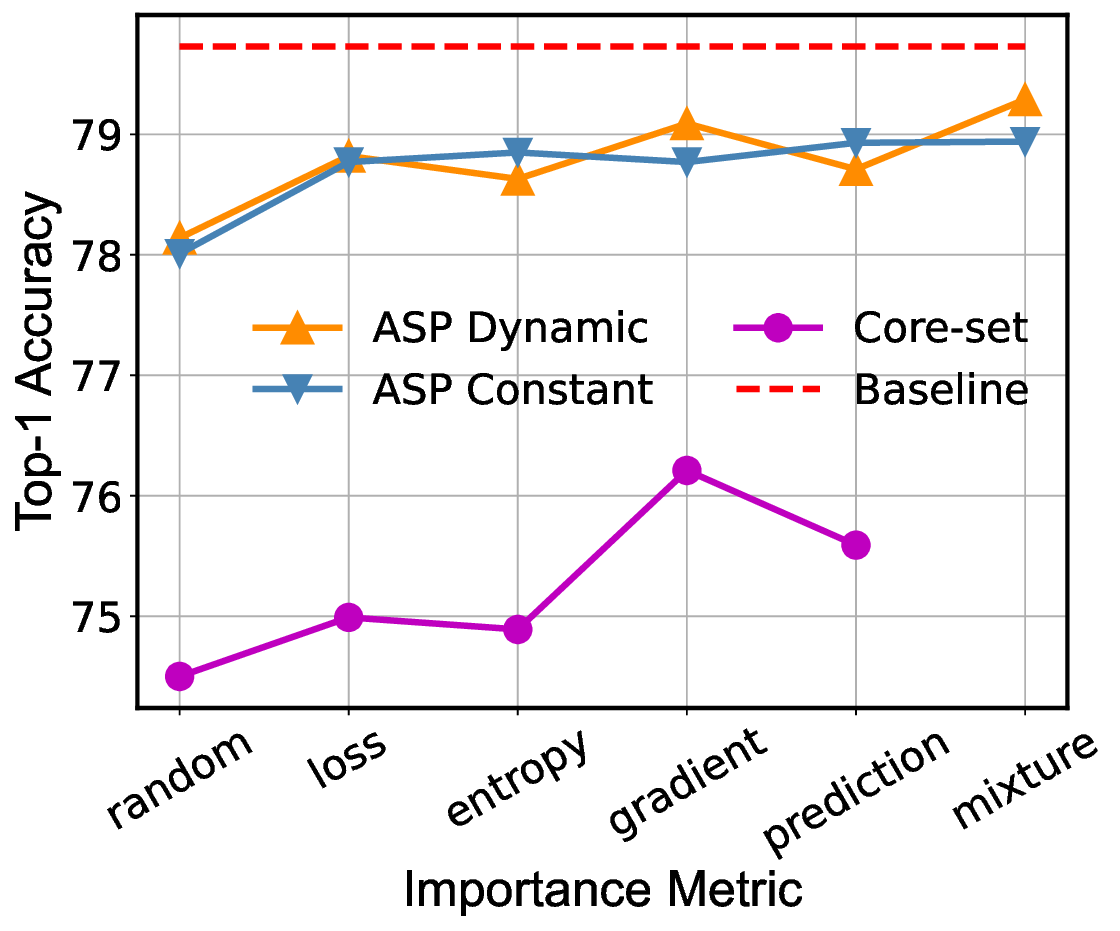}
}
\caption{Test performance (Top-1 Accuracy(\%)) of ResNet-101 on CIFAR-100 under various data sampling ratios, data selection methods, and importance metrics.}
\label{fig:101_c100}
\end{center}
\end{figure*}

\begin{figure*}[!h]
\begin{center}
\subfigure[Data Ratio 10\%]{
\includegraphics[width=.235\linewidth]{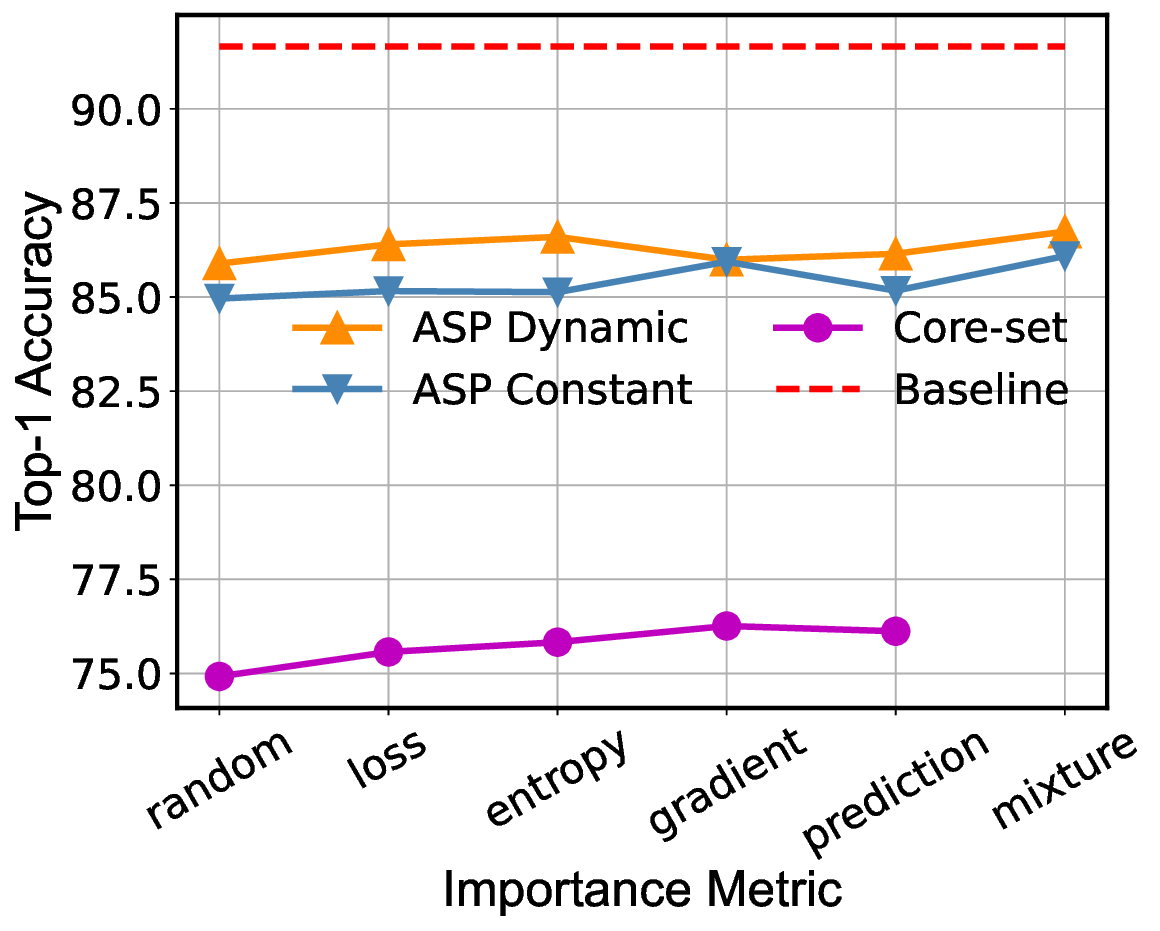}
\label{fig:20_c10_a}
}
\subfigure[Data Ratio 30\%]{
\includegraphics[width=.23\linewidth]{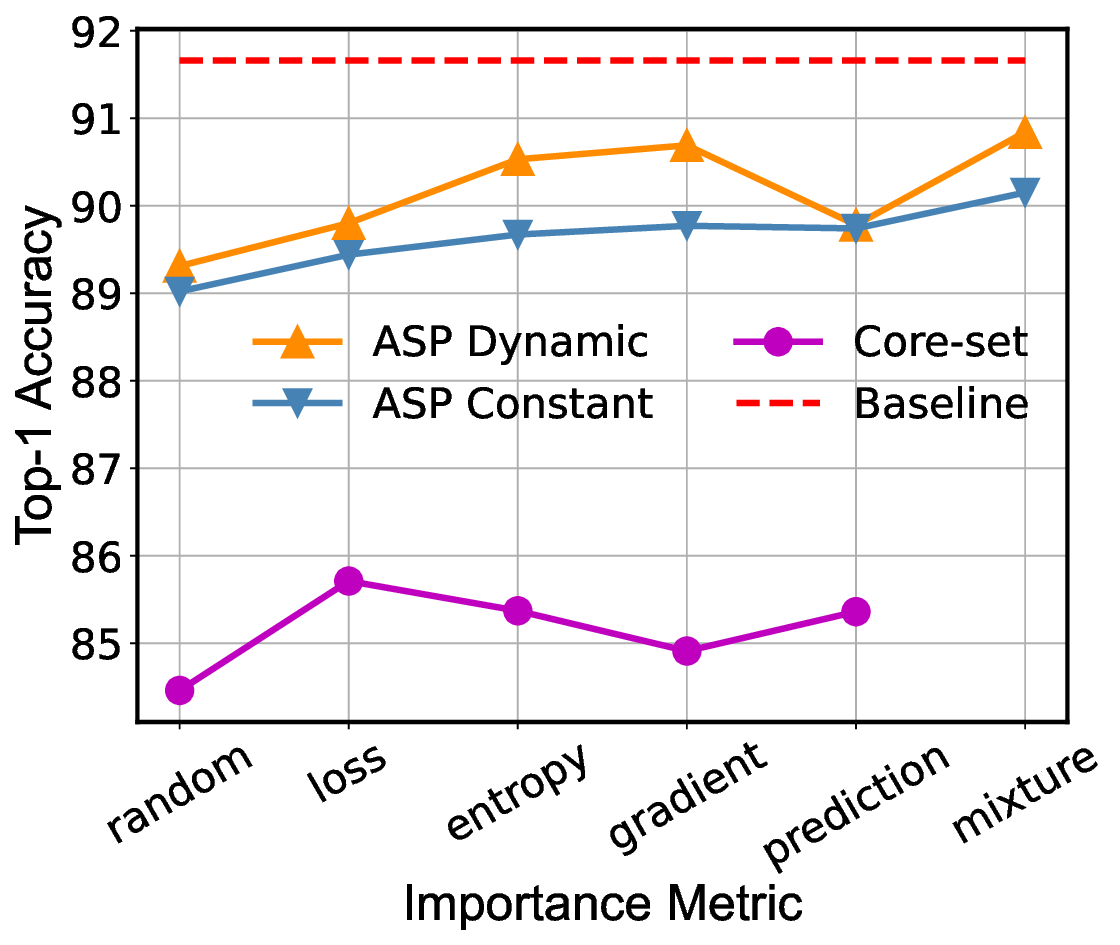}
}
\subfigure[Data Ratio 50\%]{
\includegraphics[width=.23\linewidth]{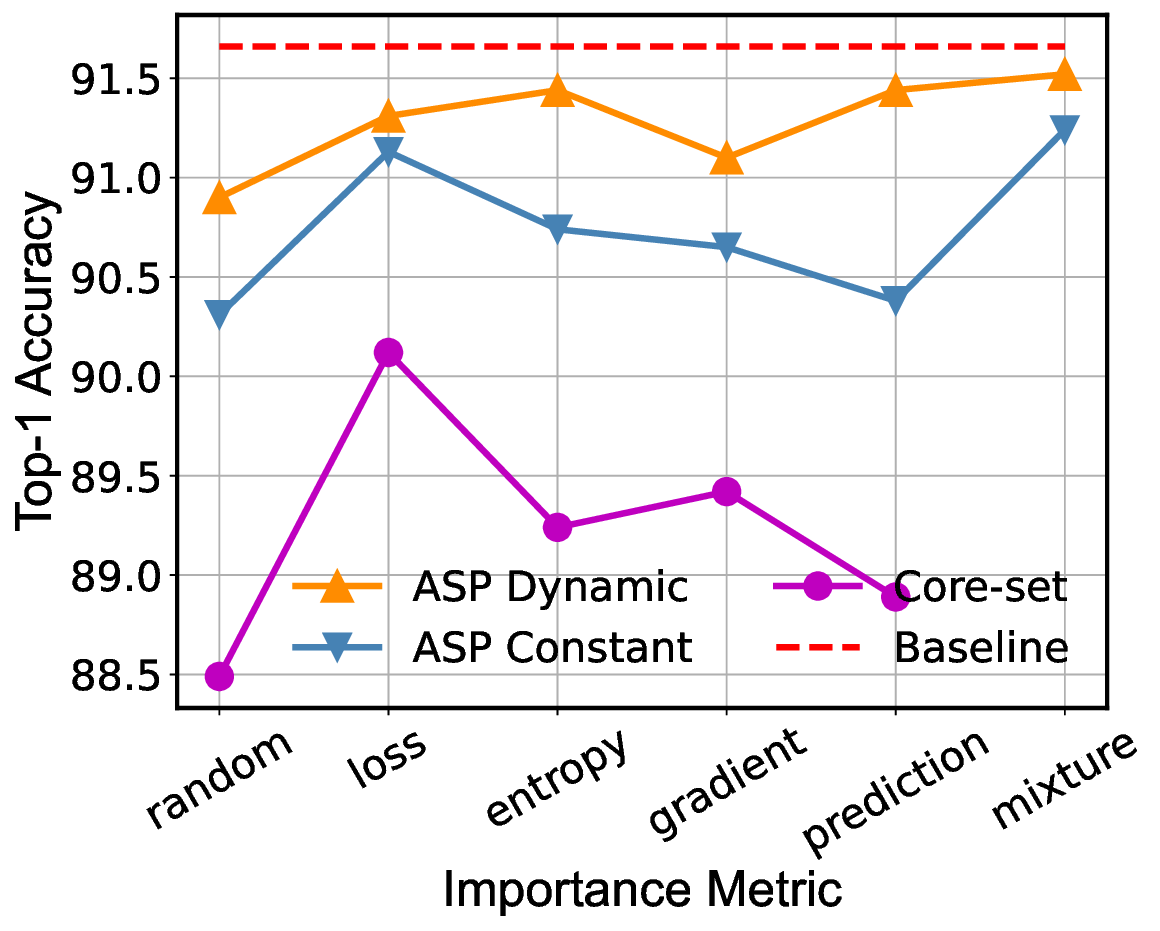}
\label{fig:20_c10_c}
}
\subfigure[Data Ratio 70\%]{
\includegraphics[width=.23\linewidth]{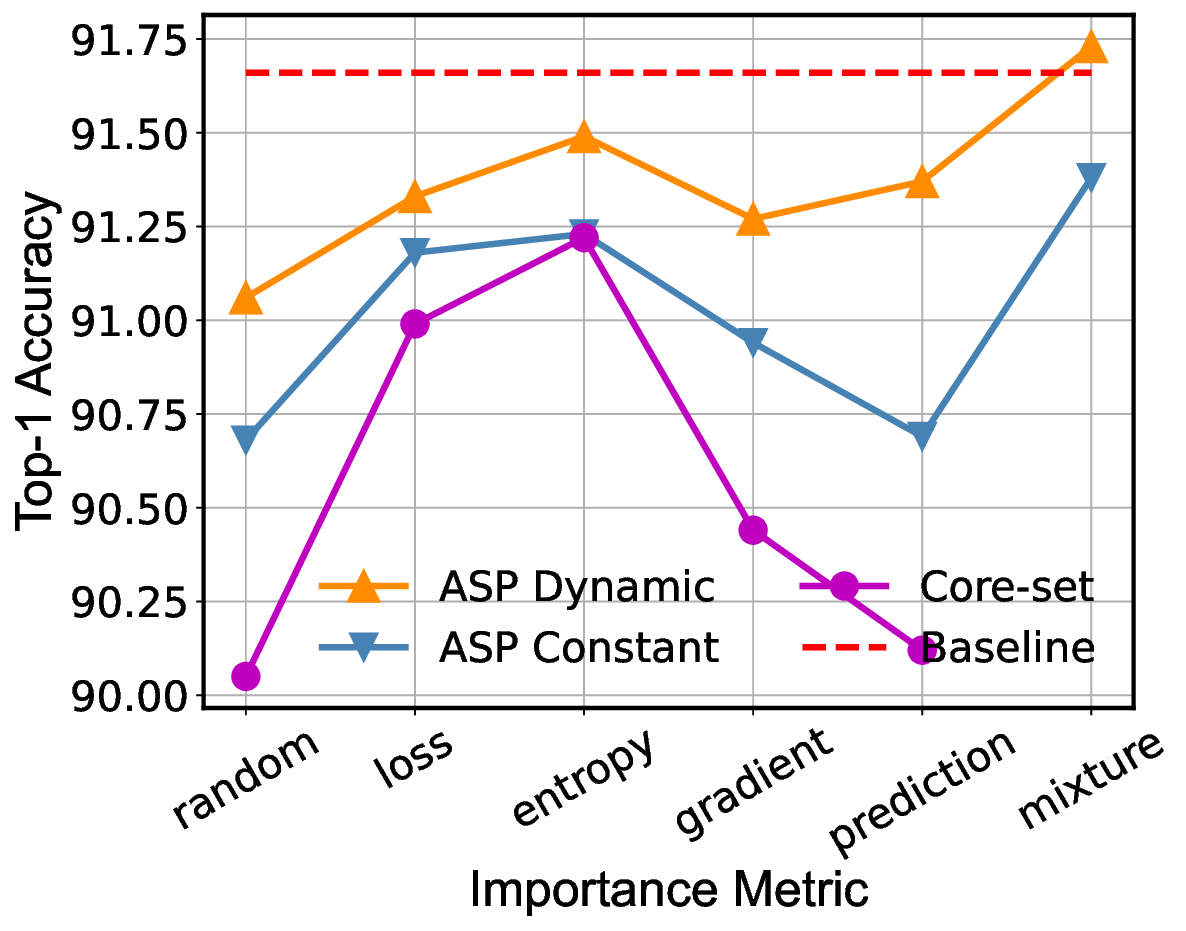}\label{fig:20_c10_d}
}
\caption{Test performance (Top-1 Accuracy(\%)) of ResNet-20 on CIFAR-10 under various data sampling ratios, data selection methods, and importance metrics.}
\label{fig:20_c10}
\end{center}
\end{figure*}

\begin{figure*}[!h]
\begin{center}
\subfigure[Data Ratio 10\%]{
\includegraphics[width=.235\linewidth]{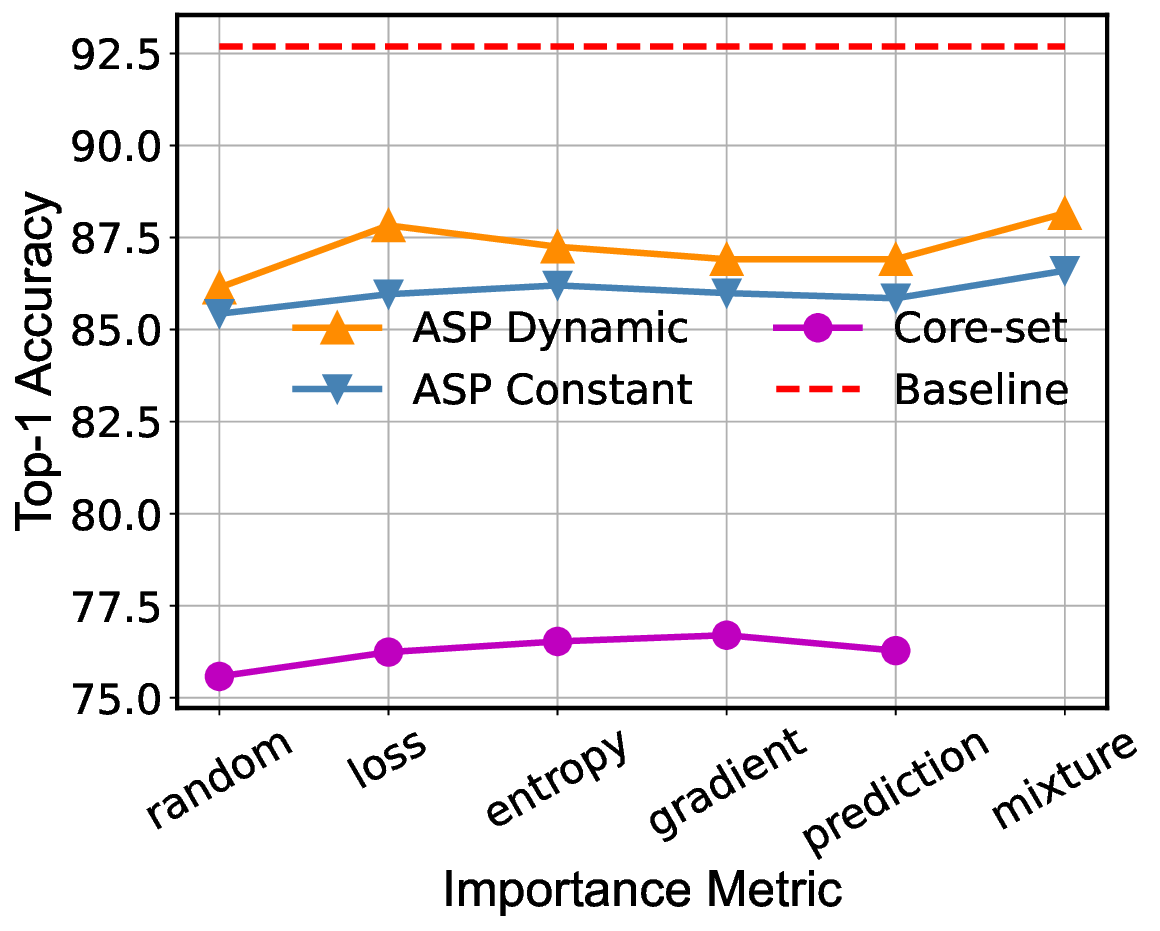}
\label{fig:32_c10_a}
}
\subfigure[Data Ratio 30\%]{
\includegraphics[width=.23\linewidth]{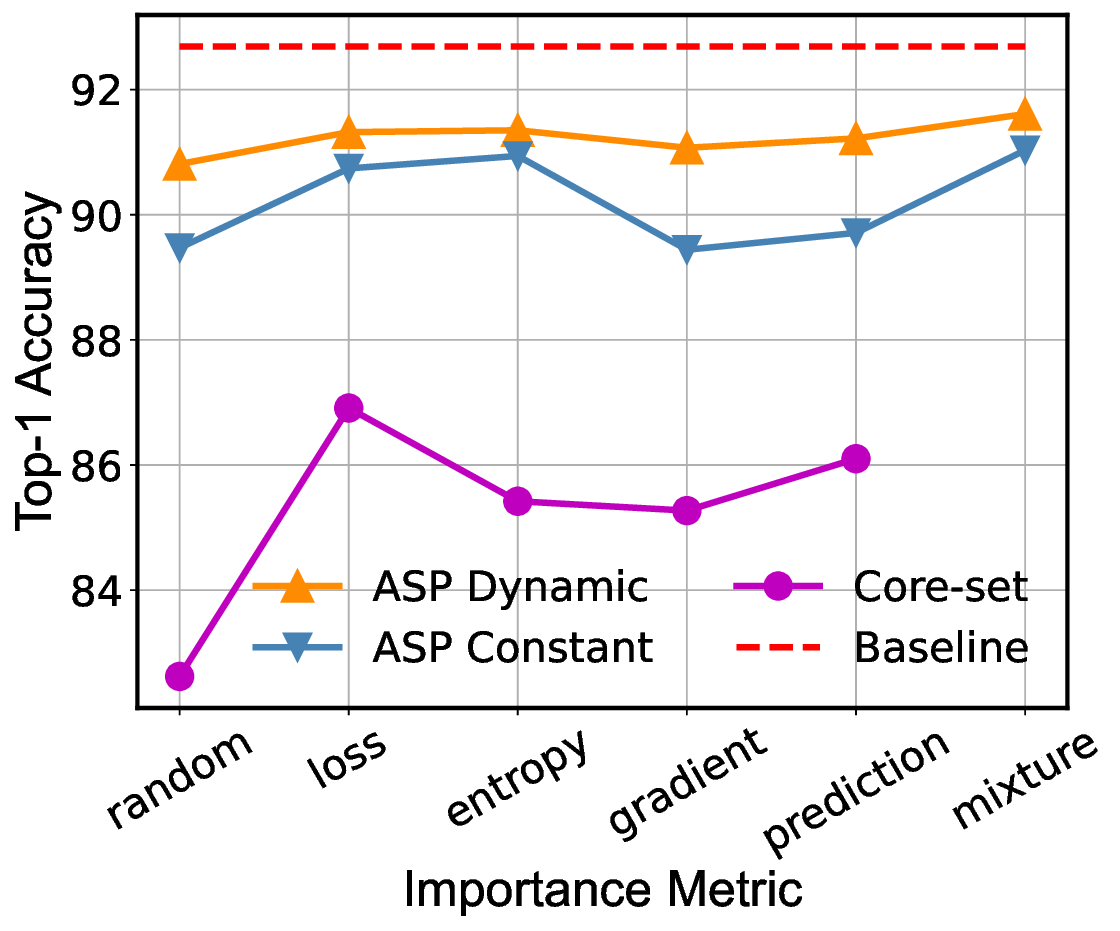}
}
\subfigure[Data Ratio 50\%]{
\includegraphics[width=.23\linewidth]{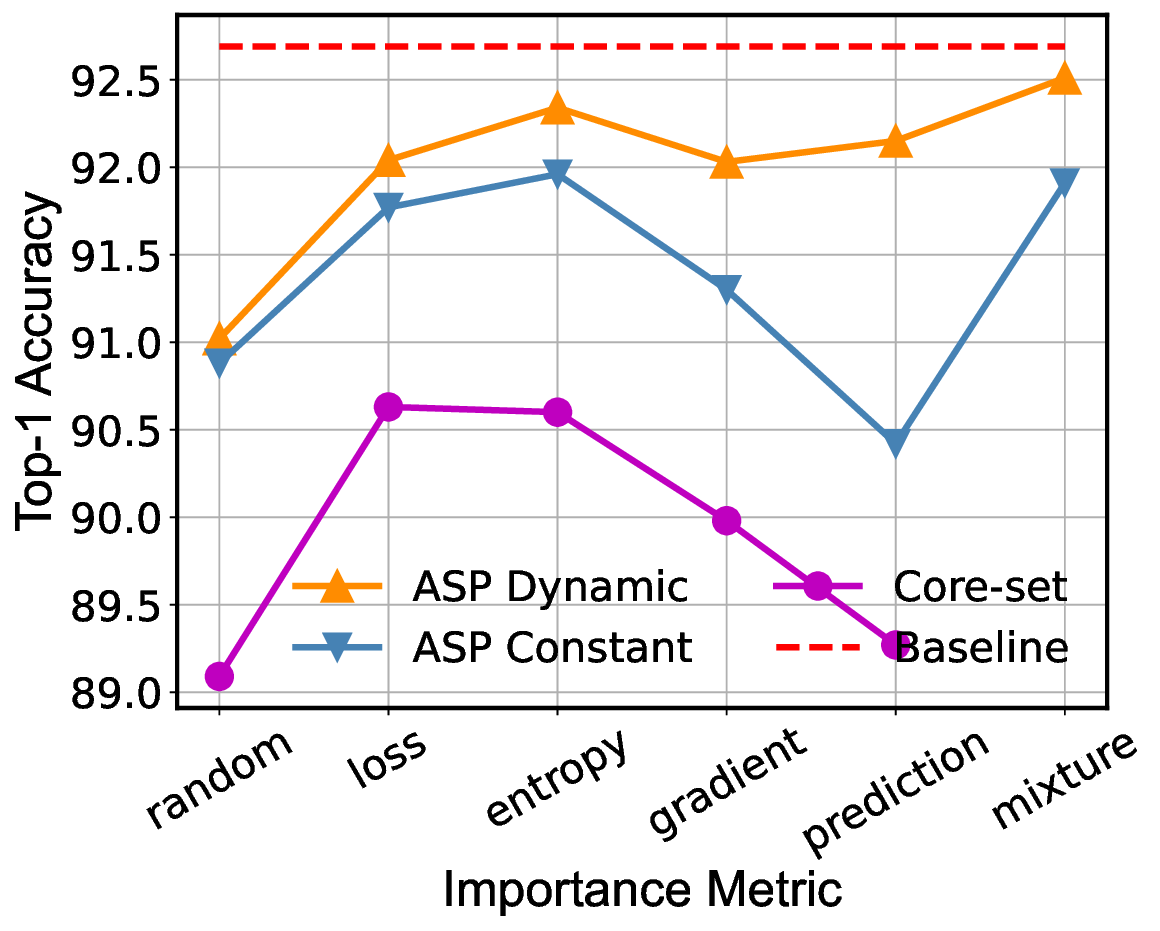}
\label{fig:32_c10_c}
}
\subfigure[Data Ratio 70\%]{
\includegraphics[width=.23\linewidth]{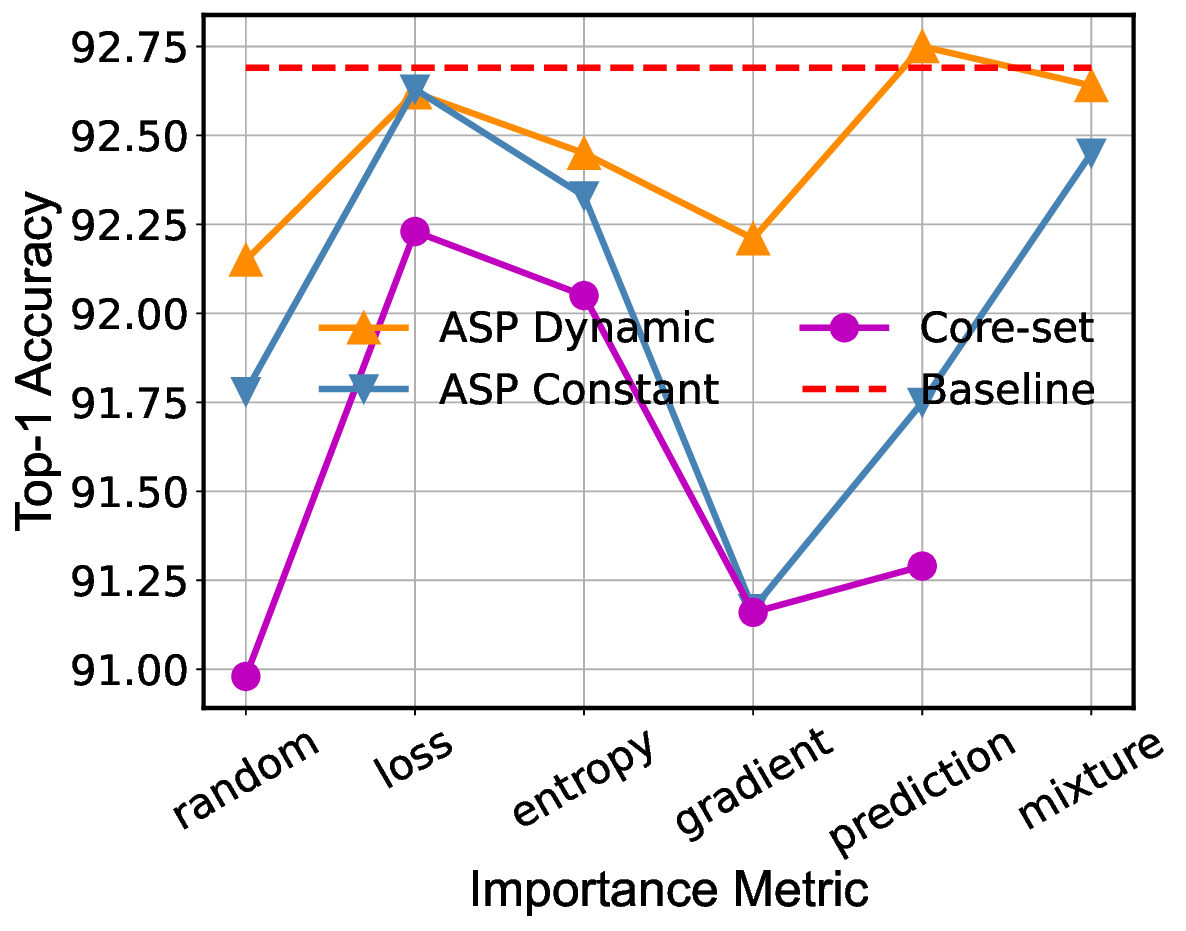}\label{fig:32_c10_d}
}
\caption{Test performance (Top-1 Accuracy(\%)) of ResNet-32 on CIFAR-10 under various data sampling ratios, data selection methods, and importance metrics.}
\label{fig:32_c10}
\end{center}
\end{figure*}

\begin{figure*}[!h]
\begin{center}
\subfigure[Data Ratio 10\%]{
\includegraphics[width=.235\linewidth]{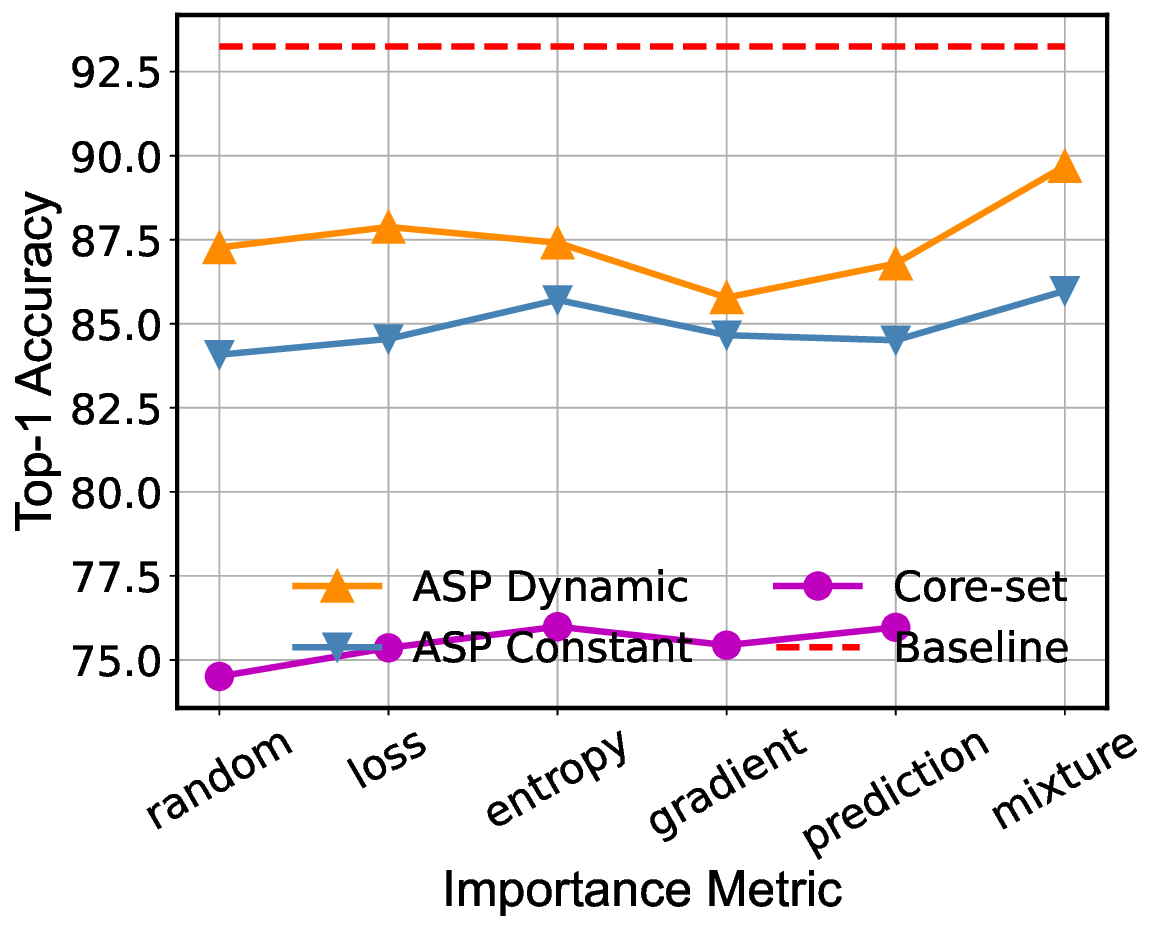}
\label{fig:44_c10_a}
}
\subfigure[Data Ratio 30\%]{
\includegraphics[width=.23\linewidth]{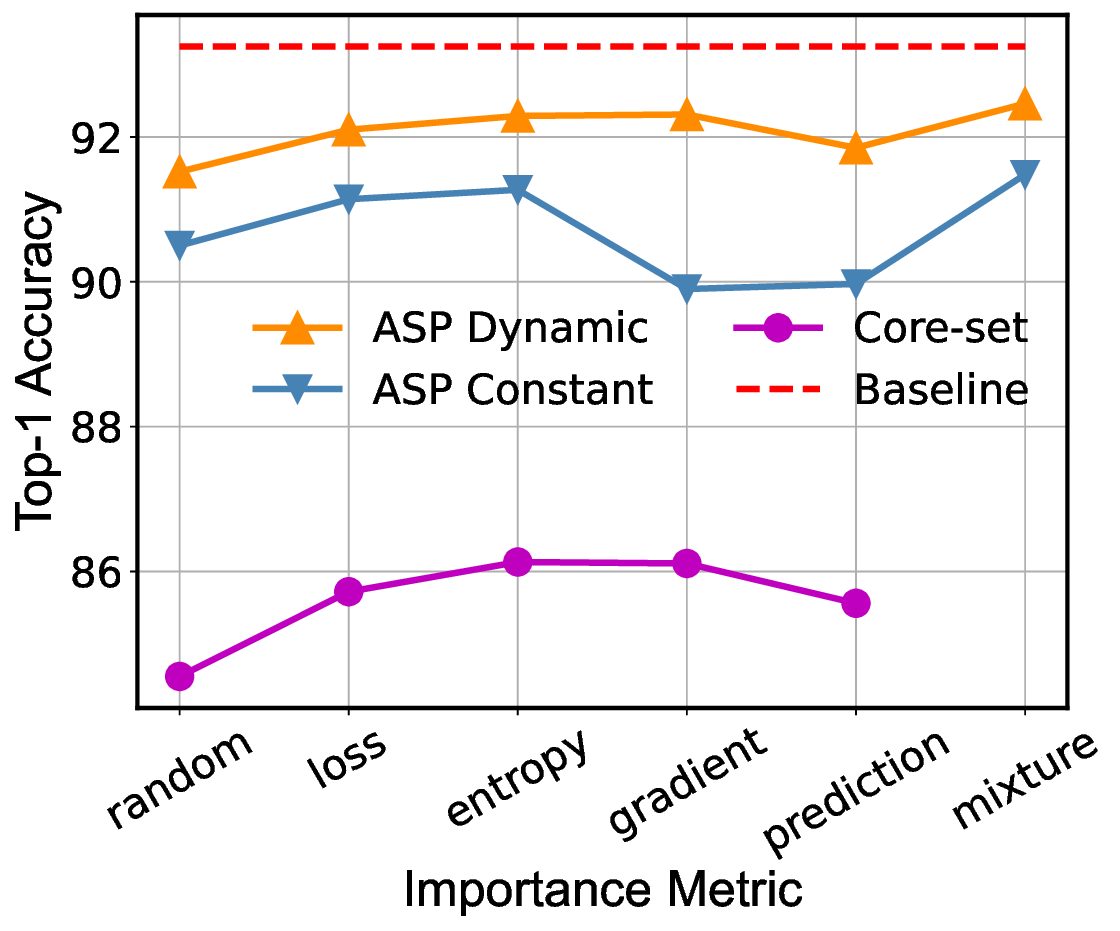}
}
\subfigure[Data Ratio 50\%]{
\includegraphics[width=.23\linewidth]{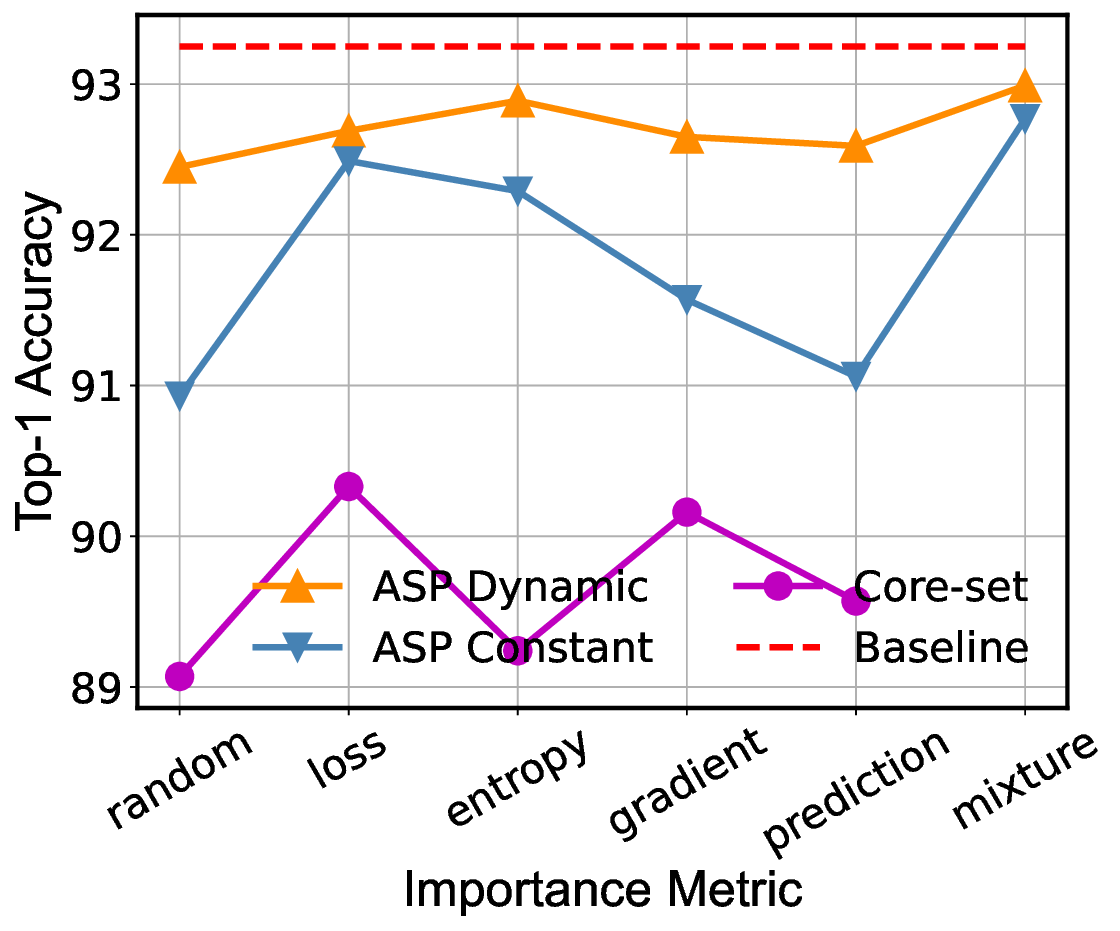}
\label{fig:44_c10_c}
}
\subfigure[Data Ratio 70\%]{
\includegraphics[width=.23\linewidth]{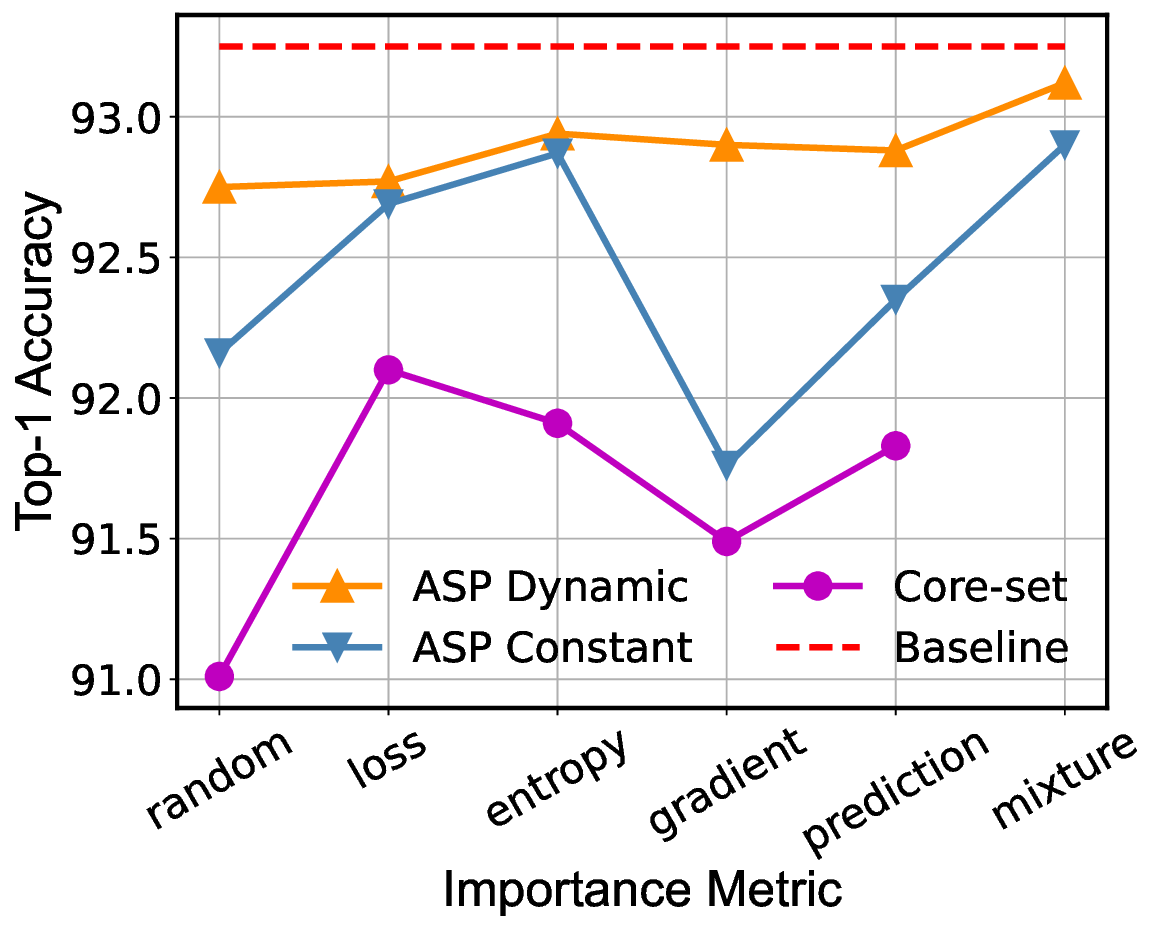}
}
\caption{Test performance (Top-1 Accuracy(\%)) of ResNet-44 on CIFAR-10 under various data sampling ratios, data selection methods, and importance metrics.}
\label{fig:44_c10}
\end{center}
\end{figure*}

\begin{figure*}[!h]
\begin{center}
\subfigure[Data Ratio 30\%]{
\includegraphics[width=.23\linewidth]{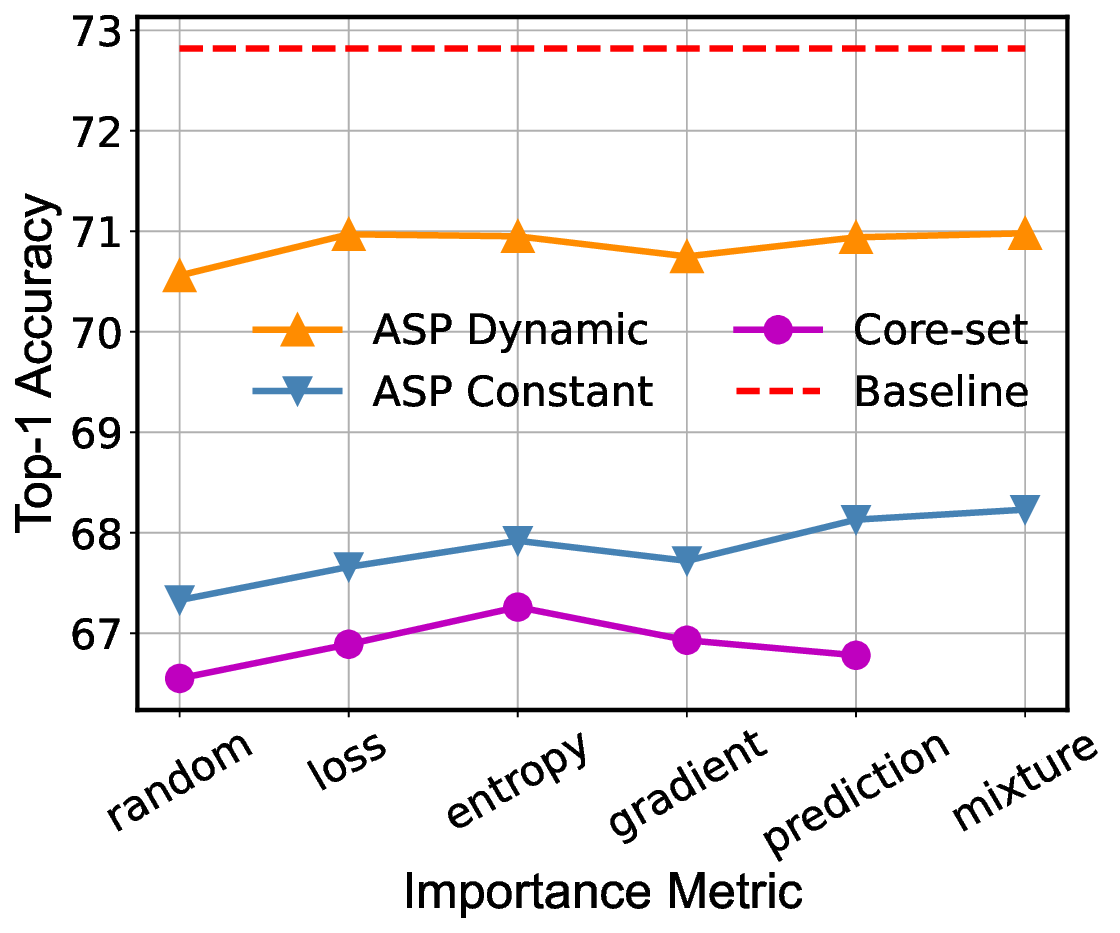}
}
\subfigure[Data Ratio 50\%]{
\includegraphics[width=.23\linewidth]{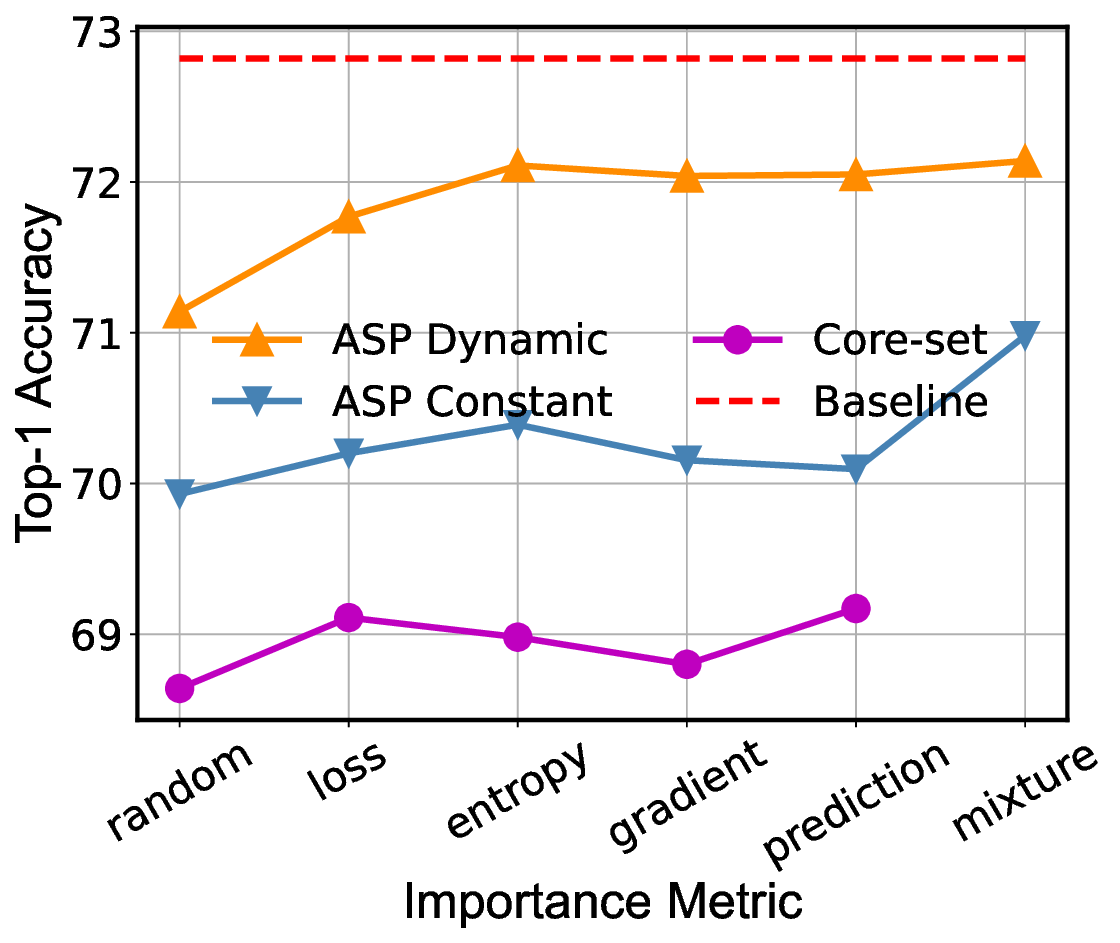}
\label{fig:44_image_c}
}
\subfigure[Data Ratio 70\%]{
\includegraphics[width=.23\linewidth]{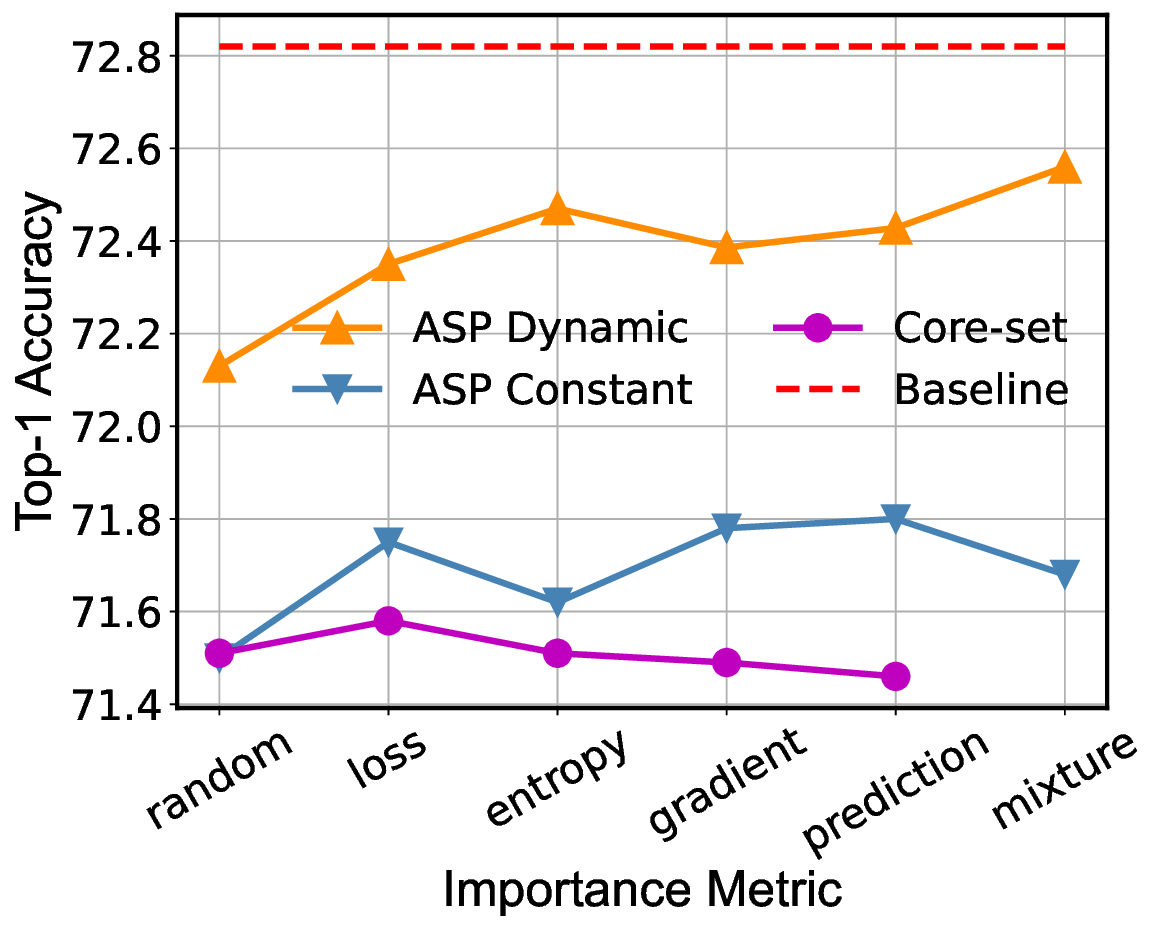}
}
\caption{Test performance (Top-1 Accuracy(\%)) of ResNet-44 on ImageNet-1k under various data sampling ratios, data selection methods, and importance metrics.}
\label{fig:44_image}
\end{center}
\end{figure*}

\paragraph{Performance of Proxy data}
To demonstrate the generalization and robustness of the ASP framework, we have conducted experiments on regular model training (ResNet-18 on the CIFAR-10) in the main text. 
Here, we show the additional experimental result of ASP on more baseline models and datasets. 
As shown in Fig.\ref{fig:18_c100} to \ref{fig:44_image}, we compare various models' performance under different sampling ratio $r$ on different datasets, where $ r \in \{0.1, 0.3, 0.5, 0.7\}$, $r=1.0$ means that using the entire dataset $\mathcal{D}_{train}$.

\textbf{From Fig.\ref{fig:18_c100} to \ref{fig:44_image}, we can observe that ASP achieves better performance compared with different core-set data selection methods on all datasets and baseline models, which demonstrate the robustness and effectiveness of ASP.
Furthermore, it is worth noting that the dynamic sampling ratio scheduler significantly outperforms the constant sampling ratio scheduler, and the mixture of metrics outperforms the single importance metric in almost all experiments, which are mutually corroborated with our previous analysis.}
As shown in Fig.\ref{fig:44_image}, the above phenomenons are more obvious on ImageNet.
What's more, Fig.\ref{fig:32_c10_d}, Fig.\ref{fig:18_c100_d}, Fig.\ref{fig:34_c10_d}, and so on show that after combining with ASP, the performance of the model can even exceed the baseline model trained with all samples, which shows that ASP can indeed deactivate some redundant samples in the dataset.

\section{Visualizations}
\begin{figure*}[!h]
\centering
\includegraphics[width=0.8\columnwidth]{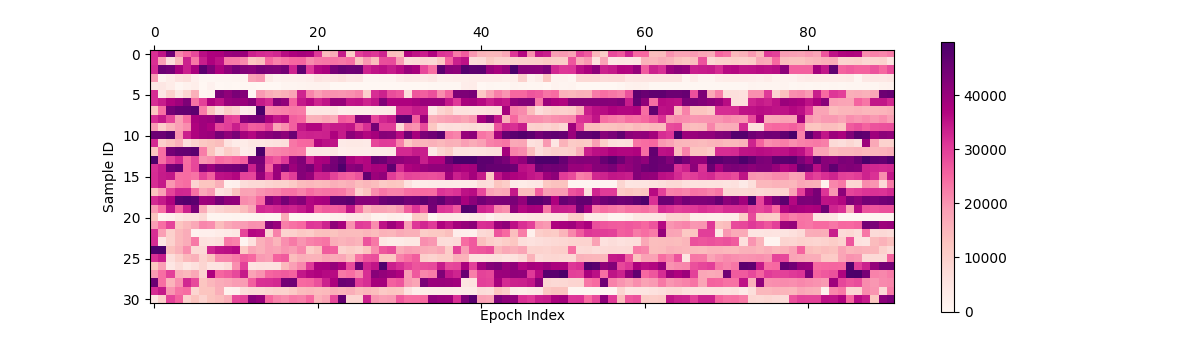}  
\caption{Visualization of input data importance ranking among all the data during training. The larger the value is, the more important the sample is and the higher the probability of being selected.}
\label{importance}
\end{figure*}

\paragraph{Visualizations of Dynamic Samples Rank}
In order to understand the data selection in ASP more intuitively, we randomly selected 30 samples and recorded the changes in their corresponding importance metrics during the 90 epochs of the training process. From Figure \ref{importance}, we can see that there indeed exist easy and hard samples. 
Some samples have high-importance metrics throughout the training process, which means that they will be activated for training with a high probability throughout the training process. On the other hand, there are also samples with a low probability of being activated throughout the training process.
Meanwhile, the importance metric of samples changes dynamically with training instead of rising or falling monotonically, which makes the replayable data selection possible.

\begin{figure*}[!h]
\centering  
\subfigure[Easiest 1 example for each class.]{
\label{Fig.sub.1}
\includegraphics[width=0.33\textwidth]{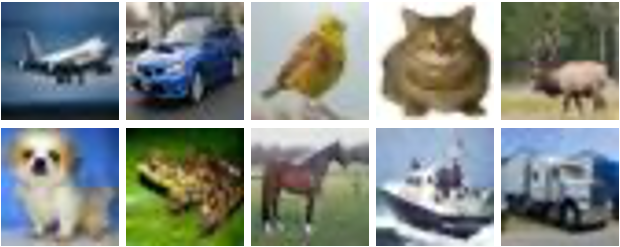}}
\subfigure[Hardest 1 example for each class.]{
\label{Fig.sub.2}
\includegraphics[width=0.33\textwidth]{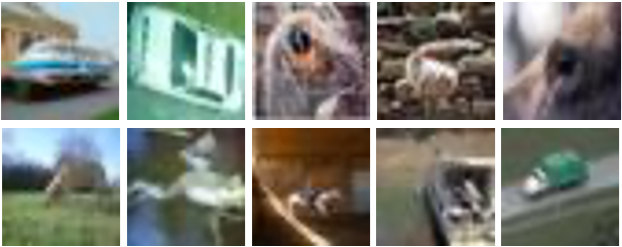}}
\subfigure[Easiest 10 examples for the dataset.]{
\label{Fig.sub.3}
\includegraphics[width=0.33\textwidth]{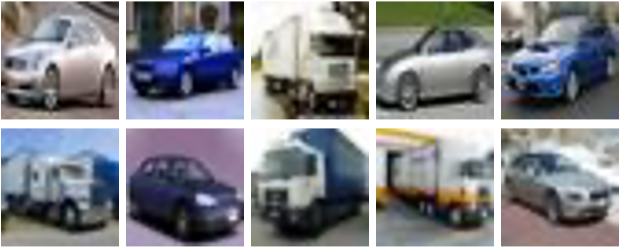}}
\subfigure[Hardest 10 examples for the dataset]{
\label{Fig.sub.4}
\includegraphics[width=0.33\textwidth]{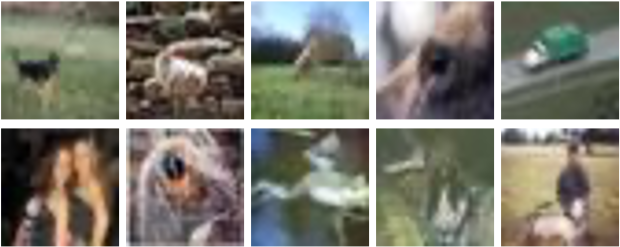}}
\caption{The easiest and hardest examples in the CIFAR10 dataset.}
\label{Fig.main}
\end{figure*}

\paragraph{Visualizations of Hard and Easy Samples}
As we mentioned in the main text, the importance metric of each sample will be updated by PMM $V$, the higher the importance metric indicates the harder the sample is to be identified, and vice versa.
We collect all the importance values of samples within 200 epochs and calculate the average importance metric over 200 epochs. 
Figure \ref{Fig.sub.1} and Figure \ref{Fig.sub.2} visualizes the easiest 10 samples and the hardest samples among the dataset respectively.
In addition, we divide all the data by category and show the easiest and hardest sample of each category in Figure \ref{Fig.sub.3} and Figure \ref{Fig.sub.4}.
Obviously, this result is consistent with human intuition, automobiles and trucks are more easily to be recognized, since they have a more regular shape, and the color background in the image is simple. Nevertheless, the 10 hardest images have blurred backgrounds, making it difficult to recognize objects in the images. 

\begin{figure*}[!ht]
\begin{center}
\subfigure[normal cell]{
\includegraphics[width=.4\linewidth]{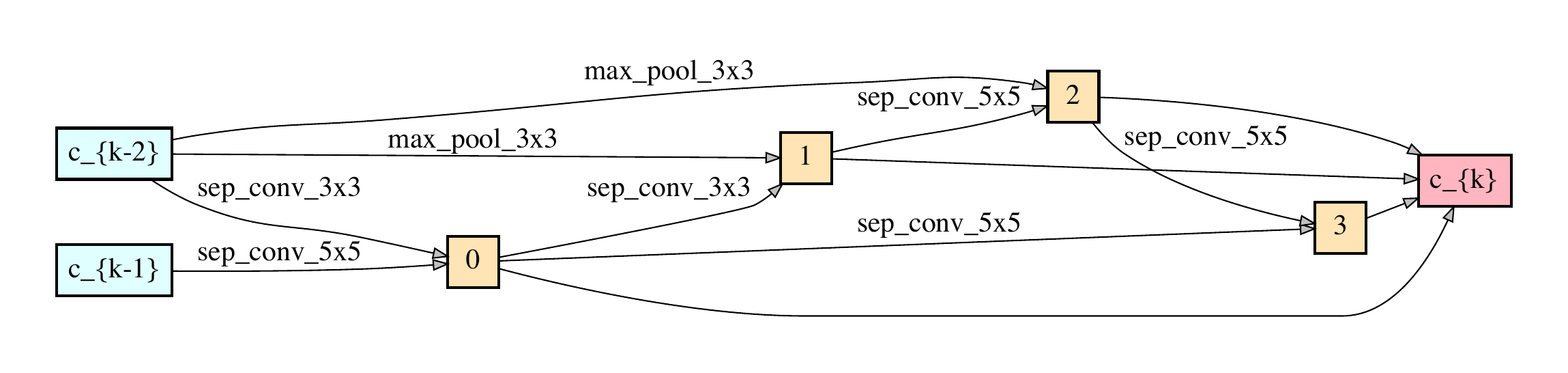}
\label{fig:darts_asp_A_n}
}
\subfigure[reduction cell]{
\includegraphics[width=.4\linewidth]{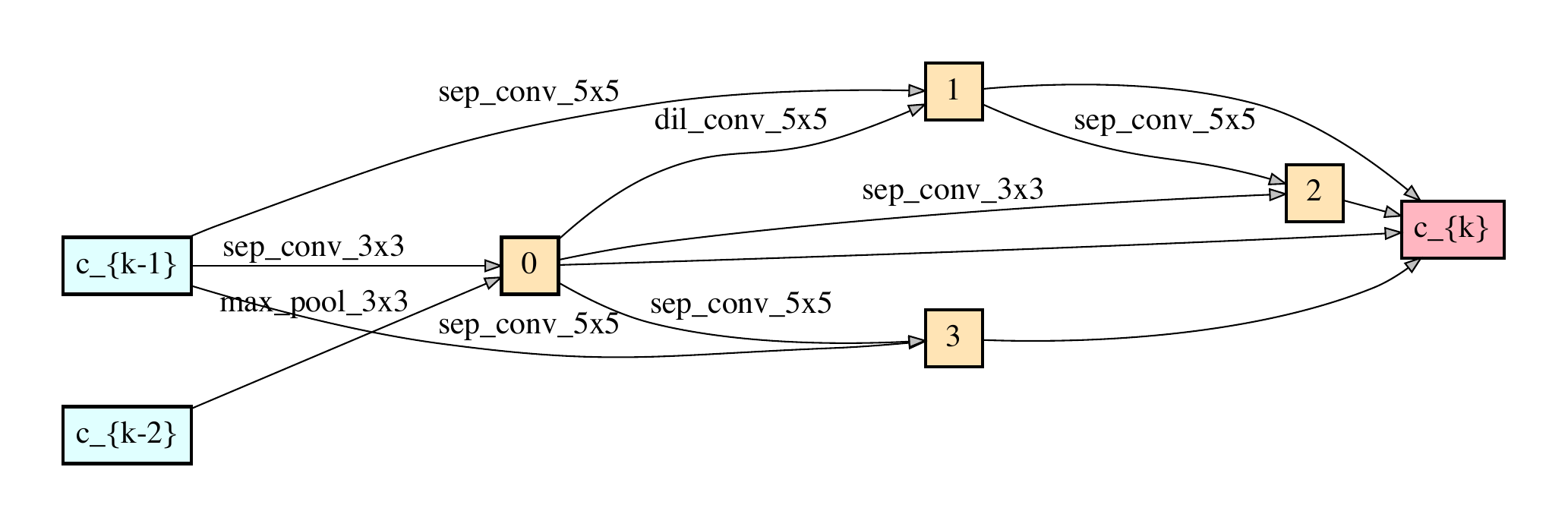}\label{fig:darts_asp_A_r}
}
\caption{DARTS-ASP A.}
\label{fig:darts_asp_A}
\end{center}
\end{figure*}

\begin{figure*}[!ht]
\begin{center}
\subfigure[normal cell]{
\includegraphics[width=.4\linewidth]{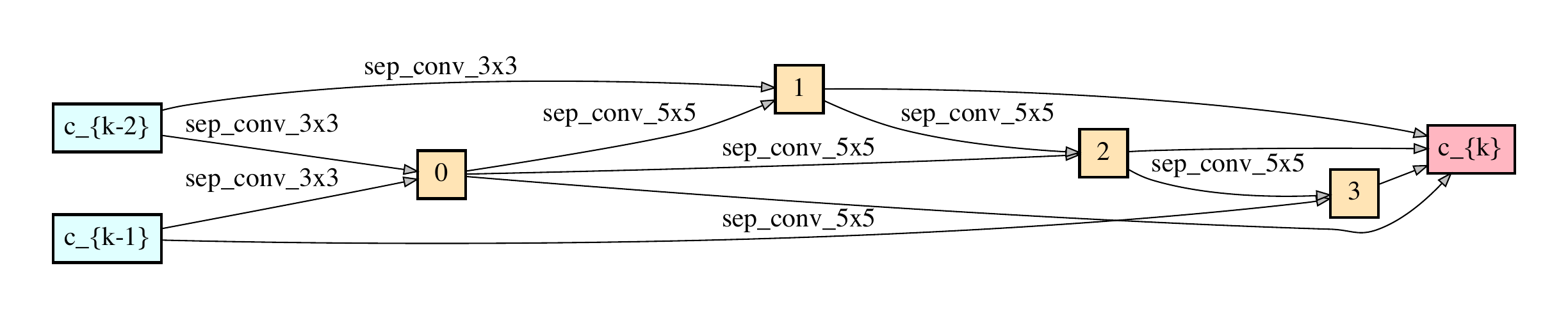}
\label{fig:darts_asp_B_n}
}
\subfigure[reduction cell]{
\includegraphics[width=.4\linewidth]{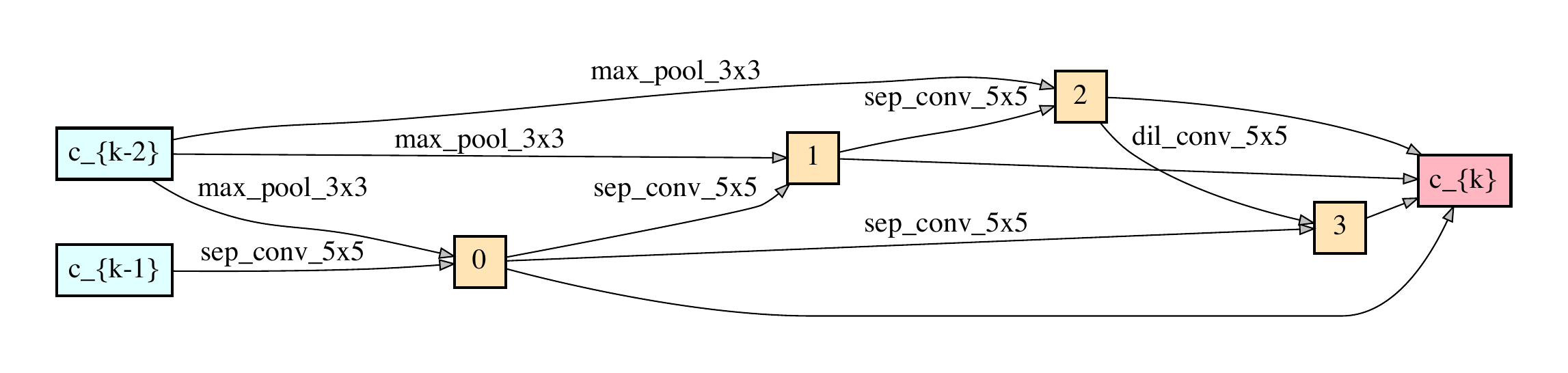}\label{fig:darts_asp_B_r}
}
\caption{DARTS-ASP B.}
\label{fig:darts_asp_B}
\end{center}
\end{figure*}

\begin{figure*}[!ht]
\begin{center}
\subfigure[normal cell]{
\includegraphics[width=.4\linewidth]{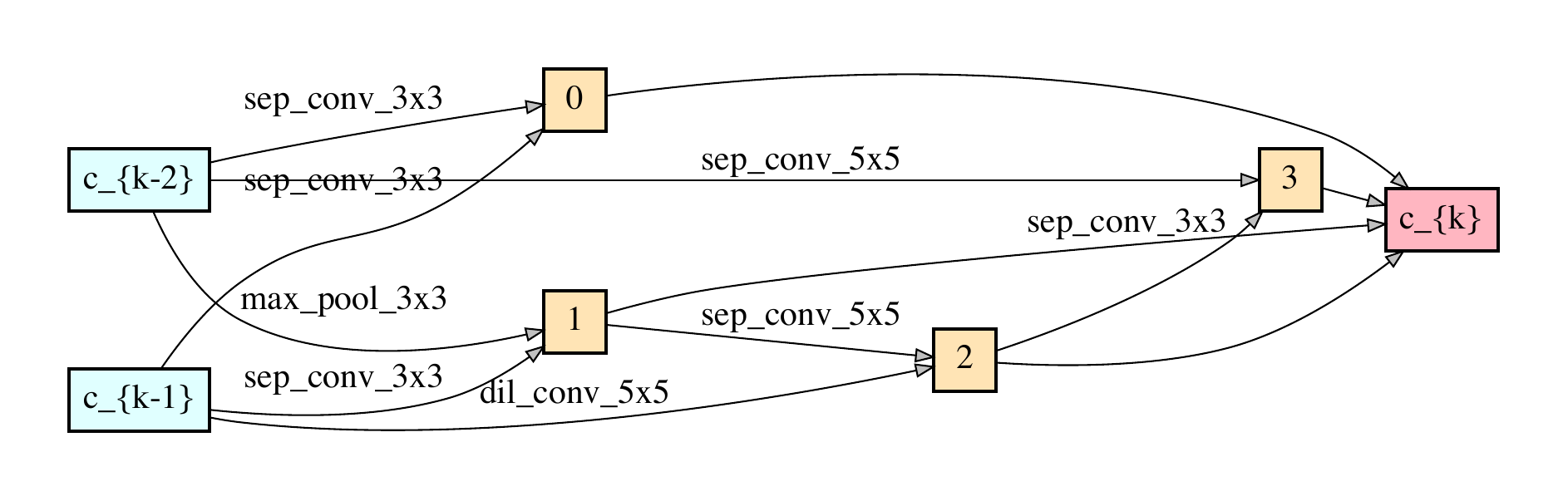}
\label{fig:darts_asp_C_n}
}
\subfigure[reduction cell]{
\includegraphics[width=.4\linewidth]{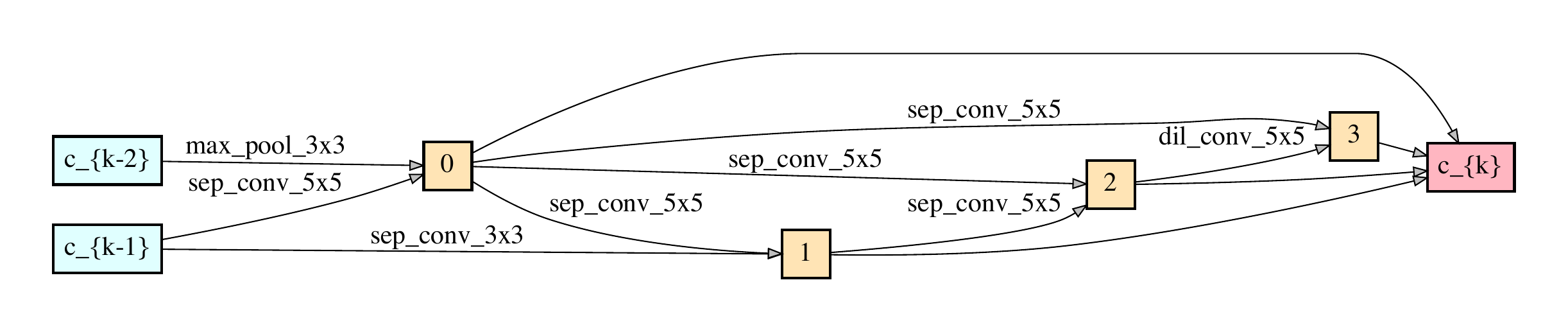}\label{fig:darts_asp_C_r}
}
\caption{DARTS-ASP C.}
\label{fig:darts_asp_C}
\end{center}
\end{figure*}

\paragraph{Visualizations of NAS Architecture}
We visualize the searched architecture on the DARTS search space in Figure \ref{fig:darts_asp_A} to Figure \ref{fig:darts_asp_C}, noted as DARTS-ASP.


\section{Implementation Details}
\subsection{Dataset}
\paragraph{CIFAR}
CIFAR-10 and CIFAR-100 datasets~\cite{krizhevsky2009learning} both consist of 50,000 training images and 10,000 test images distributed equally across all classes and are resized into 32x32 pixels. CIFAR-10 has 10 classes and each class has 5,000 training images and 1,000 test images. CIFAR-100 has 100 fine-grained classes and each class has 500 training images and 100 test images. 
\paragraph{ImageNet}
ImageNet-1k~\cite{krizhevsky2012imagenet} is a more scalable dataset that consists of 1.28 million training images and 50,000 validation images with 1,000 classes. The images from the ImageNet-1k dataset are not always consistent and we need to resize them to 256$\times$256 and then randomly/centrally crop them into 224$\times$224 in the training/test stage. ImageNet-16-120~\cite{dong2019bench} is another down-sampled dataset derived from ImageNet-1k, which resizes the image into 16$\times$16 and selects images with label $\in \left[1, 120\right]$. It contains 151.7K training images, 3K validation images, and 3K test images with 120 classes.

\subsection{Baseline Models}
Except for the experiments of neural architecture search, we mainly adopt the various ResNet~\cite{he2016deep} architectures as our base models in other experiments.
Because it is the most popular architecture in deep learning and has better diversity, covering from lightweight models (e.g. ResNet-18) to large models (e.g. ResNet-101). 
It is easy for researchers to just modify the architectures a little and easily adjust them to different datasets, such as 32-pixel inputs for CIFAR and 224-pixel inputs for ImageNet-1K. 
We display several types of ResNet architectures and configures as shown in Tab.~\ref{two_res}.

\begin{table*}
\centering
\scalebox{0.55}{
\begin{tabular}{c|c|c|c|c|c|c|c}
\specialrule{0.1em}{1pt}{1pt}
 Name      &  Size        & ResNet-18      & ResNet-34        & ResNet-101          & ResNet-20       & ResNet-44          & ResNet-56                    \\ 
\specialrule{0.05em}{1pt}{1pt}
Conv1           & 32$\times$32        & \multicolumn{3}{c|}{3x3, 64, stride 1}                    & \multicolumn{3}{c}{3x3, 16, stride 1}                                 \\ 
\specialrule{0.05em}{1pt}{1pt}
Conv2\_x        & 32$\times$32        & $\begin{bmatrix} 3\times3, 64\\ 3\times3, 64 \end{bmatrix}\times 2$       & $\begin{bmatrix} 3\times3, 64\\ 3\times3, 64 \end{bmatrix}\times 3$      & $\begin{bmatrix} 1\times1, 64\\ 3\times3, 64\\ 1\times1, 256 \end{bmatrix}\times 3 $  
                                      & $\begin{bmatrix} 3\times3, 16\\ 3\times3, 16 \end{bmatrix}\times 3 $             & $\begin{bmatrix} 3\times3, 16\\ 3\times3, 16 \end{bmatrix}\times 7 $          & $\begin{bmatrix} 3\times3, 16\\ 3\times3, 16 \end{bmatrix}\times 9 $   \\ 
\specialrule{0.05em}{1pt}{1pt}
Conv3\_x        & 16$\times$16        & $\begin{bmatrix} 3\times3, 128\\ 3\times3, 128 \end{bmatrix}\times 2$     & $\begin{bmatrix} 3\times3, 128\\ 3\times3, 128 \end{bmatrix}\times 4$     & $\begin{bmatrix} 1\times1, 128\\ 3\times3, 128\\ 1\times1, 512 \end{bmatrix}\times 4 $
                                    & $\begin{bmatrix} 3\times3, 32\\ 3\times3, 32 \end{bmatrix}\times 3 $             & $\begin{bmatrix} 3\times3, 32\\ 3\times3, 32 \end{bmatrix}\times 7 $          & $\begin{bmatrix} 3\times3, 32\\ 3\times3, 32 \end{bmatrix}\times 9 $   \\
\specialrule{0.05em}{1pt}{1pt}
Conv4\_x         & 8$\times$8         & $\begin{bmatrix} 3\times3, 256\\ 3\times3, 256 \end{bmatrix}\times 2$     & $\begin{bmatrix} 3\times3, 256\\ 3\times3, 256 \end{bmatrix}\times 6$     & $\begin{bmatrix} 1\times1, 256\\ 3\times3, 256\\ 1\times1, 1024 \end{bmatrix}\times 23 $
                                    & $\begin{bmatrix} 3\times3, 64\\ 3\times3, 64 \end{bmatrix}\times 3 $             & $\begin{bmatrix} 3\times3, 64\\ 3\times3, 64 \end{bmatrix}\times 7 $          & $\begin{bmatrix} 3\times3, 64\\ 3\times3, 64 \end{bmatrix}\times 9 $   \\ 
\specialrule{0.05em}{1pt}{1pt}
Conv5\_x         & 4$\times$ 4        & $\begin{bmatrix} 3\times3, 512\\ 3\times3, 512 \end{bmatrix}\times 2$     & $\begin{bmatrix} 3\times3, 512\\ 3\times3, 512 \end{bmatrix}\times 3$     & $\begin{bmatrix} 1\times1, 512\\ 3\times3, 512\\ 1\times1, 2048 \end{bmatrix}\times 3 $
                                    &                                                           &                                                       &  \\ 
\specialrule{0.05em}{1pt}{1pt}
               & 1$\times$1              & \multicolumn{6}{c}{average pool, 10-d fc, softmax}                \\ 
\specialrule{0.05em}{1pt}{1pt}
\multicolumn{2}{c|}{Params(M)}                   & 11,173         & 21,282         & 42,512          & 269             & 658                      & 853                           \\ 
\specialrule{0.1em}{1pt}{1pt}
\end{tabular}
}
\caption{Two different hidden channel numbers for ResNet architectures for CIFAR-10. Downsampling is performed by conv3\_1, conv4\_1, and conv5\_1 with a stride of 2.}
\label{two_res}
\end{table*}

\subsection{Hyper-parameters}
\paragraph{ResNet Settings}
We run all the experiments on NVIDIA 2080Ti GPU and PyTorch 1.5.1 version. 
All the models are trained with 200 epochs and a batch size of 256 for ImageNet-1k and 128 for others. 
We use the SGD optimizer with a learning rate of 0.1 and a momentum of 0.9.
For the ResNet-20, ResNet-32, and ResNet-44, we set the weight decay of the optimizer as 1e-4 and adopt the MultiStepLR scheduler, which starts with a learning rate of 0.1 and decays to 0.1 times at 100 epochs and 150 epochs.
As for ResNet-18, ResNet-34, and ResNet-101, we set the weight decay of the optimizer to be 5e-4 and adopt CosineAnnealingLR \cite{loshchilov2016sgdr} scheduler with a starting learning rate of 0.1 and gradually decay to 1e-5.
By default, except for standard data preprocessing, i.e. random horizontal flip, a random crop of 224 pixels for ImageNet-1k, and 32 for others, We don't adopt any other data augmentation unrelated to the ASP purpose, such as Cutout \cite{devries2017improved} and Mixup\cite{zhang2018mixup}.

\paragraph{NAS-Bench-201 Settings}
NAS-Bench-201 \cite{dong2019bench} constructs the supernet by repeated 15 cells and each cell is represented by a densely connected DAG, which assigns six transforming edges between four data nodes. 
Each edge has five types of operations that are commonly used in the NAS literature: $O=\{3\times 3 Conv, 3\times 3 Avg, 1\times 1 Conv, Skip, Zeroize\}$. 
Finally, it has a total of 15,625($5^6$) sub-model structures. \cite{dong2020bench} provides the validation and test accuracy of each sub-model in CIFAR10, CIFAR100 and ImageNet-16-120 datasets. 
We train the supernet for 250 epochs and only adopt the ASP to select proxy data for search acceleration. Specifically, we use the SGD optimizer with a 0.05 learning rate and 2.5e-4 weight decay value. We use the SPOS \cite{guo2020single} search algorithm and randomly sample a path to update in the training.

\paragraph{DARTS Settings}
DARTS \cite{liu2018darts} introduces a node-like search space and each cell consists of N=7 nodes, among which the output node is defined as the depthwise concatenation of all the intermediate nodes (input nodes excluded). there are 8 candidate operations between any two nodes, 
i.e.$O$: 3 x 3 and 5 x 5 separable convolutions, 3 x 3 and 5 x 5 dilated separable convolutions, 3 x 3 max pooling, 3 x 3 average pooling, identity, and zero. 
In the search stage, we adopt the ASP to select the proxy subset and then split it into training and valid sets. The other training setting on DARTS search space is similar to DARTS settings \cite{liu2018darts}.
Specifically, we search for the best architecture for 50 epochs with 16 as initial channels, and then we re-train the searched architecture for 600 epochs with 36 initial channels. Meanwhile, we also adopt the Cutout with the Cutout length of 16 to augment the training dataset.

\paragraph{HPO Settings}
Hyper-parameters optimization is an extremely time-consuming task and in order to apply HPO algorithms in large-scale data scenarios, we adopt three algorithms with high efficiency as our baseline. i.e. Hyperband \cite{li2017hyperband} , BOHB \cite{falkner2018bohb}, and Hyper-Tune \cite{li2022hyper}. We conduct all the HPO-related experiments with the ResNet-18 model and the same setting with \textbf{ResNet Settings}. 
We adopt an asynchronous parallel acceleration method \cite{li2022hyper} and use 8 processes to work in parallel with 2 GPU days. So we set $2\times8=16$ GPU days as the total time budget. When we adopt the ASP to select proxy data with the sampling ratio r, e.g., 10\%, we will adjust the time budget to r (10\%) of the original time budget.

\end{document}